\documentclass[examplefnt,biber]{nowfnt} 

\usepackage[utf8]{inputenc}
\usepackage{etoolbox}

\usepackage{amsfonts}
\usepackage{amsmath} 
\usepackage{amssymb}
\usepackage{stfloats}
\usepackage{graphicx}
\usepackage{caption}
\usepackage{subcaption}
\usepackage{tikz}
\usepackage{placeins}
\usepackage{booktabs}
\usepackage{tabularx}
\usepackage{array}

\DeclareMathOperator*{\argmax}{arg\,max}
\DeclareMathOperator*{\argmin}{arg\,min}

\renewcommand{\S}{\mathcal{S}}
\newcommand{\A}{\mathcal{A}}
\newcommand{\E}{\mathbb{E}}
\renewcommand{\H}{\mathcal{H}}
\newcommand{\Real}[1]{\mathbb{R}^{#1}}

\title{Ergodic Control and Controlled Diffusion for Robot Learning}

\subtitle{Review and Tutorial}

\maintitleauthorlist{
Max Muchen Sun \\
Northwestern University \\
msun@u.northwestern.edu
\and
Cem Bilaloglu \\
Idiap Research Institute \\
cem.bilaloglu@epfl.ch
\and 
Ananya Rao \\
Carnegie Mellon University \\
ananyara@cs.cmu.edu
\and 
Stefan Ivic \\
University of Rijeka \\
stefan.ivic@riteh.uniri.hr
\and 
Guillaume Sartoretti \\
National University of Singapore \\
guillaume.sartoretti@nus.edu.sg
\and 
Kathleen Fitzsimons \\
Pennsylvania State University \\
k-fitzsimons@psu.edu 
\and 
Ian Abraham \\
University of Sydney/Yale University \\
ian.abraham@sydney.edu.au
\and 
Sylvain Calinon \\
Idiap Research Institute \\
sylvain.calinon@idiap.ch
\and 
Todd Murphey \\
Northwestern University \\
t-murphey@northwestern.edu
}

\issuesetup
{%
 }

\usepackage{mwe}

\author[1]{Max Muchen Sun}
\author[2]{Cem Bilaloglu}
\author[3]{Ananya Rao}
\author[4]{Stefan Ivic}
\author[5]{Guillaume Sartoretti}
\author[6]{Kathleen Fitzsimons}
\author[7,8]{Ian Abraham}
\author[2]{Sylvain Calinon}
\author[1]{Todd Murphey}

\affil[1]{Northwestern University}
\affil[2]{Idiap Research Institute}
\affil[3]{Carnegie Mellon University}
\affil[4]{University of Rijeka}
\affil[5]{National University of Singapore}
\affil[6]{Pennsylvania State University}
\affil[7]{University of Sydney}
\affil[8]{Yale University}

\begin{document}

\makeabstracttitle
\allowdisplaybreaks

\begin{abstract}
Diffusion learning leverages the statistical mechanism of diffusion processes for learning, reasoning, and inferring complex distributions from data. Recent advances in diffusion learning have been transformative, with robot learning emerging as a key opportunity area, with applications spanning perception, control, and decision-making. At the same time, the statistical mechanism of diffusion processes can be controlled to shape the temporal evolution of the state distribution underlying robot trajectories, inducing ergodic behavior in robotic systems. The frameworks of controlled diffusion and ergodic control were developed around the same time as diffusion learning, and their theories and algorithms have increasingly converged. Ergodicity induced by controlled diffusion has several significant implications for robot learning, distinct from applying diffusion learning to robotics problems: it formally enforces statistical properties required to ensure optimality of robot learning, enables non-myopic search over uncertain information landscapes for data collection, and enables behavior specification based on spatial rather than temporal characteristics of trajectories. This survey introduces the intuition behind controlled diffusion for robot learning, explores its connection to diffusion learning, presents theoretical foundations and numerical tutorials for solving controlled diffusion and ergodic control problems, and reviews applications across robotics. Finally, we discuss key challenges and future opportunities in leveraging controlled diffusion for robot learning.
\end{abstract}

\chapter{Introduction}
\label{chap:intro}

Learning is fundamentally about generating predictions from examples. This process has two core aspects. The first is \textbf{pattern recognition}: how to effectively extract, represent, and generalize patterns from data so that predictions can be accurate and robust. The second is \textbf{data collection}: determining what kind of data is needed to capture the structure of a problem and provide the right information for learning. Over the past two decades, deep learning has driven transformative progress across nearly every aspect of society by advancing both of these dimensions. Models and architectures have grown more powerful, while datasets have become larger, more diverse, and more carefully curated for specific domains.

\begin{figure}[t!]
    \centering
    \includegraphics[width=\linewidth]{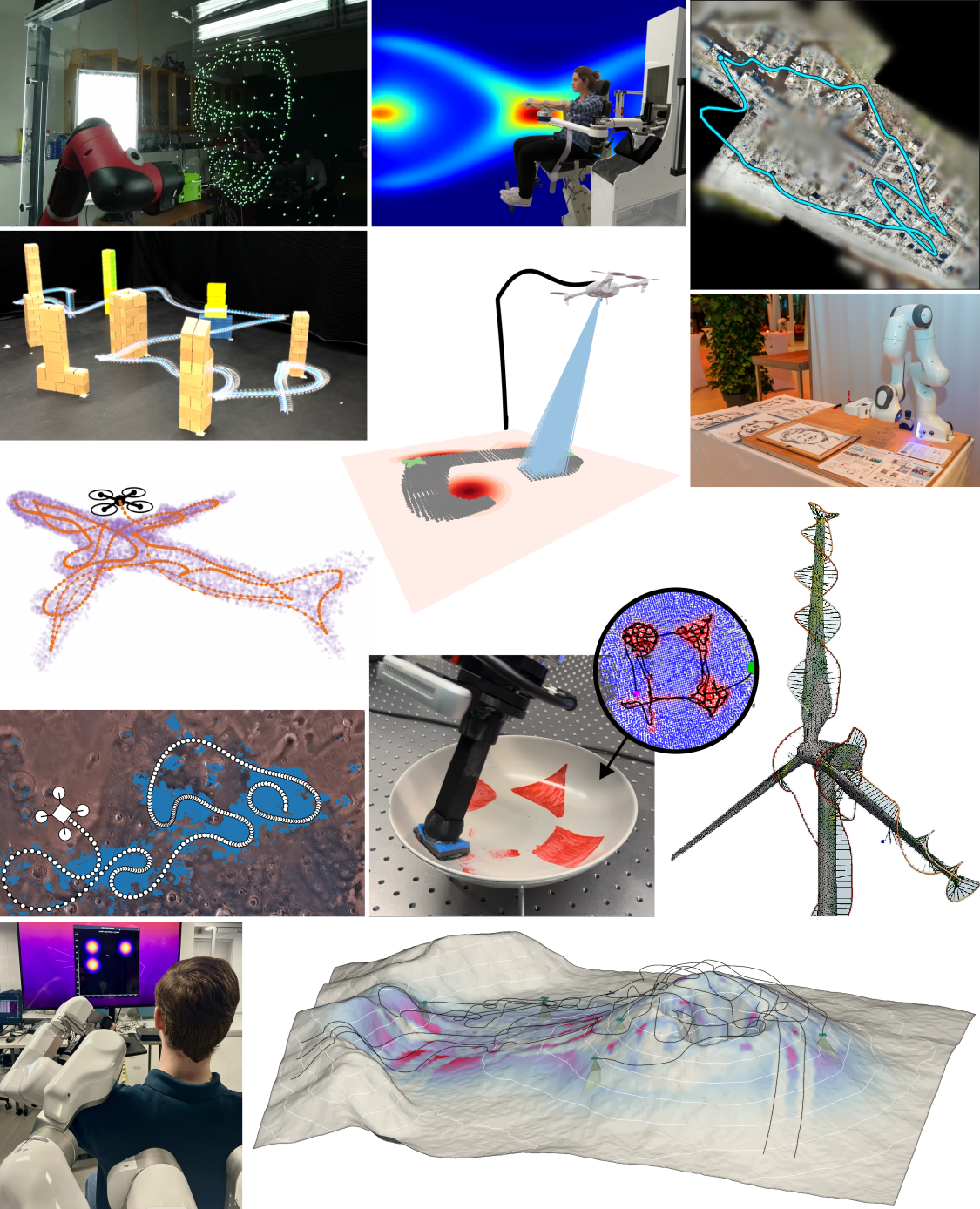}
    \caption{Illustrations of ergodic trajectories and ergodic control in action.}
    \label{fig:overview}
\end{figure} 

Robotics is one of the fields most deeply influenced by these advances in learning. Realizing the full potential of robots in society requires them to operate autonomously in unstructured, dynamic, and entirely novel environments. Robots must continuously adapt to their surroundings—including human interactions—without constant human intervention. While robotics research has established strong foundations in perception, control, and decision-making, conventional approaches often lack the complexity and adaptability needed for this vision. Learning-based methods hold the promise to close this gap. Recent breakthroughs already demonstrate this potential: high-fidelity perception systems enabling self-driving cars, learning-based controllers for agile locomotion in difficult terrains, dexterous manipulation of complex objects, and large model-driven autonomy that allows robots to navigate and collaborate in human-centered spaces.

The rapid progress in the pattern recognition aspect of machine learning has brought profound benefits to robotics. The breakthrough performance of AlexNet~\citep{krizhevsky_imagenet_2012} on the ImageNet dataset unlocked the potential of semantic understanding as a foundation for robotic perception. Deep reinforcement learning (DRL) ~\citep{levine_end--end_2016} has leveraged neural network-based policy representation, combined with computationally accessible simulators, to produce policies capable of highly agile and flexible locomotion and dexterous manipulation. Over the last five years, diffusion models have emerged as one of the most powerful machine learning methods~\citep{sohl-dickstein_deep_2015, ho_denoising_2020, lipman_flow_2022}. These models employ the statistical mechanism of diffusion processes to approximate and sample from highly complex probability distributions through iterative denoising, enabling precise modeling of multimodal data in a computationally tractable manner. The strength of diffusion models has been demonstrated across a broad range of domains—language, image, audio, and video generation—where they achieve unprecedented quality and diversity, and their application to robotics is growing rapidly. In perception, diffusion-enabled models are pushing robot sensing and scene understanding closer than ever to human-level capabilities, allowing richer, probabilistic representations of geometry, semantics, and temporal dynamics that can support decision-making under uncertainty. In planning and control, diffusion-based policy representations have emerged as a flexible and expressive framework that can capture multimodal distributions over trajectories, enabling robots to generate smooth, diverse, and context-aware action sequences. These models have shown promise in imitation learning, where they leverage high-quality human demonstrations to reproduce complex behaviors, and in deep reinforcement learning, where they exploit large-scale simulation data to improve sample efficiency and generalization. By unifying generative modeling and control, diffusion models open the door to scalable, probabilistic approaches for robot learning that integrate perception, reasoning, and decision-making. Taken together, these developments point toward a broader vision for robotics: developing foundation models for robotics—large-scale, versatile models that unify perception, reasoning, and control while leveraging the strengths of diffusion modeling.

On the other hand, the question of what data is needed and how to collect it—complementing the progress in pattern recognition—remains largely unanswered. The prevailing strategy behind large language models and image generation models has been to exhaustively collect and train on vast amounts of data. While this data-at-scale paradigm has driven transformative progress in certain domains, it also introduces fundamental limitations, including escalating energy consumption, privacy and security concerns, and a lack of formal guarantees around safety and reliability. This challenge is even more complex in robotics, where data collection is inherently expensive and where the structure of data differs fundamentally from that of images, text, or other static modalities. Robots are embodied agents operating under physical constraints such as rigid-body geometry, actuation limits, and nonlinear dynamics, all of which govern not only what data they can gather but also how that data is distributed. These constraints introduce strong inductive biases: observations are local and partial, samples are tied to the robot’s physical configuration and trajectory, and data is inherently shaped by the robot’s motion and perception. A particularly critical issue is the violation of the independently and identically distributed (i.i.d.) assumption, which underpins most learning algorithms. Unlike images or text samples that can be shuffled and batched independently, robotic data is sequential, with heavy temporal correlations caused by physical dynamics and closed-loop control. Each action or motion constrains future states and observations, producing highly correlated trajectories through state space rather than independent samples. These correlations are further compounded by environmental dependencies, non-stationary dynamics (e.g., wear, shifting payloads, changing terrain), and interactions with other agents including humans, leading to datasets that are non-i.i.d., non-stationary, and context-dependent. This fundamental issue cannot be solved by naively scaling data collection, making it necessary to rethink how robots acquire and structure data and develop methods that explicitly account for embodiment and temporal correlation.

Interestingly, around the same time that diffusion learning models were being developed, the same statistical mechanism of diffusion was also applied to robot motion synthesis from a control optimization perspective. In diffusion learning, a diffusion process defines a forward–reverse mapping between distributions: data samples are progressively perturbed with noise until they approximate a simple prior, and a learned denoising process then iteratively transforms samples from this prior into complex samples that match the statistics of a data distribution. Controlled diffusion applies this same iterative principle to robot trajectories rather than static data samples. Starting from a simple, dynamically feasible trajectory, a control law derived from the diffusion process incrementally adjusts the trajectory so that the ensemble of visited states matches a specified spatial distribution, which can either be analytically designed or represented by data. In this setting, the diffusion mechanism defines a stochastic process over deterministic robot dynamics, resulting in trajectories whose long-term spatial statistics converge to a stationary distribution. This property, known as ergodicity, means that the time-averaged statistics of a system’s trajectory are equivalent to its space-averaged or ensemble statistics. Controlled diffusion therefore adapts the same mathematical foundation used in generative modeling—iteratively refining a simple, tractable distribution into a structured target distribution—to the design of robot motion, focusing on achieving statistical coverage of an environment based on a desired distribution.

Ergodicity induced by controlled diffusion has several profound implications for robot data collection, which are distinct from merely applying diffusion learning models to robotics tasks. First, it has been formally established that ergodic, diffusion-driven trajectories reduce or eliminate temporal correlations in robot data, enabling datasets that approximate the i.i.d. assumption while remaining consistent with the robot’s dynamics and physical constraints. This is a critical property because nearly all modern learning algorithms rely on i.i.d. data—a condition rarely satisfied in embodied systems without explicitly designed motion strategies. Second, ergodic trajectories inherently support optimal, non-myopic exploration in environments characterized by non-convex, sparse, or uncertain information distributions. By optimizing spatial statistics rather than local objectives, they ensure that robots systematically cover all regions of interest and avoid short-sighted, greedy exploration strategies. Lastly, ergodicity offers a principled behavioral specification framework rooted in spatial coverage patterns rather than explicit temporal sequencing, allowing designers to prescribe where and how frequently a robot should sample regions of its environment without dictating exact trajectories. The algorithmic framework for achieving this property, known as ergodic control, has seen significant theoretical and computational advances over the past decade and has been deployed in diverse robotics applications, including autonomous information gathering, large-scale environmental monitoring, dexterous manipulation, and human–robot interaction. These developments demonstrate that ergodic control is not merely a motion planning technique but a general mathematical framework for designing efficient and task-agnostic data collection strategies for embodied systems.

Diffusion learning and controlled diffusion were developed around the same time, and in recent years, their underlying theory and algorithms have begun to converge, supported by advances in computational resources and numerical methods. This paper reviews the progress made in controlled diffusion and ergodic control for robot learning, highlight the connections between diffusion learning and controlled diffusion, introduce the theoretical foundations and algorithmic implementations of ergodic control to the broader robotics community, and examine representative applications. By consolidating this knowledge, the review seeks to motivate future research directions and provide insight into a central question for the field: \textbf{how should robots collect data to learn effectively, efficiently, and autonomously?} This convergence of ideas presents an exciting opportunity to treat diffusion not only as a powerful paradigm for pattern recognition, as demonstrated by recent breakthroughs in generative modeling, but also as a unifying principle for robot action and data acquisition. By leveraging the mathematical foundations of diffusion processes, the robotics community can begin to develop systems that do more than passively consume data—they can actively shape their own data streams, optimize information gathering in complex environments, and close the loop between representation and interaction. The rest of the paper is organized as follows:
\begin{itemize}
    \item \textbf{Chapter 2} introduces a toy example to illustrate the statistical principles of data collection for robot learning, which motivates controlled diffusion and ergodic control.
    \item \textbf{Chapter 3} gives an overview of diffusion learning, controlled diffusion, and ergodic control.
    \item \textbf{Chapter 4} presents numerical methods and implementation recipes for solving ergodic control problems.
    \item \textbf{Chapter 5} presents extending the standard ergodic control problem for different practical requirements.
    \item \textbf{Chapter 6} surveys ergodic control as a motion synthesis strategy across different robotics applications.
    \item \textbf{Chapter 7} discusses controlled diffusion as a strategy for closed-loop data collection in robot perception learning.
    \item \textbf{Chapter 8} discusses controlled diffusion as a strategy for data generation in embodied reinforcement learning.
    \item \textbf{Chapter 9} outlines open challenges and future directions in controlled diffusion and ergodic control.
\end{itemize}

\chapter{Data Collection for Robot Learning: A Toy Example}

Imagine a simple world with only axis-aligned rectangles. Each rectangle can be described by four parameters: the x- and y-coordinates of its center, and its width and height. However, these parameters are not directly measurable. Instead, the only way to interact with this world is by querying any location in space and observing a binary measurement: the sensor returns \emph{positive} if the point lies inside a rectangle and \emph{negative} otherwise. This setup is identical to the classic ``rectangle game'' example in computational learning theory \citep{kearns_introduction_1994}. The goal is to infer the hidden ground-truth rectangle from binary measurements. We will notice that “learning’’ a rectangle model from data in this case is remarkably easy with the following supervision method: given a hypothetical rectangle, the hypothesis aligns with the data if all the positive measurements fall within the rectangle and all the negative measurements fall outside, and vice versa.

\begin{figure}[t]
    \centering
    \includegraphics[width=1.0\linewidth]{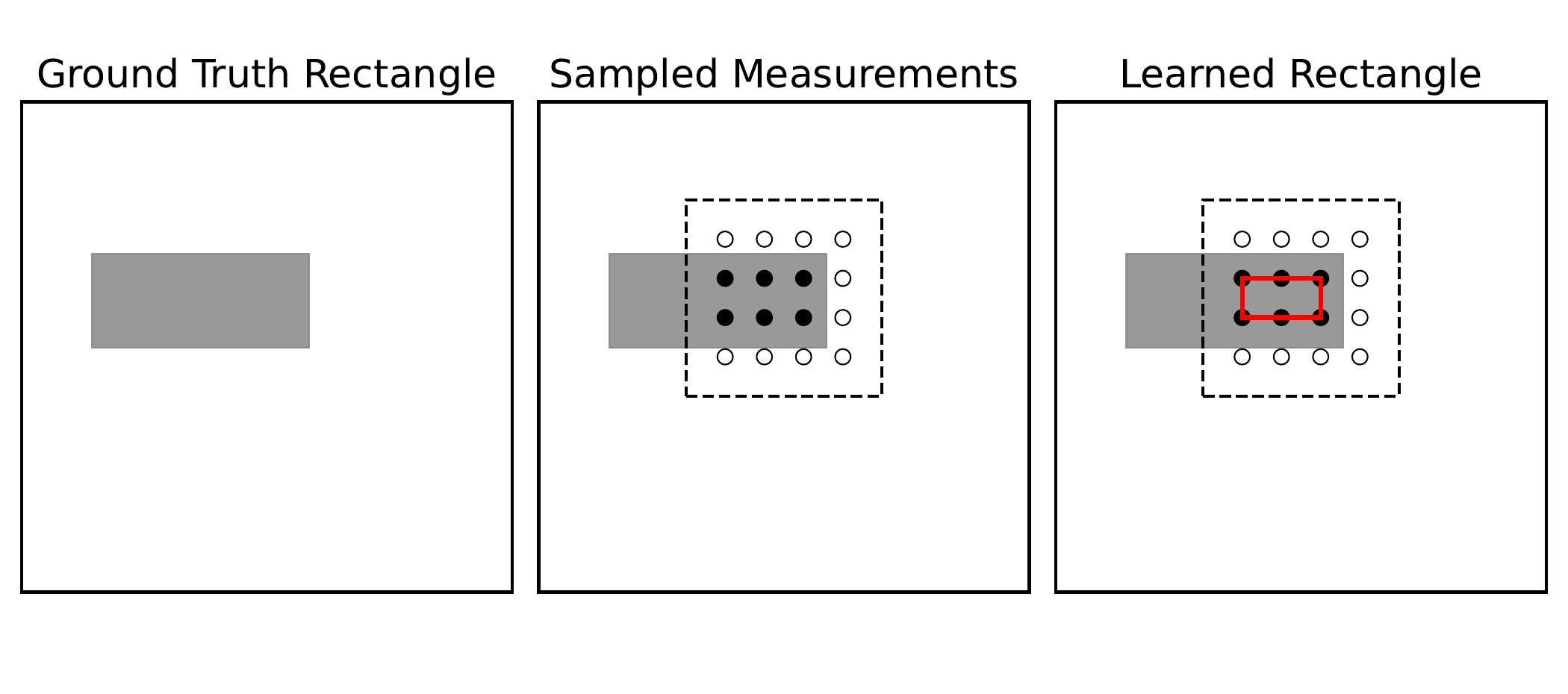}
    \vspace{-3.5em}
    \caption{Illustration of the ``rectangle game''. We assume the binary measurements are sampled from an evenly distributed grid within an arbitrary but fixed squared area. The learned rectangle is computed as the smallest rectangle that contains all the positive signals and none of the negative signals.}
    \label{fig:pac_illustration_1}
    
    \centering
    \includegraphics[width=1.0\linewidth]{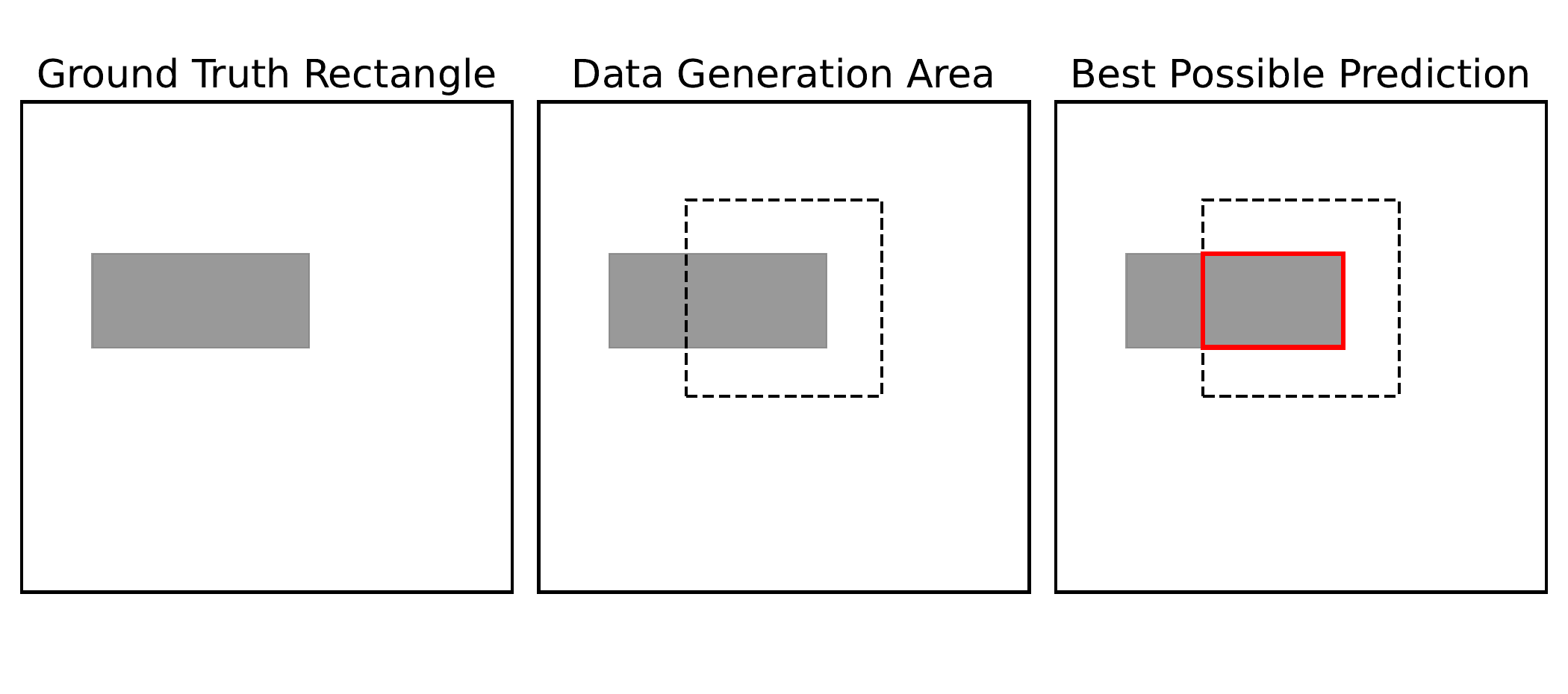}
    \vspace{-3.5em}
    \caption{Given a fixed squared data collection area, there exists a best possible rectangle prediction this learning algorithm can produce.}
    \label{fig:pac_illustration_2}
    \vspace{-1em}
\end{figure}

The purpose of creating this simple ``learning problem'' is twofold. First, most of the machine learning literature focuses on training and inference, emphasizing model architectures that extract patterns from datasets. In contrast, this example is designed so that neither training nor inference is the bottleneck. This raises a natural question: if perfect training and inference are available, is learning solved? Clearly not. The existence of such perfect methods highlights a deeper limitation: what can be learned is fundamentally constrained by the data we collect and how we collect it. Second, nothing about the rectangle environment restricts the insights from this toy example to only this environment. ``Rectangle'' here is really just a simple instance of a parameterized function representation. We can replace the ``rectangle'' with a ``triangle'' or a “circle,’’ or use a Gaussian process or a neural network as the learned representation, and everything we have said about the environment---including how to assess whether a hypothesis is correct---still holds. Therefore, this toy example also offers intuitive yet highly generalizable insights into the data collection problem in robot learning.

\section{What Data Makes Learning Correct?}

The first question we will ask is, what kind of data collection strategy will lead to ``correct'' learning. We will choose a simple learning algorithm to illustrate the example here: given a set of measurements (assuming at least two positives), we construct the smallest axis-aligned rectangle that contains all the positive points and excludes all negatives (Figure \ref{fig:pac_illustration_1}). This allows us to define the correctness of the learning process the difference bewteen the learned rectangle and the ground-truth rectangle. 

Suppose measurements are taken on a uniform grid over a squared region $\mathcal{S}$. Let the total number of measurements be $N=n^2$, corresponding to an $n\times n$ grid (Figure~\ref{fig:pac_illustration_1}). This grid-based sampling strategy provides two important insights:

\noindent\textbf{(A ``best possible'' rectangle)} Even with infinite resolution, the knowledge of the learned model is restricted to the boundaries of $\mathcal{S}$; thus the best we could ever hope to recover is the intersection of the ground-truth rectangle with $\mathcal{S}$ (see Figure~\ref{fig:pac_illustration_2}). 

\begin{figure}[t]
\centering
    \includegraphics[width=1.0\linewidth]{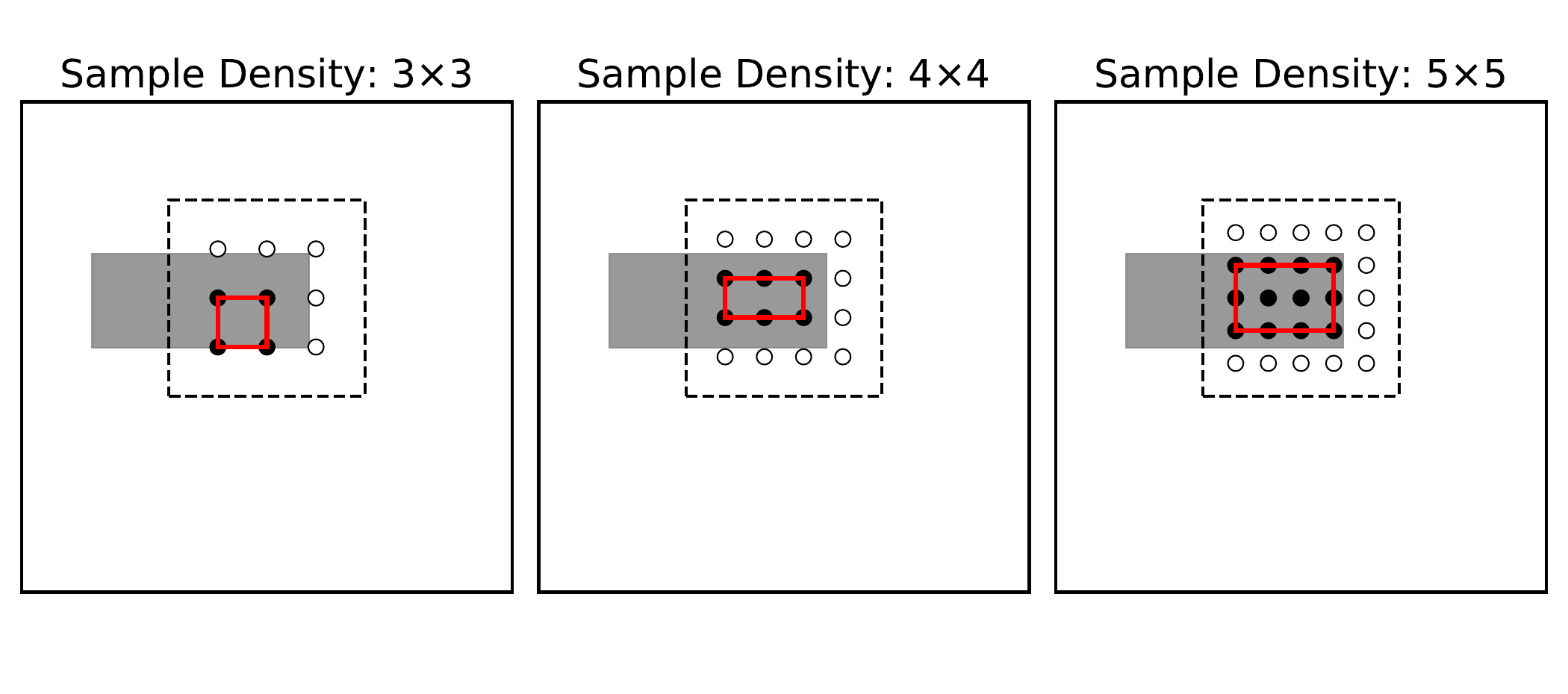}
    \vspace{-3.5em}
    \caption{There exists a computable lower bound of the error between the ground truth rectangle and the learned rectangle from a finite number of samples. Furthermore, this bound decreases as the number of samples increases.}
    \label{fig:pac_illustration_3}
    
    \centering
    \includegraphics[width=1.0\linewidth]{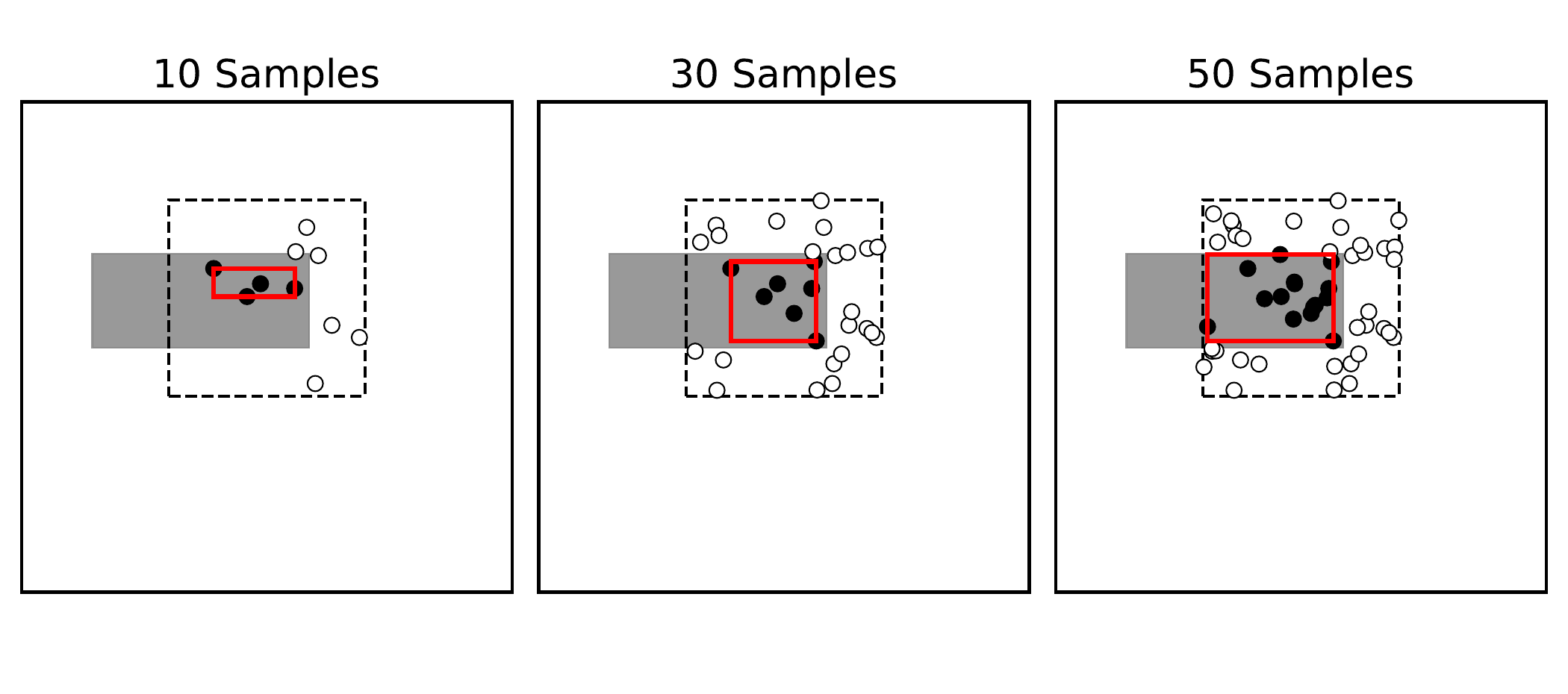}
    \vspace{-3.5em}
    \caption{The computable bound on the worst possible ``error'' of the model learned from a finite number samples exists for other, more ``random'', sampling strategies, too.}
    \label{fig:pac_illustration_4}
    \vspace{-1em}
\end{figure}

\noindent\textbf{(Bounded finite-sample error)} Because the samples form a uniform grid, the maximum ``error'' between the learned rectangle and this best-possible rectangle is bounded by the grid spacing. Intuitively, each side of the learned rectangle can be at most one grid cell away from the the true boundary, meaning the error shrinks proportionally to $\frac{1}{n}$ as sample density increases. This gives us a \emph{computable} worst-case error bound that decreases as the number of samples increases (see Figure~\ref{fig:pac_illustration_3}). 

This simple learning problem combined with a simple data collection strategy captures the essence of \emph{probably approximately correct} (PAC) learning. For any fixed hypothesis class (axis-aligned rectangles) and data distribution (uniform grid over $\mathcal{S}$), we can bound how far the learned model is from the best possible model in that class. In more general settings, instead of a uniform grid, we assume samples are drawn independently from an arbitrary but fixed data distribution (see Figure~\ref{fig:pac_illustration_4}), and instead of hand-derived geometric bounds, we use concentration inequalities to guarantee correctness.

\begin{definition}[PAC learning bound]
With probability at least $1-\delta$, the true error of the hypothesis $h$ returned by a learning algorithm is within $\epsilon$ of the smallest achievable error in the hypothesis class:
\[
    \Pr\big(\mathrm{err}(h) \leq \mathrm{err}(h^*) + \epsilon\big) \geq 1 - \delta,
\]
where $h^*$ is the best hypothesis in the class. 
\end{definition}

PAC learning is one of many tools for formally analyzing the properties of a learning algorithm. Our purpose here is not to provide a comprehensive treatment of PAC learning, but rather to use it as a lens to highlight the critical role of data collection in determining the performance guarantees of a learning algorithm. More importantly, this discussion offers insight into what principles a data collection strategy should satisfy and what implications these principles carry if we \emph{automate and optimize} data collection with a robot.

Returning to our toy example, one implicit but crucial design choice was to assume a \emph{stationary data distribution}. This assumption guarantees the existence of a “best possible model” that can be learned, as the target distribution remains fixed over time. The second design choice was to use a \emph{uniform grid} for sampling, a deterministic analogue of IID sampling. This choice enabled computable, finite-sample error bounds derived from the geometry of the grid. Together, these two principles—(1) a stationary target distribution and (2) a decorrelated, uniformly covering sampling strategy—form a foundation for designing effective data collection strategies.

Now, consider automating this process with a robot. Unlike static sampling, a robot must \emph{physically move} through space to take measurements. The assumptions of stationarity and IID sampling, while natural in abstract learning theory, become problematic when translated into robot control synthesis. Instead of reasoning about temporal actions alone, we must explicitly reason about the \emph{spatial distribution} that the robot’s trajectory induces over time, ensuring that (i) this induced distribution converges to a desired target distribution and (ii) successive samples along the trajectory are sufficiently decorrelated.

This shift in perspective—from temporal decision-making to spatial distribution design—naturally motivates the formalism of \emph{ergodic control}. At the foundation of ergodic theory is Birkhoff’s Ergodic Theorem, which establishes that under certain conditions (irreducibility and aperiodicity), the time average of a trajectory converges to the spatial average defined by a stationary distribution. We begin by recalling the formal definition of an ergodic Markov decision process (MDP):

\begin{definition}[Ergodic MDP]
A Markov decision process is said to be \emph{ergodic} if it is both irreducible and aperiodic. In this case, for any stationary policy, the induced Markov chain admits a unique stationary distribution $\pi$, and for any measurable function $f$, the time-average along a trajectory converges to the spatial expectation:
\[
\lim_{T \to \infty} \frac{1}{T} \sum_{t=1}^T f(x_t) = \int f(x) \pi(x) dx.
\]
\end{definition}


\begin{figure}[t]
    \centering
    \includegraphics[width=1.0\linewidth]{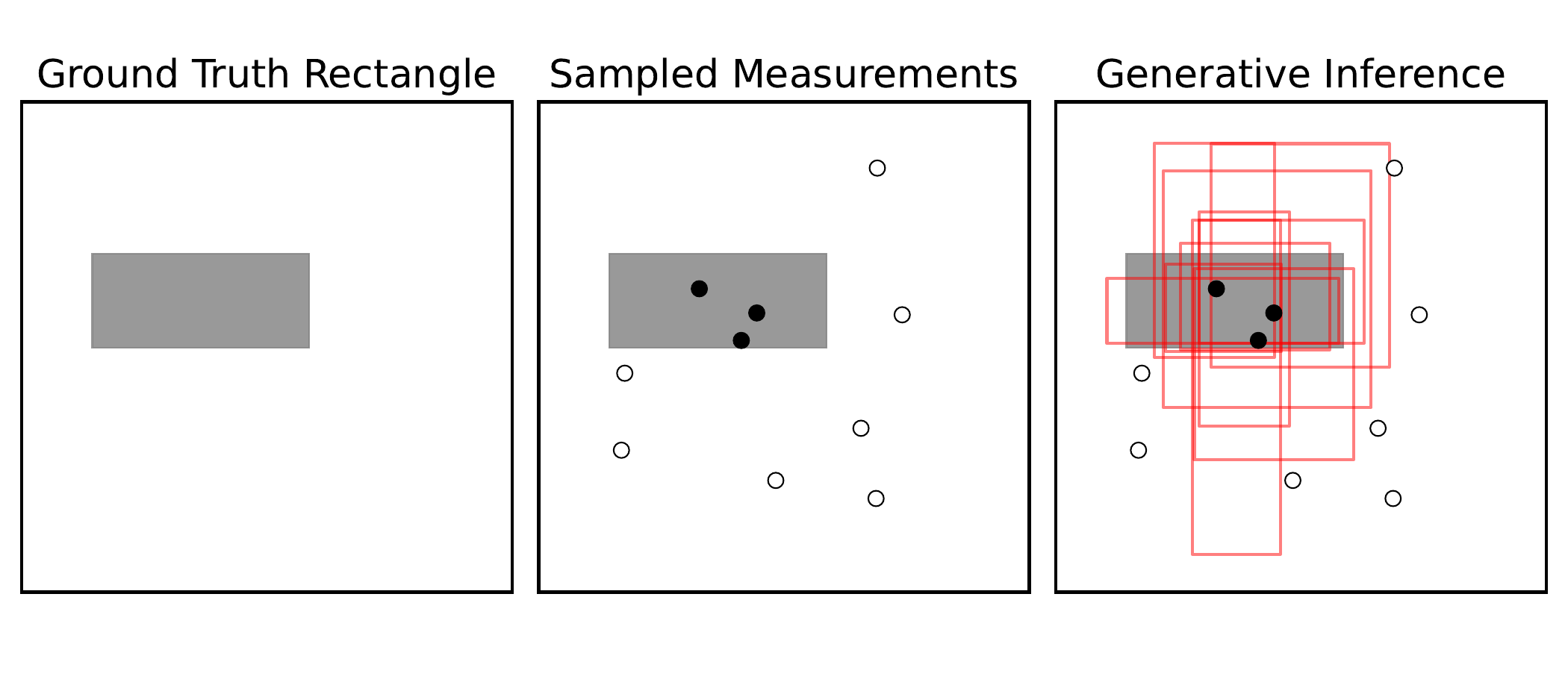}
    \vspace{-3em}
    \caption{Illustration of the BoxGPT example. Training and inference of the generative model are straightforward; what is not straightforward is data collection.}
    \label{fig:boxgpt_env}
\end{figure}

\section{What Data Makes Learning Efficient?}

In the above, we have seen that the importance of i.i.d. data collection with a stationary distribution. However, another important question that is unanswered above is the specification of the data distribution itself, what distribution leads to better efficiency of the learning process. 

To improve efficiency, we need to know what data distribution provides more information for the learned model. This means we will need a little help from the learning process. While the learning method---fitting the smallest rectangle that contains all positive signals and none of the negative signals---works, it is limited in terms of informing us what is missing in the model. Instead, we will generalize this learning method a little bit, to make a tiny ``generative model'' to capture a posterior distribution instead of solely a deterministic estimation: think of maximum likelihood estimation (MLE) and Bayesian inference. 

Suppose we have already taken a number of binary measurements, with these measurements as the training data, we want to create a ``generative model'' that can predict the underlying rectangles. The learning procedure is simple: a generated rectangle sample is valid if all positive signals fall inside it and all negative signals fall outside. This simple rule, combined with rejection sampling, allows us to construct a globally optimal generative model that asymptotically recovers the ground-truth posterior distribution (see Figure~\ref{fig:boxgpt_env}). Let's call this model ``BoxGPT.''

\begin{figure}[t]
    \centering
    \includegraphics[width=1.0\linewidth]{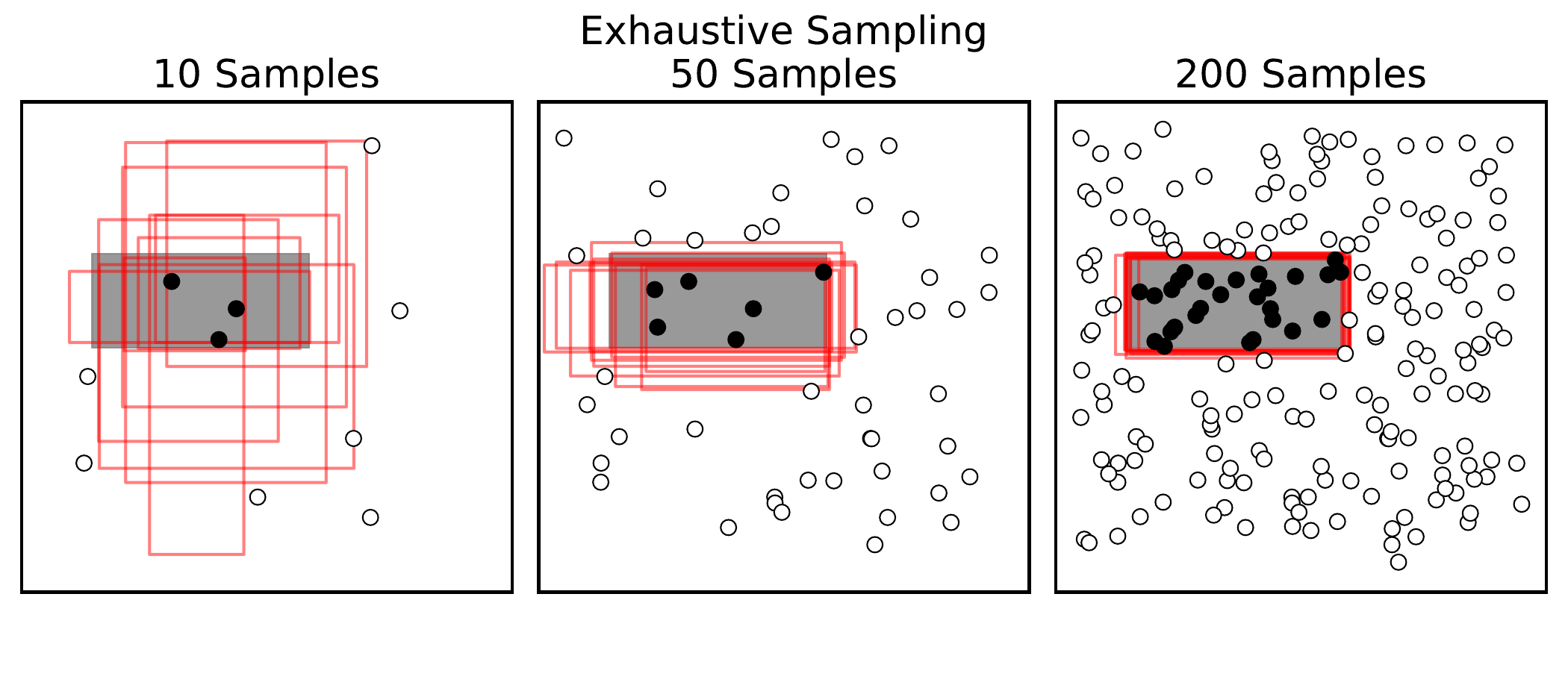}
    \vspace{-3em}
    \caption{Exhaustive sampling is an effective but inefficient data collection strategy.}
    \label{fig:exhaustive_sampling}
\end{figure}

One strategy to create a ``good dataset'' for BoxGPT is to exhaustively sample measurements from the entire space. Over time, the predicted rectangle samples will converge to the ground-truth rectangle (see Figure~\ref{fig:exhaustive_sampling} for examples). As shown in the previous section, this strategy is not only intuitive but also provably optimal under standard assumptions, such as measurements being independent and identically distributed (i.i.d.), with formal guarantees supported by PAC-learning theory. In fact, this exhaustive sampling principle underlies the success of many large foundation models: given a capable training pipeline and inference model, their generality is closely tied to the ``uniformity'' of the training data.

However, exhaustive sampling is not efficient. Most data points are redundant and contribute little to improving the model. Increasing the sampling rate does not necessarily yield better models, and in fact typically will not yield a better model without a change in data collection strategy; instead, it increases computational costs. Even in this simple BoxGPT example, validating a hypothetical rectangle scales as $\mathcal{O}(N)$ for a dataset of size $N$. As a result, exhaustive sampling produces datasets that may not improve performance while significantly slowing both training and inference. This cost is even more pressing in real-world foundational models, where the energy demands of large-scale training are a growing sustainability concern~\citep{crawford_generative_2024, chen_generative_2025}. Moreover, exhaustive sampling complicates adaptation: if the underlying data distribution changes significantly (e.g., the rectangle shifts), the entire dataset must be re-collected from scratch.

\begin{figure}[t]
    \centering
    \includegraphics[width=1.0\linewidth]{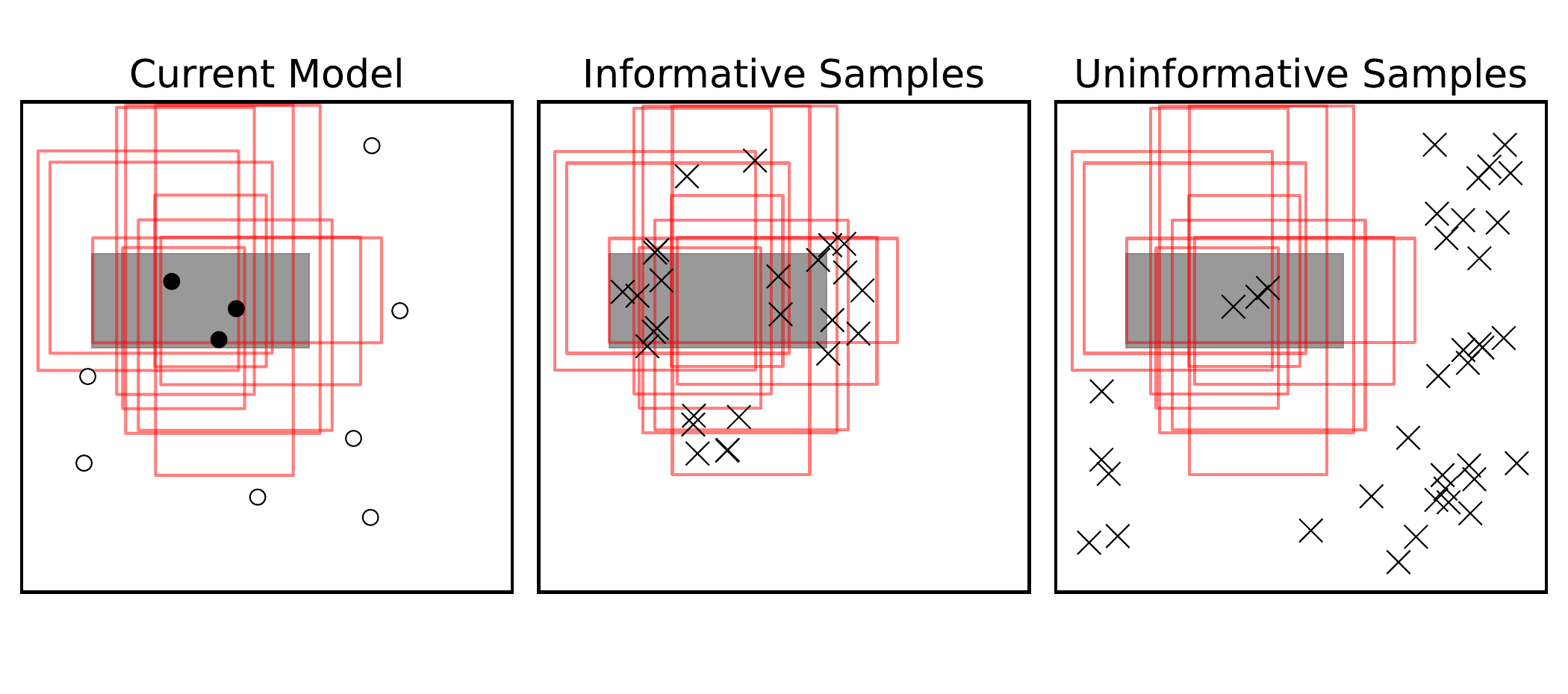}
    \vspace{-3em}
    \caption{Based on the model’s current predictions, sampling measurements from different regions---regardless of their exact values---has varying impacts on model improvement. Identifying the optimal data distribution to sample from is the key to optimal data collection for learning.}
    \label{fig:high_low_information}
    \vspace{1em}
    \centering
    \includegraphics[width=1.0\linewidth]{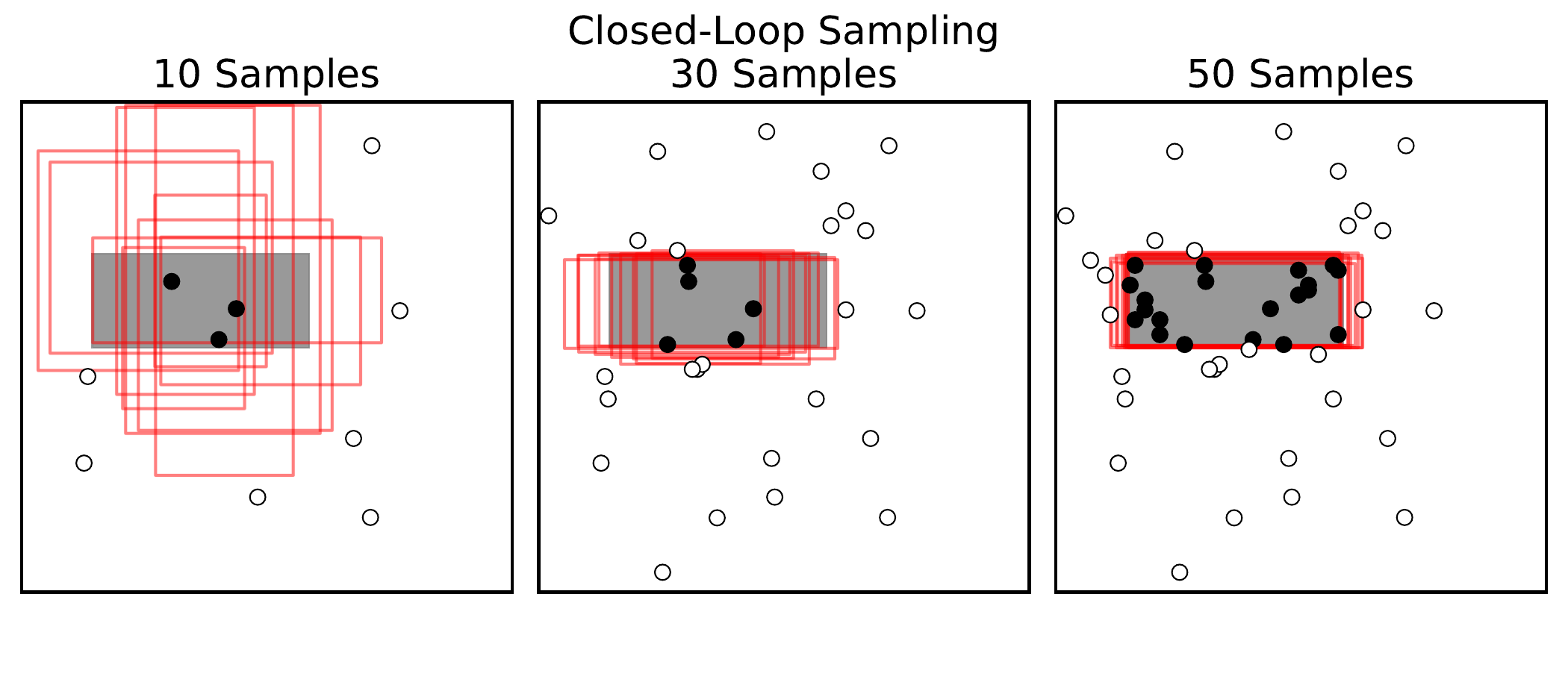}
    \vspace{-3em}
    \caption{Closed-loop data collection can significantly improve data collection efficiency while preserving asymptotic optimality.}
    \label{fig:closed_loop_generation}
\end{figure}

\emph{How can we improve the efficiency of data collection while preserving the same asymptotic optimality as exhaustive sampling?}  

Exhaustive sampling is fundamentally tied to \emph{offline training}, where model training and inference are separate phases. Because data must be collected without knowledge of the trained model, measurements are taken over the largest plausible domain, with a high sampling rate to avoid ambiguity. With strong prior knowledge of the task, this is optimal for offline training, but it enforces heavy data requirements.

To address this limitation, we shift focus to \emph{closed-loop data collection}. Rather than fixing a dataset before training, we iteratively expand the dataset during training, guided by the model’s current predictions. Generative models define probabilistic distributions, making prediction uncertainty a natural measure for guiding new measurements. In BoxGPT, for example, sampling from regions that are consistently predicted as inside or outside all rectangles adds little value. Conversely, regions where predictions are split—half inside, half outside—correspond to high entropy and are the most informative (see Figure~\ref{fig:high_low_information}).  

A simple closed-loop strategy emerges: start with a small, uniformly sampled dataset from the largest plausible domain, then iteratively sample new measurements from the information distribution informed by the generative model based on the existing data (see Figure~\ref{fig:closed_loop_generation}). This approach preserves the i.i.d. property that is required for the correctness of the learning process, but it creates a new decision-making problem for dynamically updating the distribution from which samples are collected, progressively improving model quality while retaining the same asymptotic guarantees as exhaustive sampling, but at far lower data and computational cost.

\begin{figure}[t]
    \centering
    \includegraphics[width=1.0\linewidth]{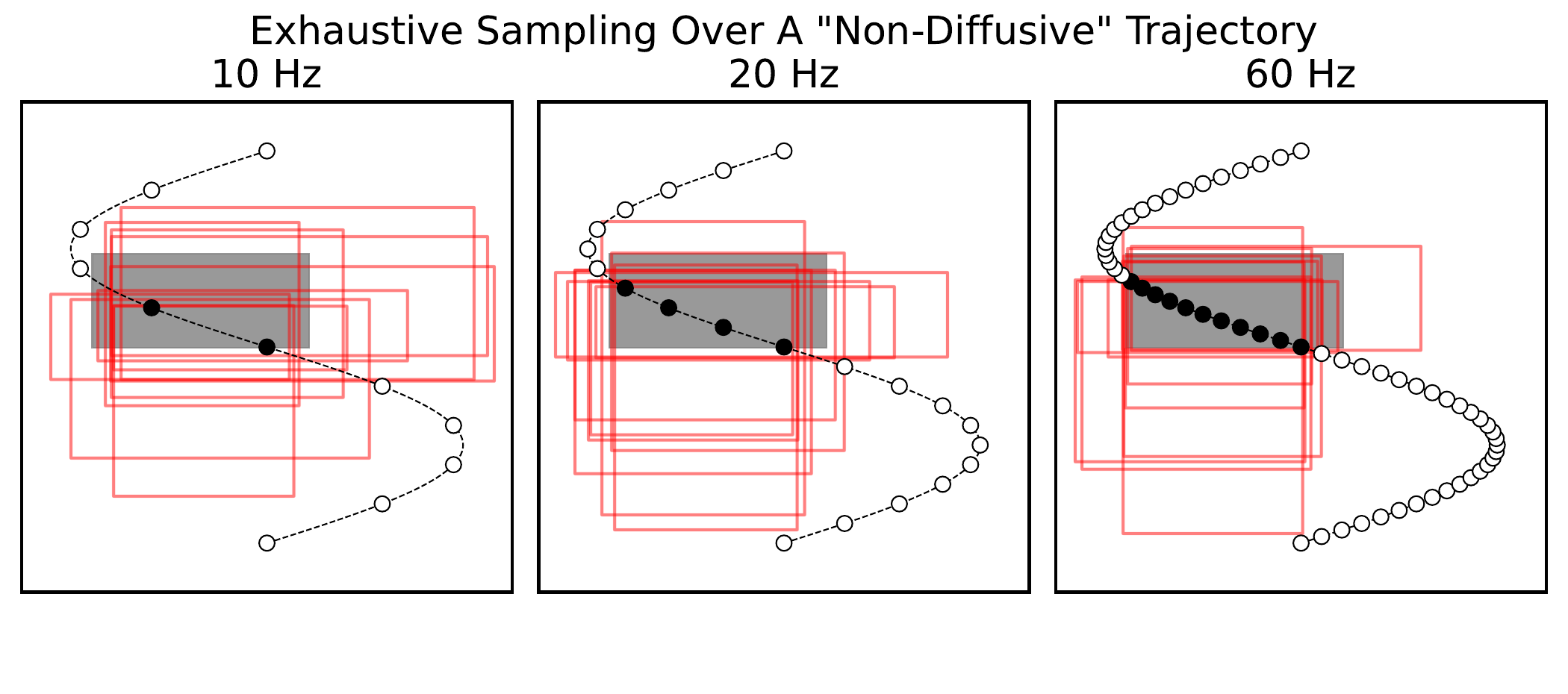}
    \vspace{-3em}
    \caption{Increasing the sampling frequency along a trajectory where the time-averaged statistics fail to converge to a stationary distribution may not yield a better dataset.}
    \label{fig:exhaustive_traj_sampling}
    \vspace{+1em}
    \centering
    \includegraphics[width=1.0\linewidth]{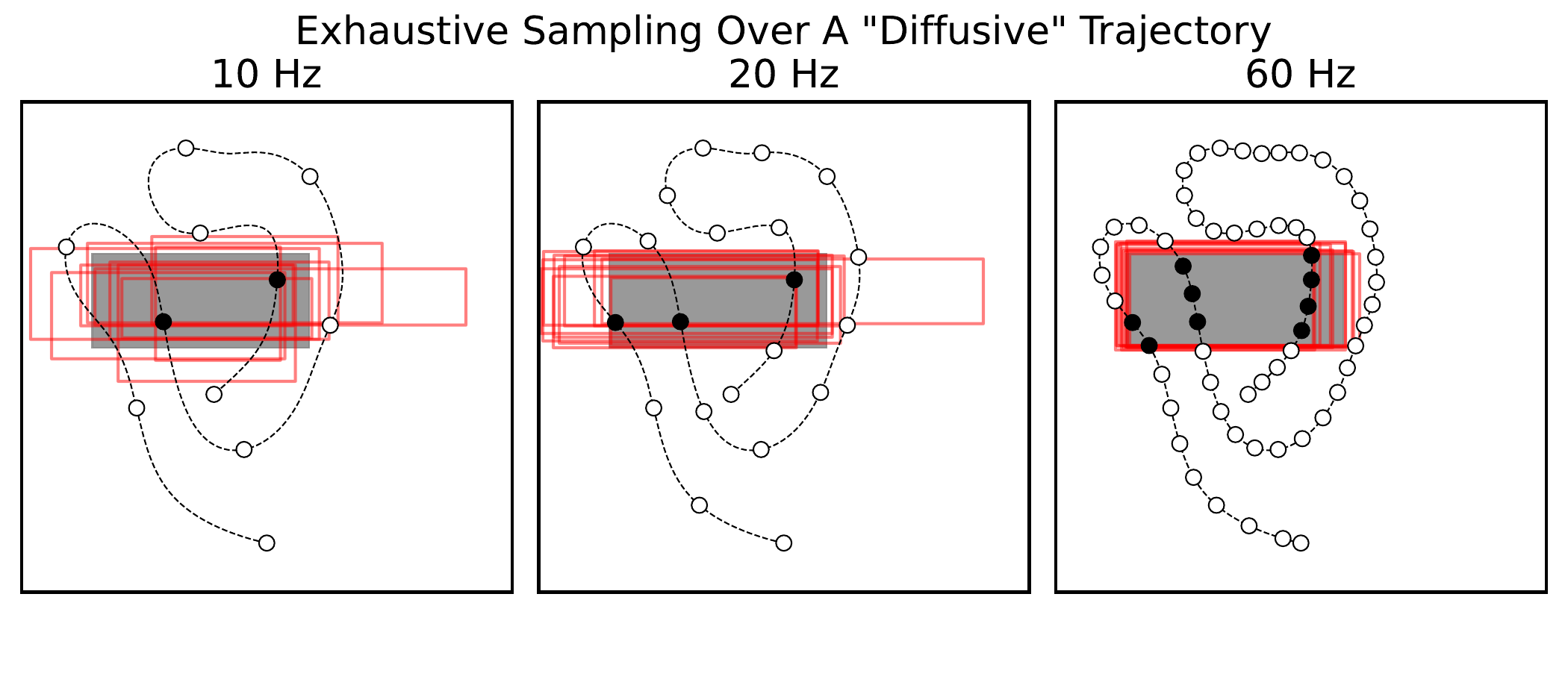}
    \vspace{-3em}
    \caption{Controlled diffusion generates trajectories based on desired time-averaged statistics, in which case increasing the sampling rate is effective for producing a better dataset.}
    \label{fig:ergodic_traj_sampling}
    \vspace{-1em}
\end{figure}

\section{From Sampling to Motion Synthesis}

Now consider closed-loop data collection for robots. While sampling was trivial in the BoxGPT example, it is far more complex for embodied agents constrained by the continuity and dynamics of physical environments. On a robot, data is produced through motion: new information arises only when the robot expends energy to interact with the environment. As a result, the data collection problem becomes a motion synthesis problem, and “sampling” new measurements based on model uncertainty is no longer straightforward.

Even exhaustive sampling is not trivial here. A common intuition is that increasing the sampling frequency along a trajectory will yield a better dataset. However, this intuition only holds when samples can be treated as i.i.d. draws from a stationary distribution. In that case, denser sampling produces a clear “filling in” effect: gaps between data points shrink, the coverage of the domain improves, and model errors can be statistically bounded (as shown in Figure~\ref{fig:pac_illustration_3} and Figure~\ref{fig:pac_illustration_4}). By contrast, when data are sampled from a trajectory, the governing dynamics introduces strong temporal correlations between successive states. These correlations mean that the trajectory, when viewed as a collection of samples, no longer behaves like draws from a stationary distribution. The result is that both the stationary distribution property and the i.i.d. assumption are lost, unless the trajectory is deliberately synthesized with these properties in mind.

This distinction is crucial. In exhaustive sampling, increasing the sample rate directly improves the dataset. For trajectories of embodied agents, increasing the sample rate along the same path does not eliminate correlations or restore the stationary distribution. It simply produces more redundant points that fail to close the statistical gaps. As a result, higher frequency alone cannot fix the problem: without motion synthesis that enforces stationary distribution and i.i.d. properties, the dataset remains fundamentally limited (see Figure~\ref{fig:exhaustive_traj_sampling}).

Therefore, for data collection in robot learning, what we need is a motion synthesis strategy that optimizes the time-averaged spatial statistics of the trajectory. The objective is to destroy correlations between successive states and to spread out the trajectory so that the states behave as if they were i.i.d. samples from a stationary distribution. This idea is closely related to the diffusion process, where particles evolve over time to spread out and, at equilibrium, assemble a prescribed spatial distribution. The generated trajectory in this case, although constrained by the dynamics of the system, can be constructed to exhibit the same statistical property as i.i.d. samples from a stationary distribution. This property is called ergodicity: an ergodic system ensures that long-run averages along the trajectory are statistically equivalent to averages over independent samples drawn from the distribution.

Controlled diffusion builds on this principle: it is a control synthesis framework that enforces desired time-averaged spatial statistics under dynamics constraints, and the technique for solving such problems is known as ergodic control. Controlled diffusion therefore directly enforces the two principles of robot data collection discussed above: the stationary distribution and the i.i.d. property. As a result, increasing the sampling frequency alongside such ``diffusive'' trajectories yields a better dataset, because the additional samples improve coverage in the same way as increasing the number of i.i.d. samples would (see Figure~\ref{fig:ergodic_traj_sampling}).

\begin{figure}[t]
    \centering
    \includegraphics[width=0.9\linewidth]{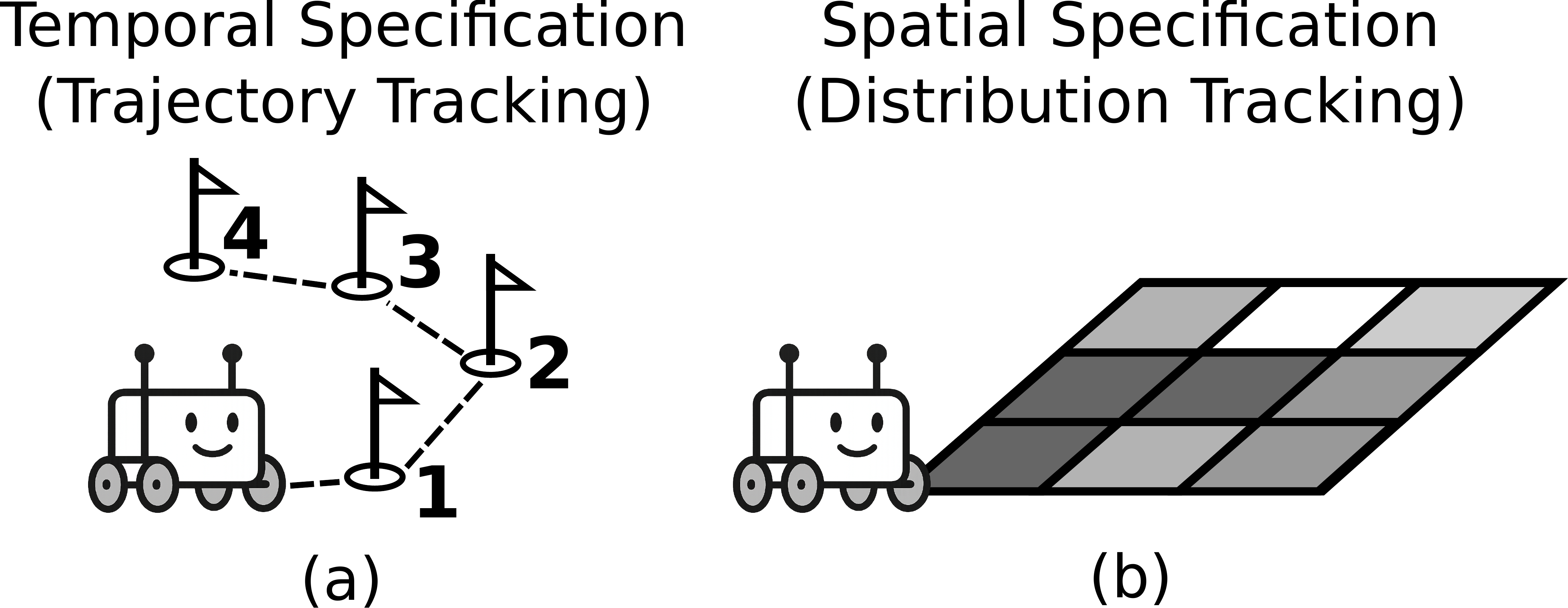}
    \caption{Temporal vs. spatial specifications for motion synthesis. (a) Conventional methods focus on the temporal sequence of robot actions. (b) Controlled diffusion focuses on the spatial statistics of the robot trajectory.}
    \label{fig:trajectory_vs_distribution_tracking}
    \vspace{+1em}
    \centering
    \includegraphics[width=1.0\linewidth]{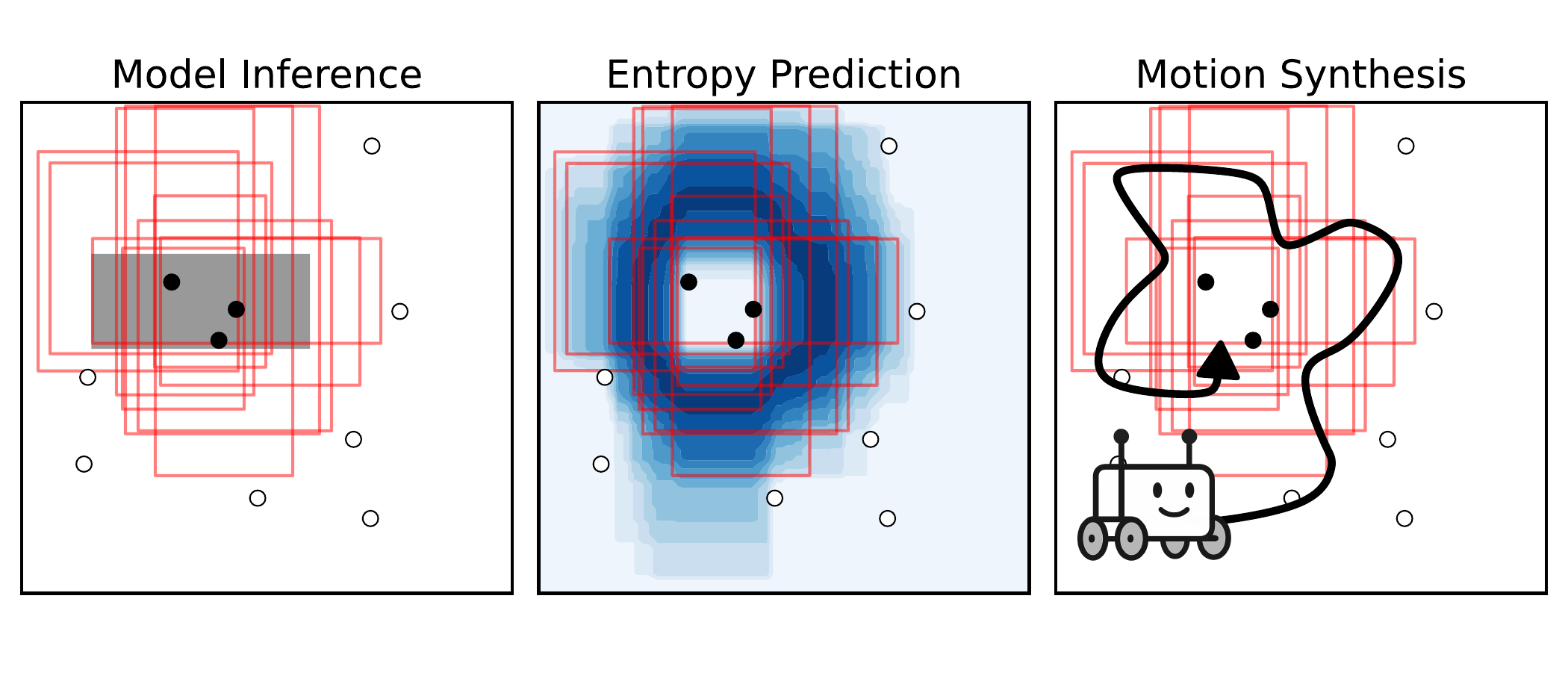}
    \vspace{-3em}
    \caption{Automating closed-loop data collection on robots requires a spatial-statistics-based specification for motion synthesis, which leads to the formulation of controlled diffusion and ergodic control.}
    \label{fig:robot_data_generation}
\end{figure}

The BoxGPT example illustrates three levels of thinking about data collection for robot learning. First, the statistical properties of the collected data, namely the requirements of a stationary distribution and the i.i.d. assumption. Second, the necessity of closed-loop data collection. Third, for robots, these two principles demand an entirely new perspective on motion synthesis: motion must be synthesized based on the time-averaged spatial statistics rather than the temporal sequence of states (see Figure~\ref{fig:trajectory_vs_distribution_tracking}). This perspective leads to the framework of controlled diffusion, a control synthesis framework for shaping the statistical properties of motion in embodied agents, and to ergodic control, the family of techniques that solve the controlled diffusion problem.

Taken together, these ideas provide an outline for closed-loop robot data collection: given the current model inference results from existing data, we first generate a spatial distribution that encodes the desired statistics of future data (e.g., entropy prediction). We then synthesize robot motion to produce trajectories whose time-averaged statistics match the target distribution, thereby generating data that satisfies both the stationary distribution and i.i.d. principles (see Figure~\ref{fig:robot_data_generation}). 

Lastly, we provide an interactive demonstration of BoxGPT at \url{https://murpheylab.github.io/boxgpt/}. Users can control a virtual robot to collect data manually or enable an ergodic controller for automated data collection. The project page also provides the single-file Gymnasium environment for prototyping other data collection strategies, example policies in Google Colaboratory notebooks, and a detailed mathematical formulation.

\chapter{Controlled Diffusion and Ergodic Control: Overview}

\section{Diffusion Learning}

\subsection{Diffusion-based generative models}

The core objective of generative models is to capture the probability distribution underlying a dataset and to generate statistically consistent samples from it. The central challenge lies in balancing model expressiveness with computational tractability. Simple models—most notably the Gaussian distribution, and more generally variational autoencoders (VAEs)~\citep{kingma_auto-encoding_2013}—are easy to train and sample from, but limited in their fidelity when representing high-dimensional, complex data. By contrast, highly expressive models such as generative adversarial networks (GANs)~\citep{goodfellow_generative_2014} can approximate rich data distributions, but at the cost of heavy computational demands in both training and inference, which constrains their practicality.

Diffusion-based models provide a way to resolve this tension. They significantly improve fidelity and expressiveness without incurring prohibitively high costs. The key distinction of diffusion learning, compared to encoder–decoder mappings or energy-based formulations, is its reliance on an iterative process of noising and denoising. Starting from a simple, tractable distribution (e.g., Gaussian), the model progressively learns to transform it toward the complex target distribution. This iterative framework not only enables the capture of intricate structures in data but also, through later refinements, reduces inference cost.

The foundation of diffusion learning was established in the seminal work on deep unsupervised learning with nonequilibrium thermodynamics~\citep{sohl-dickstein_deep_2015}. In this formulation, the forward process incrementally corrupts the data distribution with Gaussian perturbations until it approaches a pure Gaussian. A neural network is then trained to approximate the reverse denoising process by modeling conditional Gaussian transitions. Training is performed by minimizing a variational lower bound on the KL divergence between the learned reverse process and the true reverse dynamics. This framework formalizes diffusion as a thermodynamics-inspired process, giving rise to the first generation of diffusion learning. Despite improving fidelity over VAEs and being more stable than GANs, its reverse process remained Gaussian-approximated and computationally heavy, while performance was highly sensitive to heuristically chosen noise schedules.

To overcome these limitations, denoising diffusion probabilistic models (DDPMs) were introduced~\citep{ho_denoising_2020}. DDPMs replace the variational objective with a simple mean-squared error loss, training a neural network to directly predict the injected noise. This reparameterization improves computational efficiency while preserving theoretical guarantees. The framework also provides principled guidance on noise scheduling and leverages U-Nets with attention for the denoising step. Compared with the original diffusion approach, DDPMs deliver substantial gains in both fidelity—on par with GANs for image generation—and efficiency, establishing diffusion models as a mainstream generative modeling paradigm.

Parallel to DDPMs, a complementary view emerged from nonequilibrium thermodynamics: score-based models~\citep{song_score-based_2021}. The score function, defined as the gradient of the log density, naturally arises in Langevin dynamics as a mechanism for sampling. Score-based generative modeling learns to approximate this function directly via score matching. Training is improved by perturbing data with noise and conditioning the score network on the noise level, leading to noise-conditioned score networks. This construction is mathematically equivalent to DDPM’s noise-prediction objective under reparameterization, revealing that both approaches are different perspectives on the same principle.

A third perspective formulates generative modeling in terms of continuous flows. Continuous normalizing flows (CNFs)~\citep{chen_neural_2018} define a generative model as a time-dependent vector field that transforms samples from a simple prior to the data distribution. The vector field, parameterized by a neural network, is trained via maximum likelihood using the change-of-variables formula. Subsequent work introduced flow matching~\citep{lipman_flow_2022}, which simplifies training by directly fitting the neural vector field with a mean-squared error loss. This flow-based view expands the flexibility of diffusion models: it accommodates conventional noise-driven transformations while also enabling optimal transport–based flows that improve efficiency and sample quality. These extensions have pushed flow-based diffusion methods to state-of-the-art performance in recent years, cementing their role as a unifying framework within generative modeling.

\subsection{Diffusion models in robot learning}

The success of diffusion learning has had a significant impact on robotics. The key reason is that diffusion models provide the much-needed fidelity for learning perception and decision-making of high complexity while maintaining affordable computational cost. A representative milestone is the use of DDPMs as policy representations for imitation learning~\citep{chi_diffusion_2024}, enabling the direct learning of highly capable visuomotor policies from human demonstrations. Since then, the role of diffusion models in robot learning has expanded drastically, encompassing new model variants such as flow-based approaches~\citep{yang_pointflow_2019} and applications in reinforcement learning~\citep{wang_diffusion_2022, mcallister_flow_2025}.

More importantly, improved model fidelity enables scaling laws: model capability grows substantially with larger datasets. This has fueled the trend of generalist robot policies—policies that predict actions directly from sensor inputs across diverse tasks, trained on massive collections of pre-collected data. With inference fidelity no longer the primary bottleneck, data collection itself has emerged as the next challenge. Collecting robot data is costly due to the physical nature of robots, and data quality has been shown to critically influence performance. Yet an open question remains: how can we quantify data quality and automate or optimize the data collection process for robot learning?

The parallel rise of diffusion learning and the rising importance of robot data collection make it timely to introduce controlled diffusion. Controlled diffusion adopts the same principle of transforming a simple distribution into a complex one, but with a different purpose: not to learn patterns directly from datasets, but to generate robot actions in response to uncertainty, and to design automated, optimal strategies for data collection in robot learning.

\section{Controlled Diffusion, Ergodicity, and Ergodic Control}

\subsection{Preliminaries}

We denote the state space of the robot as $\mathcal{S}$ and the control space as $\mathcal{U}$. 
The robot's state trajectory is denoted as a function of time $s:[0,T]\mapsto\mathcal{S}$, with $T$ being the planning horizon. 
Similarly, the robot's control trajectory is denoted as $u:[0,T]\mapsto\mathcal{U}$. 
The state evolution of the robot is subject to the dynamics
\begin{align}
    \dot{s}(t)= f(s(t), u(t)), 
\end{align}
with initial state $s_0$ given a priori. 
The task space is denoted as $\mathcal{X}$, and we assume there exists a differentiable mapping $h:\mathcal{S}\mapsto\mathcal{X}$ from the robot state space to the task space. 
We denote the space of probability distributions over $\mathcal{X}$ as $\mathcal{P}(\mathcal{X})$, and the space of continuous functions over $\mathcal{X}$ as $\mathcal{C}(\mathcal{X})$.

\subsection{Controlled diffusion}

The core idea in controlled diffusion is that the trajectory of a robot can be interpreted as samples from a probability distribution over $\mathcal{X}$. 
This distribution can be represented as an empirical measure based on the Dirac delta function.

\begin{definition}[Dirac delta function]
    The Dirac delta function, denoted $\delta(x)$, is a generalized function~\citep{lighthill_introduction_1958} defined by its action on test functions:
    \begin{align}
        \int_{\mathcal{X}} \delta(x-s)\, f(x)\, dx = f(s), \quad \forall s\in\mathcal{X}, \ \forall f\in\mathcal{C}(\mathcal{X}).
    \end{align}
\end{definition}

\begin{definition}[Trajectory empirical distribution]
    Given a state trajectory $s(t)$, the \emph{trajectory empirical distribution} is defined as the time-averaged probability measure with density
    \begin{align}
        p_{s}(t, x) = \frac{1}{t} \int_0^t \delta\big( x - h(s(\tau)) \big)\, d\tau.  \label{eq:emp_distr}
    \end{align}
\end{definition}

The temporal evolution of $p_s(t,x)$ is governed by the robot's dynamics and the applied controls. 
Thus, the trajectory empirical distribution can be viewed as a diffusion process that can be steered through control inputs. 
We refer to this process as \emph{controlled diffusion}. 
The objective is to regulate this process so that the trajectory empirical distribution converges to a target distribution.

\begin{definition}[Controlled diffusion]
    Given a target distribution $q(x)\in\mathcal{P}(\mathcal{X})$, controlled diffusion is the process of generating controls $u(t)$ such that
    \begin{align}
        p_{s}(t,\cdot) \xrightarrow{t\to\infty} q(\cdot), 
    \end{align}
    where the convergence is in the sense of weak convergence of measures, i.e.,
    \begin{align}
        \lim_{t\to\infty} \int_{\mathcal{X}} f(x)\, p_s(t,dx) 
        = \int_{\mathcal{X}} f(x)\, q(dx), \quad \forall f\in\mathcal{C}(\mathcal{X}),
    \end{align}
    subject to the dynamics
    \begin{align}
        s(t) = s_0 + \int_0^t f(s(\tau), u(\tau))\, d\tau.
    \end{align}
\end{definition}

\paragraph{Intuition.} 
Controlled diffusion means steering the robot so that the state visitation frequency of its trajectory matches a desired spatial distribution $q$. In other words, the robot is not simply driven to a particular goal state, but instead guided so that its long-run presence in task space is statistically indistinguishable from samples drawn from $q$. This definition coincides with the notion of \emph{ergodicity}: a trajectory is ergodic with respect to $q$ if the time averages along the trajectory equal the spatial averages under $q$.

\subsection{Ergodic systems and ergodic control}

In the classical setting, ergodicity is defined with respect to an invariant probability distribution. Let $k(x'|x)$ denote the transition kernel of the closed-loop system. A distribution $q\in\mathcal{P}(\mathcal{X})$ is said to be \emph{invariant}, or \emph{stationary}, if
\begin{align}
    q(x') = \int_{\mathcal{X}} k(x'|x)q(x),dx. \label{eq}
\end{align} Thus, if the current state is distributed according to $q$, its distribution remains $q$ after the system evolves. Ergodicity further requires that long-run averages along individual trajectories recover expectations under this invariant distribution. In ergodic control, $q$ is specified as a desired target distribution, and the control objective is to synthesize closed-loop behavior whose long-run visitation statistics converge to $q$.

\begin{definition}[Ergodic system]
    A dynamical system $s(t)$ is said to be \emph{ergodic} with respect to a target distribution $q(x)\in\mathcal{P}(\mathcal{X})$ if its trajectory empirical distribution $p_s^T(x)$ converges weakly to $q(x)$ as $T\rightarrow\infty$. 
    Equivalently,
    \begin{align}
        \lim_{T\rightarrow\infty} \mathbb{E}_{x\sim p_s^T}[\phi(x)] 
        = \mathbb{E}_{x\sim q}[\phi(x)], 
        \quad \forall \phi\in\mathcal{C}(\mathcal{X}). \label{eq:ergodic_system}
    \end{align}
\end{definition}

\begin{remark}
    In classical ergodic theory, ergodicity is defined for a measure-preserving dynamical system by requiring that every invariant measurable set has either zero or full measure. Under the assumptions of Birkhoff's ergodic theorem, this measure-theoretic definition implies the convergence of time averages to expectations under the invariant measure for almost every initial condition. We adopt this consequence as a working definition because it directly describes the trajectory-level property relevant to control and data collection.
\end{remark}

\begin{lemma}
    By the definition of the Dirac delta, the weak convergence in (\ref{eq:ergodic_system}) can also be written as
    \begin{align}
        \lim_{T\rightarrow\infty} \frac{1}{T} \int_0^T \phi\big(h(s(t))\big)\, dt 
        = \mathbb{E}_{x\sim q}[\phi(x)], 
        \quad \forall \phi\in\mathcal{C}(\mathcal{X}).
    \end{align}
\end{lemma}

\begin{theorem}[Ergodicity via empirical distribution] \label{thm:ergodicity}
    As an immediate consequence, a system is ergodic with respect to $q(x)$ if and only if
    \begin{align}
        p_s^T(x) \xrightarrow{T\to\infty} q(x) \quad \text{in distribution}.
    \end{align}
\end{theorem}

\begin{remark}
    The convergence above is weak convergence of probability measures and does not, in general, imply pointwise convergence of probability densities. Pointwise density convergence requires additional regularity or an explicit smoothing of the empirical measure.
\end{remark}

\begin{remark}
    Another related concept is \emph{mixing}, which describes how temporal dependence between states decays as the separation between them increases~\citep{mathew_multiscale_2005, scott_capturing_2009}. Ergodicity and mixing are related because mixing generally implies ergodicity, while ergodicity alone does not guarantee that temporally separated samples become approximately independent.
\end{remark}

The definition above enables us to \emph{quantify} ergodicity by measuring the statistical discrepancy between the empirical distribution $p_s^T(x)$ induced by a trajectory and the target distribution $q(x)$. 

\begin{definition}[Ergodic metric] \label{def:ergodic_metric}
    An \emph{ergodic metric} is a nonnegative functional
    \begin{align}
        D:\mathcal{P}(\mathcal{X})\times\mathcal{P}(\mathcal{X}) \to \mathbb{R}_{\geq 0},
    \end{align}
    that quantifies the discrepancy between the trajectory empirical distribution $p_s^T$ and the target distribution $q\in\mathcal{P}(\mathcal{X})$. 
    It satisfies the identity of indiscernibles:
    \begin{align}
        D(p,q) = 0 \quad \Leftrightarrow \quad p = q \ \text{as measures on }\mathcal{X}.
    \end{align}
\end{definition}

\begin{figure}[t]
    \centering
    \includegraphics[width=1.0\linewidth]{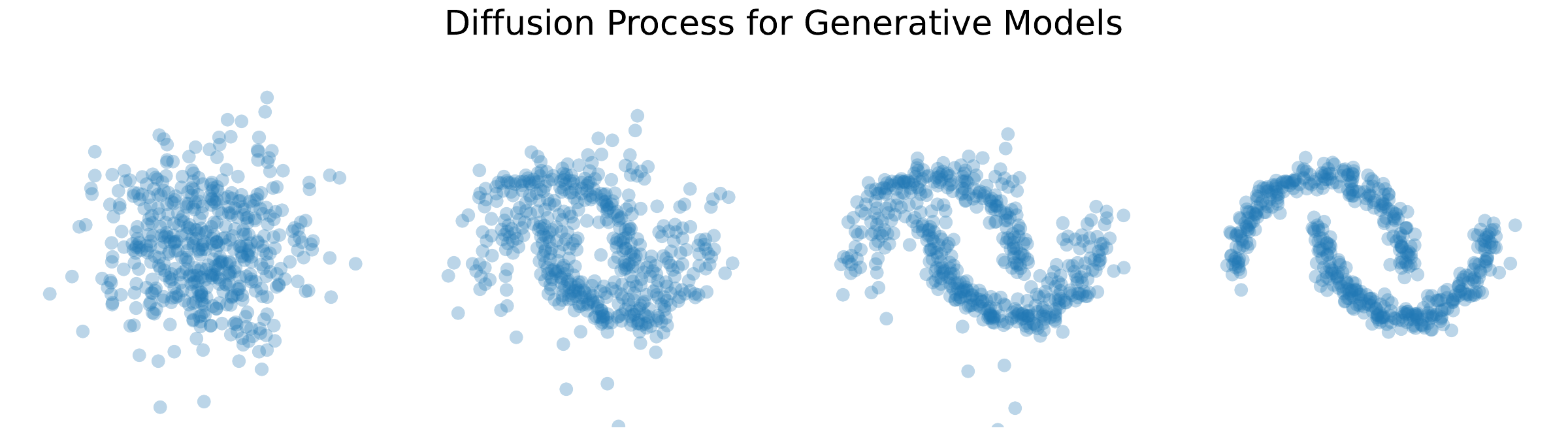}
    \centering
    \includegraphics[width=1.0\linewidth]{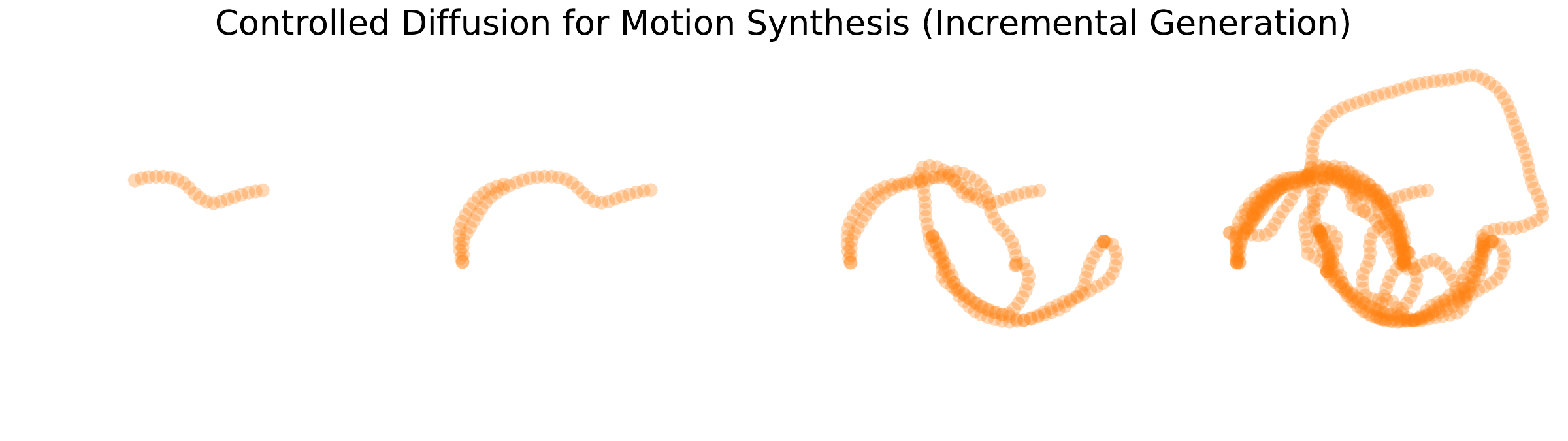}
    \centering
    \includegraphics[width=1.0\linewidth]{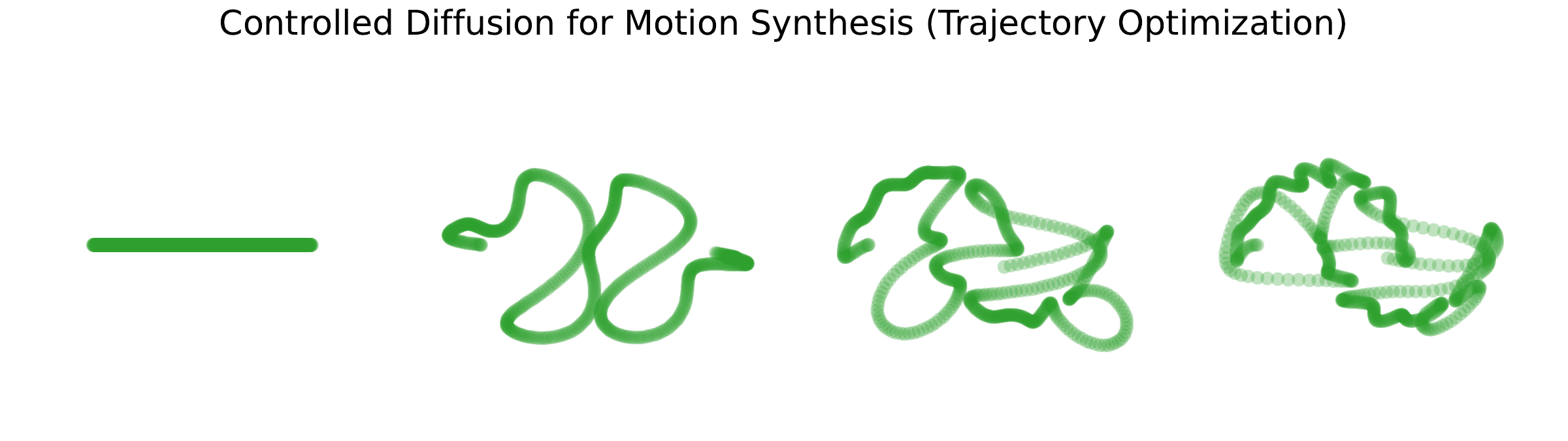}
    \vspace{-1.5em}
    \caption{For generative models, the diffusion process transforms a simple distribution into a complex one. Controlled diffusion follows the same principle, but applies the diffusion process to motion synthesis, generating trajectories that match desired time-averaged statistics. There are two perspectives on controlled diffusion: in the first, trajectories are generated incrementally over time (incremental generation), while in the second, they are iteratively optimized across the entire horizon (trajectory optimization).}
    \label{fig:diffusion_illustration}
    \vspace{-1em}
\end{figure}

We will present detailed numerical formulations of ergodic metrics in Section~4. Since the ergodic metric evaluates the discrepancy between a trajectory and a target distribution, it provides a natural basis for an optimal control problem. 

\begin{definition}[Ergodic control]
    Over a finite horizon $T$, the ergodic control problem is defined as
    \begin{gather}
        u^*(t) = \argmin_{u(t)} D\big( p_s^T(x), q(x) \big), \\
        \text{s.t. } s(t) = s_0 + \int_0^t f(s(\tau), u(\tau))\, d\tau. \nonumber
    \end{gather}
\end{definition}

\paragraph{Two perspectives on ergodic control.} 
In practice, ergodic control can be interpreted from two complementary perspectives:  

1. \emph{Incremental evolution.}  
   The system is viewed as incrementally generating a trajectory, with its empirical distribution $p_s^t$ evolving in time. Control is applied so that this evolving distribution converges toward $q$ as $t\to\infty$.  

2. \emph{Trajectory optimization.}  
   The entire trajectory over a finite horizon $T$ is optimized iteratively so that its induced empirical distribution $p_s^T$ minimizes discrepancy with $q$.  

Both perspectives are consistent with the definition of controlled diffusion: in the incremental view, the diffusion process is controlled in real time, while in the optimization view, the control problem is solved by iteratively refining the whole horizon. 
Together, these perspectives frame ergodic control as a bridge between distributional objectives and trajectory-level decision making.

\section{What Does Ergodicity Offer?}  

As mentioned above, in robot learning, the fundamental bottleneck is often not model capacity but data: the quality, diversity, and statistical structure of the collected dataset directly determine learning performance. Ergodicity provides a principled way to address these bottlenecks by shaping trajectories so that the data they generate has desirable distributional properties. We highlight three such advantages---invariance, non-myopic search, and the i.i.d. property---each directly tied to the needs of data collection for robot learning.  

\subsection{Enumeration, Permutation, and Scale Invariance}

From the perspective of motion synthesis, ergodic control specifies behaviors through their time-averaged statistics rather than explicit state sequences. This distribution-driven specification offers three invariance properties that directly benefit data collection.  

First, \emph{enumeration invariance}: conventional methods that require visiting a list of states (e.g., the traveling salesman problem) face combinatorial explosion, as the number of possible sequential orderings grows exponentially. Ergodic control generalizes this by treating the list of target states as samples from a distribution, transforming the problem into a single distribution-matching task. For robot learning, this means that datasets can be generated without explicit sequencing, avoiding combinatorial complexity while still ensuring coverage.  

Second, \emph{permutation invariance}: many tasks—such as drawing, cleaning, patrolling, or physical therapy—are better described by their aggregate outcomes than by the order of individual actions. Ergodic control ensures that the collected data reflects the desired time-averaged statistics, regardless of the order in which states are visited. This flexibility is crucial in data collection, where different action orders can yield equally valid datasets.  

Third, \emph{scale invariance}: in multi-robot systems, conventional planning approaches scale poorly with the number of agents. By specifying collective behavior through time-averaged statistics, ergodic control ensures that dataset generation remains consistent as the number of robots increases, making multi-robot data collection scalable.  

\subsection{Non-Myopic Search with Asymptotic Guarantees}

Search is fundamental to embodied intelligence and directly impacts the efficiency of data collection. In many tasks, the utility function guiding exploration is a spatial information distribution, such as mutual information. Greedy strategies maximize immediate information gain but often produce biased datasets by overfitting to local features of the landscape. Uniform coverage ensures thoroughness but wastes resources by collecting redundant data in low-value regions.  

Ergodic control systematically balances these two extremes. By biasing coverage with respect to the information distribution, ergodic trajectories allocate more time to regions with higher information density while still ensuring that lower-density regions are not ignored. This \emph{proportionality property} guarantees that the collected dataset is both informative and comprehensive, concentrating samples where they matter most while preserving global coverage.  

Moreover, since the trajectory empirical distribution converges to the information distribution asymptotically, ergodic control maintains the thoroughness of exhaustive search: any region with nonzero information density will eventually be represented in the dataset. For robot learning, this ensures that collected data supports both exploitation of critical regions and exploration of the full domain, improving sample efficiency and generalization.

\subsection{Independent and Identically Distributed Property}

One of the most important consequences of ergodicity is its connection to the independent and identically distributed (i.i.d.) assumption that underlies modern machine learning. Batch-based training methods rely on the idea that randomly drawn subsets of data are representative of the overall dataset, a property guaranteed when the data are sampled independently from a fixed underlying distribution. However, robot data are collected sequentially along trajectories subject to dynamics, which introduces temporal correlations and can also prevent the data-generation process from immediately representing a stationary distribution.

Ergodicity provides a bridge between sequential trajectory data and these two aspects of the i.i.d. requirement. First, an ergodic process evolves with respect to a stationary distribution $q(x)$, providing a fixed underlying population from which the long-run statistics of the data are defined. Second, if a system is ergodic with respect to $q(x)$, then the trajectory empirical distribution $p_s^T$ converges to $q$.  Equivalently, by Birkhoff’s ergodic theorem~\citep{birkhoff_proof_1931, walters_introduction_2000}, for any continuous test function $\phi \in \mathcal{C}(\mathcal{X})$, the long-run time average of the trajectory satisfies
\begin{align}
    \lim_{T\to\infty} \frac{1}{T}\int_0^T \phi(h(s(t))) dt = \mathbb{E}_{x\sim q}[\phi(x)].
\end{align} This equality implies that the statistics of sequentially collected data along an ergodic trajectory are statistically indistinguishable from those obtained by drawing i.i.d. samples from $q(x)$. 

In other words, ergodicity enforces a measure-preserving property: although data points are generated sequentially constrained by physical dynamics, they retain the statistical guarantees of i.i.d. sampling in the limit. For robot learning, this makes ergodic control uniquely suited as a motion synthesis strategy for data collection, ensuring that datasets collected on hardware or in simulation are representative of the target distribution and statistically compatible with the assumptions of learning algorithms.

\section{Three Perspectives on Control-As-Inference}

In recent years, the broad framework of control-as-inference has gained significant attention~\citep{todorov_linearly-solvable_2006, toussaint_robot_2009, haarnoja_soft_2018, levine_reinforcement_2018}. This framework views control synthesis as an inference problem, where controls are generated as samples from an inferred probability distribution based on the task, including the reward function, observations, and offline data. This framework can be traced back to reinforcement learning, where the optimal policy is interpreted as a probability distribution over actions that is proportional to the exponential of the action-value function. Sampling-based control methods such as model predictive path integral (MPPI) control make this perspective explicit by sampling a distribution over control sequences and reweighting the samples according to their trajectory costs~\citep{kappen_linear_2005, theodorou_generalized_2010, williams_aggressive_2016, williams_information_2017}. 

Ergodic control, however, offers a new interpretation of this framework. Instead of viewing inference as synthesizing a batch of control sequences as samples, ergodic control views inference as synthesizing a single control sequence while treating the entire state trajectory as a batch of state samples. Both interpretations are rooted in a fundamental shift in the state representation of a control problem, from the conventional point-based representation to a probabilistic distribution-based representation.

Aside from the two interpretations above, the derivation of ergodic control also leads to a third interpretation of control-as-inference. In this view, the state evolution of a nonlinear dynamical system is inferred through a distribution-based representation of the state. By representing the state as a time-varying spatial distribution, its evolution can be described as the linear evolution of functional decomposition coefficients in an infinite-dimensional observable space, connecting this perspective to Koopman operator theory~\citep{mezic_spectral_2005}. For robot learning, the Koopman operator has played an increasingly important role by providing structured representations that are naturally suited for dynamics modeling and control~\citep{abraham_active_2019, shi_koopman_2026}, while allowing one to assess how much of the system dynamics has been captured from data. From a control-as-inference perspective, learning the Koopman representation amounts to inferring an observable space in which the system evolution becomes predictable and control becomes tractable. Furthermore, the connection between ergodic theory and the Koopman operator has long been recognized through the evolution and spectral decomposition of observables under measure-preserving dynamics. Ergodic theorems relate long-time trajectory averages to expectations under an invariant measure, while the Koopman spectrum characterizes the recurrent and mixing structure of the system~\citep{arbabi_ergodic_2017, otto_koopman_2021, brunton_modern_2022}.

Consider a continuous-time dynamical system $\dot{s}(t) = f(s(t))$, where $s(t)$ denotes the state at time $t$. An alternative representation of the state is a density concentrated at $s(t)$, denoted as $p(t,x) = \delta(x-s(t))$. This time-varying spatial distribution represents the trajectory of the system as a moving delta distribution in the state space.

We denote a set of basis functions $\{F_k(x)\}_{k=1}^{K}$, such that any function $f(x)$ can be reconstructed as follow:
\begin{align}
    \lim_{K\mapsto\infty} \sum_{k=1}^{K} \langle F_k(x), F_k(x) \rangle \cdot f(x) = f(x), \ \forall x\in\mathcal{X}, 
\end{align} where $\langle,\rangle$ denotes the inner product between these two functions. A good example of a set of basis functions with the above property are the orthogonal basis functions, such as the Fourier basis functions. With the basis functions, we can rewrite the time-varying distribution that represents the state as follow:
\begin{align}
    p(t, x) = \sum_{k=1}^{K} c_k(x) \cdot F_k(x), \qquad c_k(x) = \langle F_k(x), p(t,x) \rangle.
\end{align} Therefore, we can represent the system state as a set of time-varying coefficients $C(t) = [c_1(t), \dots, c_K(t)]$. Furthermore, based on the property of the Dirac delta function, the coefficients can be rewritten as:
\begin{align}
    c_k(t) = \langle F_k(x), p(t,x) \rangle = \langle F_k(x), \delta(x - s(t)) \rangle = F_k(s(t)). 
\end{align} A key observation here is that although the original state dynamics $\dot{s}(t) = f(s(t))$ may be nonlinear in $s(t)$, the corresponding temporal evolution of density function $p(t,x)$ is linear in $p(t,x)$, based on the Liouville equation (also called transport equation)\footnote{The Liouville equation connects this derivation to flow matching~\citep{lipman_flow_2023}, another generative modeling paradigm closely related to diffusion models.}:
\begin{align}
    \frac{\partial p(t,x)}{\partial t} = -\nabla_x \cdot \big(f(x)p(t,x)\big). \label{eq:liouville}
\end{align} We show the detailed derivation at the end of the chapter.

Based on the Liouville equation (\ref{eq:liouville}), the time derivative of the coefficients $c_k(t)$ can be written as:
\begin{align}
    \dot{c}_k(t) & = \frac{d}{dt}\langle  F_k(x),p(t,x)\rangle \\
    & = \left\langle F_k(x), \frac{\partial p(t,x)}{\partial t} \right\rangle \\
    & = \left\langle F_k(x), -\nabla_x \cdot \big(f(x)p(t,x)\big) \right\rangle.
\end{align} For the right hand side of the Liouville equation, using the linearity of the divergence operator, we have:
\begin{align}
    -\nabla_x \cdot \left( f(x)\sum_j c_j(t)F_j(x) \right) = \sum_j c_j(t) \left[ -\nabla_x \cdot \big(f(x)F_j(x)\big) \right].
\end{align} Substituting the above into the inner product for $\dot{c}_k(t)$ and using the linearity of the inner product,
\begin{align}
    \dot{c}_k(t) & = \left\langle F_k(x), \sum_j c_j(t) \left[ -\nabla_x \cdot \big(f(x)F_j(x)\big) \right] \right\rangle \\
    & = \sum_j c_j(t)
    \left\langle F_k(x), -\nabla_x \cdot  \big(f(x)F_j(x)\big) \right\rangle.
\end{align} Therefore, for the (infinite-dimensional) vector of all the coefficients $C(t) = [c_k(t)]_{k=1}^K$, which can exactly recover the $p(t,x)$ thus the original system state $s(t)$, its dynamics can be written as a linear system:
\begin{align}
    \dot{c}(t) = Ac(t), \qquad C(t)= \begin{bmatrix} c_1(t) & \cdots & c_K(t) \end{bmatrix}^\top,
\end{align} where
\begin{gather}
    \dot{c}_k(t) = \sum_j A_{kj}c_j(t), \quad A_{kj} = \left\langle F_k(x), -\nabla_x \cdot  \big(f(x)F_j(x)\big) \right\rangle.
\end{gather} 

The above derivation provides the third perspective on control-as-inference by connecting it to Koopman operator theory, which studies nonlinear dynamical systems as linear systems in an infinite-dimensional observable space. By representing the system state as a time-varying spatial distribution, this perspective enables inference of state evolution of nonlinear dynamical systems through the linear evolution of functional decomposition coefficients. This view is also closely related to the operator-theoretic interpretation of ergodicity, where the long-term behavior of a dynamical system is characterized by time averages of observables. \\

\noindent\textbf{Derivation of Liouville equation (\ref{eq:liouville}) } For any smooth test function $\phi(x)$, based on the property of the Dirac delta function,
\begin{align}
    \langle \phi(x), p(t,x) \rangle = \langle \phi(x), \delta(x-s(t)) \rangle = \phi(s(t)). \label{eq:delta_property}
\end{align}
Differentiating both sides with respect to time gives
\begin{align}
    \frac{d}{dt} \langle \phi(x), p(t,x) \rangle
    = \frac{d}{dt}\phi(s(t)) = \nabla \phi(s(t))^\top \dot{s}(t) = \nabla \phi(s(t))^\top f(s(t)).
\end{align} The right hand side of the above equation can be considered a joint function $g(\cdot) = \nabla \phi(\cdot)^\top f(\cdot)$ evaluated at $s(t)$, 
\begin{align}
    \nabla \phi(s(t))^\top f(s(t)) = \langle \nabla \phi(x)^\top f(x), p(t,x) \rangle.
\end{align} Using the product rule for divergence, we have
\begin{align}
    \nabla_x {\cdot} \big(\phi(x) f(x)p(t,x)\big) & = \nabla \phi(x)^\top f(x)p(t,x) + \phi(x)\nabla_x {\cdot} \big(f(x)p(t,x)\big) \\
    & = \nabla_x {\cdot} \big(\phi(x) f(x)p(t,x)\big) - \phi(x)\nabla_x {\cdot} \big(f(x)p(t,x)\big).
\end{align} Therefore,
\begin{align}
    \nabla \phi(x)^\top f(x) p(t,x) = \nabla_x \cdot \big(\phi(x) f(x)p(t,x)\big) - \phi(x)\nabla_x \cdot \big(f(x)p(t,x)\big).
\end{align}
Integrating both sides over the state space gives
\begin{align}
    & \int \nabla \phi(x)^\top f(x)p(t,x) dx = \nonumber \\
    & \int \nabla_x \cdot \big(\phi(x) f(x)p(t,x)\big) dx - \int \phi(x) \nabla_x \cdot \big(f(x)p(t,x)\big) dx.
\end{align}
By the divergence theorem, the first term becomes a boundary integral,
\begin{align}
    \int \nabla_x \cdot \big(\phi(x) f(x)p(t,x)\big) dx = \int_{\partial \Omega} \phi(x) f(x)p(t,x)\cdot n(x) dS,
\end{align}
where \(n(x)\) is the outward normal vector on the boundary. This boundary term vanishes, for example, if \(\phi\) has compact support, if \(p(t,x)\) decays sufficiently fast at infinity, or if there is no probability flux through the boundary. Hence,
\begin{align}
    \int \nabla \phi(x)^\top f(x)p(t,x) dx = -\int \phi(x)\nabla_x \cdot \big(f(x)p(t,x)\big) dx.
\end{align} Therefore,
\begin{align}
    \int \phi(x)\frac{\partial p(t,x)}{\partial t} dx = -\int \phi(x)\nabla_x \cdot \big(f(x)p(t,x)\big) dx.
\end{align} Since this holds for arbitrary test functions \(\phi(x)\), it follows that 
\begin{align}
    \frac{\partial p(t,x)}{\partial t} = -\nabla_x \cdot \big(f(x)p(t,x)\big).
\end{align}

\chapter{Ergodic Control: Numerical Recipes}

\begin{figure}[t!]
    \centering
    \includegraphics[width=\linewidth]{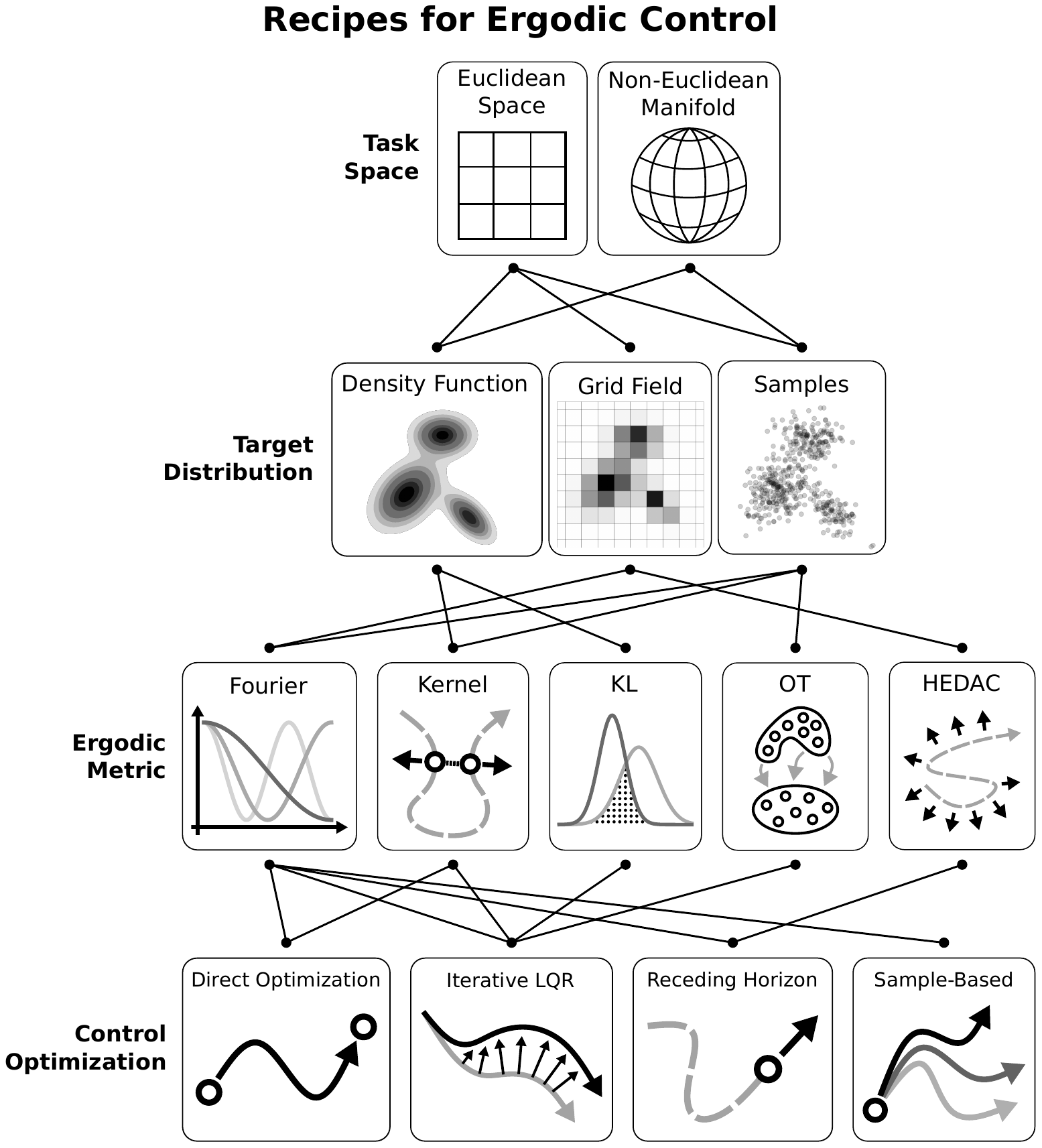}
    \caption{Diagrams for specifying an ergodic control implementation.}
    \label{fig:recipes_overview}
\end{figure}

The chapter reviews the existing numerical formulation and algorithms of ergodic control. An overview of the interactions between different numerical components of the ergodic control process is shown in Figure~\ref{fig:recipes_overview}. We provide a collection of open source implementations and tutorials of ergodic control methods in \url{https://github.com/MurpheyLab/ergodic-control-sandbox}.

\section{State and Task Space}

The task space is often a subset of the robot's state space. For example, for a wheeled robot (e.g., Mars rover) exploring an planar surface, the state space includes the robot's position and orientation, while the task space only concerns the planar position of the robot. On the other hand, for a robot arm, the state space can be the joint space of the robot, which is a subset of the Euclidean space, while the coverage task can be defined on the robot's end effector, which can be a non-Euclidean Lie group such as SE(3)~\citep{hughes_ergodic_2024}. In this case, the mapping $h$ is equivalent to the robot's forward kinematics. Note that, the choice of the robot state space should be motivated by the formulation of control synthesis while the choice of the task space should be motivated by coverage task.

\section{Target Distribution} 

There are three commonly used numerical representations for the target distribution: grid-based density function, continuous density function, and sample-based representation.

\subsection{Grid-Based Density Function} 

Without losing the generality, we assume the task space $\mathcal{X}$ is the subspace of a $d$-dimensional Euclidean space, and discretized into a grid with cells indexed by multi-indices $\mathbf{i}\in\mathbb{N}_0^{d}$, with the index set denote as $\mathcal{I}$. The target distribution is represented as a set of density values $\{q_{\mathbf{i}}\}_{\mathbf{i}\in\mathcal{I}}$. To represent a valid probability density function, the density values must satisfy the normalization requirement $\sum_{\mathbf{i}\in\mathcal{I}} q_{\mathbf{i}}=1$. Grid-based density function is the most commonly used representation for target distributions in practice, as it can be directly transformed from images and also a widely used numerical representation for modeling spatial phenomenons.

\subsection{Continuous Density Function} 

The most rigorous and comprehensive representation of a probability distribution, a continuous density function $q:\mathcal{X}\mapsto\mathbb{R}_0^+$ represents a continuous mapping from any point in the task space to the corresponding probability density value. The density function must satisfy the normalization requirement to represent a valid probability distribution $\int_{\mathcal{X}} q(x) dx = 1$. However, due to finite precision and memory limitations, it is infeasible to represent an arbitrary continuous function exactly in a digital computational framework. Therefore, practical representations for continuous density functions are based on parameterized formulations, such as through Gaussian-mixture models, kernel methods, and more recently through neural networks (NNs). 

\subsection{Sample-Based Representation} 

A target distribution can be empirically represented as a set of $N$ samples $\{q_i\}_{i=1}^{N} \subset \mathcal{X}$ from the task space, the empirical distribution of which asymptotically converges to the exact probability density function $q(x)$ as $N$ approaches infinity. Sample-based distribution representation has long been used in statistical inference, such as in particle filters. In recently years, it is also becoming common for the samples to be generated from generative models. 

\section{Ergodic Metric}

Once the target distribution is specified, we need to specify the formulation of the ergodic metric to evaluate the discrepancy between the trajectory empirical distribution and the target distribution. While there exists a wide range of statistical discrepancy measures, not all of them are feasible as the ergodic metric. The main challenge in specifying the ergodic metric lies in the use of the Dirac delta function within the trajectory’s empirical distribution. The Dirac delta is not a classical function, but rather a generalized function—also called a distribution—that is not necessarily compatible with commonly used discrepancy measures. 

Table~\ref{tab:ergodic_metric_comparison} consolidates the numerical
requirements and practical tradeoffs of the five ergodic metrics discussed
above. The appropriate choice depends primarily on the available
representation of the target distribution, the dimensionality of the task
space, and the length of the planning horizon.

\begin{table*}[t]
\centering
\caption{Comparison of ergodic metrics.}
\label{tab:ergodic_metric_comparison}
\scriptsize
\setlength{\tabcolsep}{3pt}
\renewcommand{\arraystretch}{1.25}

\begin{tabularx}{\textwidth}{
    >{\raggedright\arraybackslash}p{0.15\textwidth}
    *{5}{>{\raggedright\arraybackslash}X}
}
\toprule & \textbf{Fourier} & \textbf{Kernelized} & \textbf{MMD} & \textbf{KL} & \textbf{Sinkhorn} \\
\midrule

\textbf{Target representation} & Grid / Samples & Normalized density & Samples & Score function & Grid / samples \\
\midrule

\textbf{Time complexity} & $\mathcal{O}(MK^d)$ & $\mathcal{O}(M^2d)$ &
$\mathcal{O}((M^2+MN)d)$ & $\mathcal{O}(M^2d)$ & $\mathcal{O}(L(M^2+MN)d)$
\\
\midrule

\textbf{Space complexity} & $\mathcal{O}(K^d)$ & $\mathcal{O}(M^2)$ & $\mathcal{O}(M^2+MN)$ & $\mathcal{O}(M^2)$ & $\mathcal{O}(M^2+MN)$
\\
\midrule

\textbf{Dimension scaling} &
Exponential & Linear & Linear & Linear & Linear \\
\midrule

\textbf{Domain} & Bounded rectangular Euclidean domain & Any domain with a suitable kernel & Any domain with a suitable kernel & Euclidean; specialized extensions for manifolds & Any domain with a suitable kernel
\\
\midrule

\textbf{Advantages} & Numerically stable and robust & Scalability and flexibility across domains & Works directly with samples & Supports unnormalized target densities & Handles irregular or disjoint supports
\\
\midrule

\textbf{Limitations} & Domain restrictions; Poor dimension scaling & Target representation restrictions & Prone to local optimum & Requires an accurate score function & High computational cost
\\
\bottomrule
\end{tabularx}

\vspace{0.3em}
\begin{minipage}{0.98\textwidth}
\scriptsize
\textit{Notation:}
$M$ denotes the number of discretized trajectory points,
$N$ the number of target samples,
$d$ the task-space dimension,
$K$ the number of Fourier coefficients per dimension,
and $L$ the number of Sinkhorn iterations.
The stated space complexities assume direct storage of the pairwise kernel
or cost matrices and can be reduced in practice through parallelization.
\end{minipage}
\end{table*}

\subsection{Fourier-Based Sobolev Distance} 

\subsubsection{Overview} 

While standard $L^p$ distance metrics are not well defined over Dirac delta functions—thus the trajectory empirical distribution—the Sobolev distance in the negative Sobolev space provides an alternative distance metric compatible with the Dirac delta functions. The Fourier-based Sobolev distance is the first metric proposed for ergodic control~\citep{mathew_spectral_2009, mathew_metrics_2011}, and remains one of the most commonly used ergodic metric. We will refer to it as Fourier ergodic metric in the rest of the paper.

\subsubsection{Sobolev Space}

An $s$-th order Sobolev space $H^{s}$ is a vector space of functions equipped with a norm---called the Sobolev norm---that combines the $L^p$ norms of the function and its derivatives up the $s$-th order. A negative Sobolev space $H^{-s}$ is the dual space of the Sobolev space $H^{s}$, which consists of all continuous linear functionals on the Sobolev space $H^{s}$. The motivation of formulating the distance metric in the negative Sobolev space as the ergodic metric is threefold. while the Dirac delta function does not belong to any well-defined $L^p$ or Sobolev space, its inner product with functions from the Sobolev space is well-defined; therefore, it belongs to the negative Sobolev space, where it has a well-defined norm suitable for defining the distance metric. Second, unlike a regular $L^p$ space, the Sobolev space captures not only the magnitude of a function but also its smoothness by incorporating the derivatives of the function into the norm. This makes it robust against high-frequency noise that might exist within empirical distributions, which is a desirable property for the ergodic metric. Lastly, the Sobolev distance can be expressed in terms of the decay of Fourier coefficients---the inner products between the function and the Fourier basis functions---which leads to smooth and differentiable numerical evaluations in practice. 

\subsubsection{Formulation} 

Without loss of generality, we define the normalized Fourier basis function over a $n$-dimensional rectangular task space $\mathcal{X}{=}[0,L_1]{\times\cdots\times}[0,L_n]\subset\mathbb{R}^n$:
\begin{align}
    f_{\mathbf{k}}(x) = \frac{1}{h_{\mathbf{k}}} \prod_{i=1}^{n} \cos\left( \frac{k_i\pi}{L_i} x_i \right)
\end{align} where
\begin{gather}
        x = [x_1, x_2, \cdots, x_n] \in \mathcal{X}, \nonumber \\
        \mathbf{k} = [k_1, \cdots, k_n] \in \mathcal{K}, \nonumber \\
        \mathcal{K} = [0, 1, \cdots, K_1]\times\cdots\times [0, 1, \cdots, K_n] \subset \mathbb{N}_0^n, \nonumber
\end{gather} where the normalization term $h_{\mathbf{k}}$ ensures the $L^2$ norm of $f_{\mathbf{k}}$ equals to $1$ and can be evaluated analytically as follow:
\begin{align}
    h_{\mathbf{k}} = \sqrt{ \prod_{i=1}^{n} \left[ 
    \begin{cases}
    L_i, & \text{if } K_i = 0, \\
    \frac{L_i}{2}, & \text{if } K_i > 0.
    \end{cases}
    \right] }
\end{align} The Fourier coefficient $c_{\mathbf{k}}$ between a Fourier basis function $f_{\mathbf{k}}(x)$ and the target distribution, represented as a probability density function $q(x)$, is defined as:
\begin{align}
    c_{\mathbf{k}} = \mathbb{E}_{q}[f_{\mathbf{k}}] = \int_{\mathcal{X}} f_{\mathbf{k}}(x) q(x) dx. 
\end{align} Note that, given the orthonormality of the Fourier basis functions, the probability density function can be reconstructed through the Fourier coefficients as follow:
\begin{align}
    q(x) = \sum_{\mathbf{k}\in\mathcal{K}} c_{\mathbf{k}} \cdot f_{\mathbf{k}}(x),
\end{align} which asymptotically converges to the exact probability density function as the number of coefficients approaches infinity. Similarly, the Fourier coefficient $\phi_{\mathbf{k}}$ between the Fourier basis function $f_{\mathbf{k}}(x)$ and the trajectory empirical distribution $p_s(x)$ is defined as:
\begin{align}
    \phi_{\mathbf{k}} = \mathbb{E}_{p_s}[f_{\mathbf{k}}] = \int_{\mathcal{X}} f_{\mathbf{k}}(x) p_s(x) dx. 
\end{align} Furthermore, based on the inner product property of the Dirac delta function, the evaluation of $\phi_{\mathbf{k}}$ is equivalent to:
\begin{align}
    \phi_{\mathbf{k}} & = \int_{\mathcal{X}} f_{\mathbf{k}}(x) \left[ 
\frac{1}{T} \int_0^T \delta\Big( x - s(t) \Big) dt \right] dx \nonumber \\
    & = \frac{1}{T} \int_0^T \left[ \int_{\mathcal{X}} f_{\mathbf{k}}(x) \delta\Big( x - s(t) \Big) dx \right] dt \nonumber \\
    & = \frac{1}{T} \int_0^T f_{\mathbf{k}}\Big( s(t) \Big) dt.
\end{align} The Fourier-based Sobolev distance is defined as:
\begin{align}
    D(P_s, Q) = \sum_{\mathbf{k}\in\mathcal{K}} \lambda_{\mathbf{k}} (\phi_{\mathbf{k}} - c_{\mathbf{k}})^2, \quad \lambda_{\mathbf{k}} = \left( 1 + \Vert \mathbf{k} \Vert^2 \right)^{-\frac{d+1}{2}}. \label{eq:fourier_metric}
\end{align} where $\{\lambda_{\mathbf{k}}\}$ is a convergent series that bounds the metric. 

\begin{figure}[t!]
    \centering
    \includegraphics[width=1.0\linewidth]{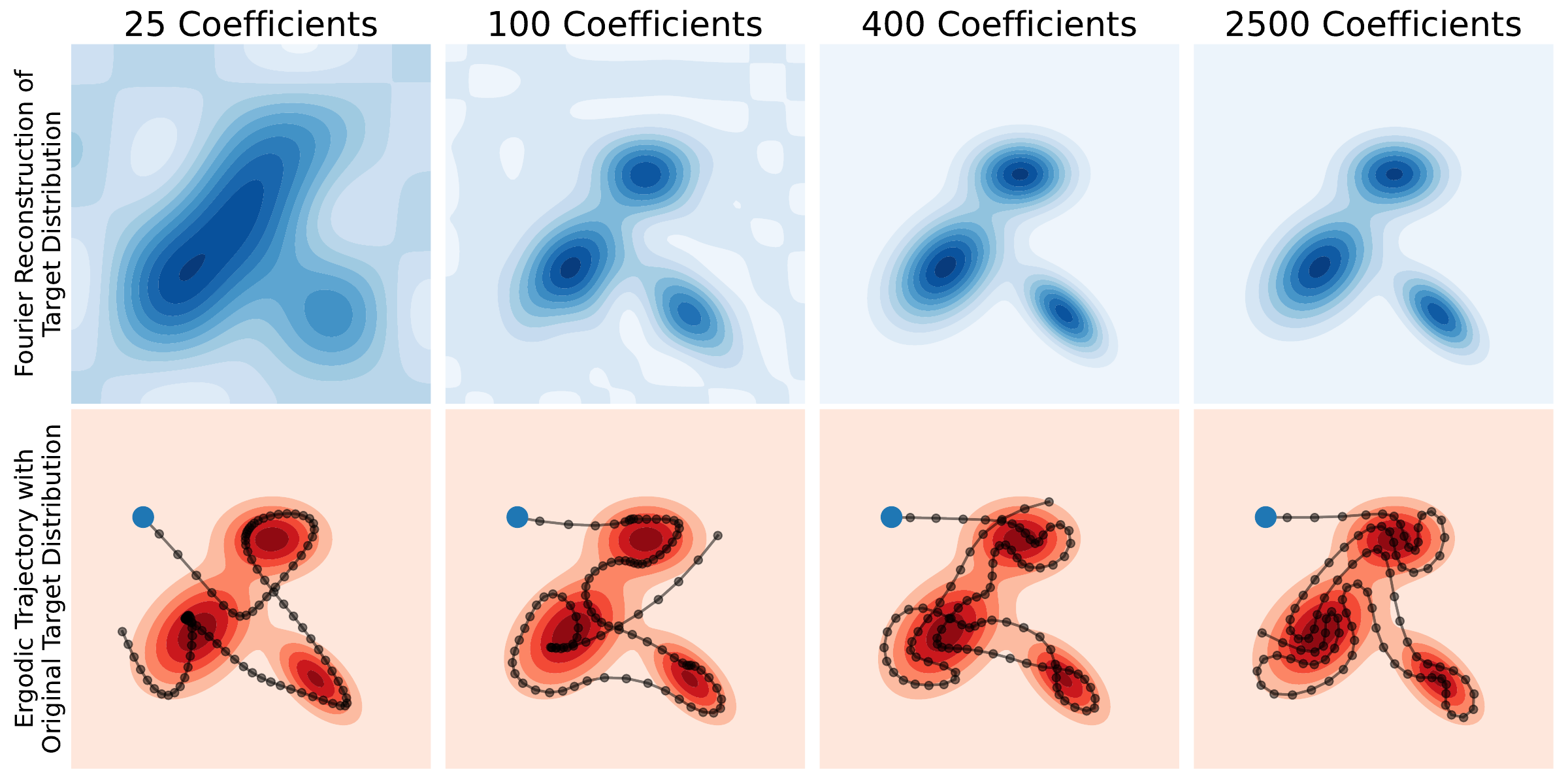}
    \caption{The number of Fourier coefficients determines the reconstruction accuracy of the target distribution and the resulting ergodic trajectory.}
    \label{fig:fourier_coefficients_comparison}
    \vspace{-1em}
\end{figure}

The formulation (\ref{eq:fourier_metric}) asymptotically converges to the exact Sobolev distance as the number of Fourier coefficients approaches infinity. In practice, a finite number of Fourier coefficients will be used to approximate the distance metric as the Fourier ergodic metric. Figure~\ref{fig:fourier_coefficients_comparison} shows how different numbers of Fourier coefficients leads to different granularity of approximation and converged trajectories. 

\subsubsection{Properties}

The Fourier ergodic metric provides a smooth ergodic metric formulation in practice, which enables gradient-based motion planning. Given a target distribution, the exact Sobolev space is convex in the function space with respect to the trajectory empirical distribution. While the control synthesis problem using the metric might not be convex due to the truncated number of Fourier coefficients, the nonlinear dynamics, and other constraints of the problem, the asymptotic function space convexity still leads to high-quality trajectories at convergence in practice. 

The computational complexity of evaluating the Fourier ergodic metric is $\mathcal{O}(T\prod_{i=1}^{d}K_i)$, excluding the computation of the Fourier coefficients for the target distribution, where $T$ is the time horizon of the trajectory and $K_i$ is the number of Fourier coefficient indices at the $i$-th dimension of the task space. In other words, the evaluation of the Fourier ergodic metric is efficient over long-horizon trajectories, but the computational cost scales exponentially over task space dimensions. The computational complexity for the Fourier coefficients of the target distribution is similar, but varies depending on the specific spatial integration scheme used. Despite the high computational complexity, the evaluation of each Fourier coefficient is independent from each other, which leads to effective acceleration schemes on GPUs. Therefore, the computational cost of the Fourier ergodic metric is sufficient for real-time decision-making for low-dimensional problems, such as 2D or 3D spaces.  

Lastly, the Sobolev space-based formulation is inherently robust against high-frequency signals, often caused by outliers either from the trajectory or the target distribution, compared to $L^p$ distance-based metrics. This property has been further applied in robot perception, where it is used as a robust metric for point cloud registration~\citep{sun_scale-invariant_2023}.

\subsection{Kernelized $L^2$ Distance} \label{sec:kernel_ergodic_control}

\subsubsection{Overview} 

Even though we cannot directly evaluate the $L^2$ distance between the target distribution and the trajectory empirical distribution, the formulation of the $L^2$ distance combined with the inner product property of Dirac delta function enables a kernelized distance metric as the ergodic metric~\citep{sun_fast_2025}. Compared to the Fourier basis functions, the kernel functions introduce extra flexibility over non-Euclidean space and improves the scalability of the ergodic metric. 

\subsubsection{Formulation} 

The formulation of the squared $L^2$ distance between the target distribution and the trajectory empirical distribution can be written as:
\begin{align}
    & \int_{\mathcal{X}} \left( q(x) - p_s(x) \right) \left( q(x) - p_s(x) \right) dx \nonumber \\
    & = \int_{\mathcal{X}} q(x) q(x) dx - 2 \int_{\mathcal{X}} q(x) p_s(x) dx + \int_{\mathcal{X}} p_s(x) p_s(x) dx. 
\end{align} The first term is a constant given the target distribution and hence can be ignored during optimization. The second term can be further expanded as:
\begin{align}
    \int_{\mathcal{X}} q(x) p_s(x) dx & = \int_{\mathcal{X}} q(x) \left[ 
\frac{1}{T} \int_0^T \delta\Big( x - s(t) \Big) dt \right] dx \nonumber \\
    & = \frac{1}{T} \int_0^T \left[ \int_{\mathcal{X}} q(x) \delta\Big( x - s(t) \Big) dx \right] dt \nonumber \\
    & = \frac{1}{T} \int_0^T q\Big( s(t) \Big) dt. \label{eq:kernel_second_term}
\end{align} The third term can also be further expanded:
\begin{align}
    & \int_{\mathcal{X}} p_s(x) p_s(x) dx \nonumber \\
    & = \int_{\mathcal{X}} \left[ 
\frac{1}{T} \int_0^T \delta\Big( x - s(t) \Big) dt \right] \left[ 
\frac{1}{T} \int_0^T \delta\Big( x - s(t^\prime) \Big) dt^\prime \right] dx \nonumber \\
    & = \frac{1}{T^2} {\int_0^T\int_0^T} \left[ \int_{\mathcal{X}} \delta\Big( x - s(t) \Big) \delta\Big( x - s(t^\prime) \Big) dx \right] dt dt^\prime \nonumber \\
    & = \frac{1}{T^2} {\int_0^T\int_0^T} \delta\Big( s(t) - s(t^\prime) \Big) dt dt^\prime. \label{eq:kernel_third_term}
\end{align} The kernelized distance metric is based on the formulations (\ref{eq:kernel_second_term}) and (\ref{eq:kernel_third_term}), where the Dirac delta function is relaxed as a kernel function $K:\mathcal{X}\times\mathcal{X}\mapsto\mathbb{R}_0^+$:
\begin{align}
    D(P_s, Q) & = \frac{1}{T^2} {\int_0^T\int_0^T} K\Big( s(t),  s(t^\prime) \Big) dt dt^\prime - \frac{2}{T} \int_0^T q\Big( s(t) \Big) dt. \label{eq:kernel_metric}
\end{align} In practice, the kernel function is often specified as a radial basis function:
\begin{align}
    K_{\alpha,\theta}(x, x^\prime) = \alpha\cdot\exp\left( -\frac{\Vert x-x^\prime \Vert^2}{\theta} \right), 
\end{align} where the kernel function parameters can be automatically optimized using the algorithm introduced in \citet{sun_fast_2025}.

\subsubsection{Properties}

\begin{figure}[t]
    \centering
    \includegraphics[width=1.0\linewidth]{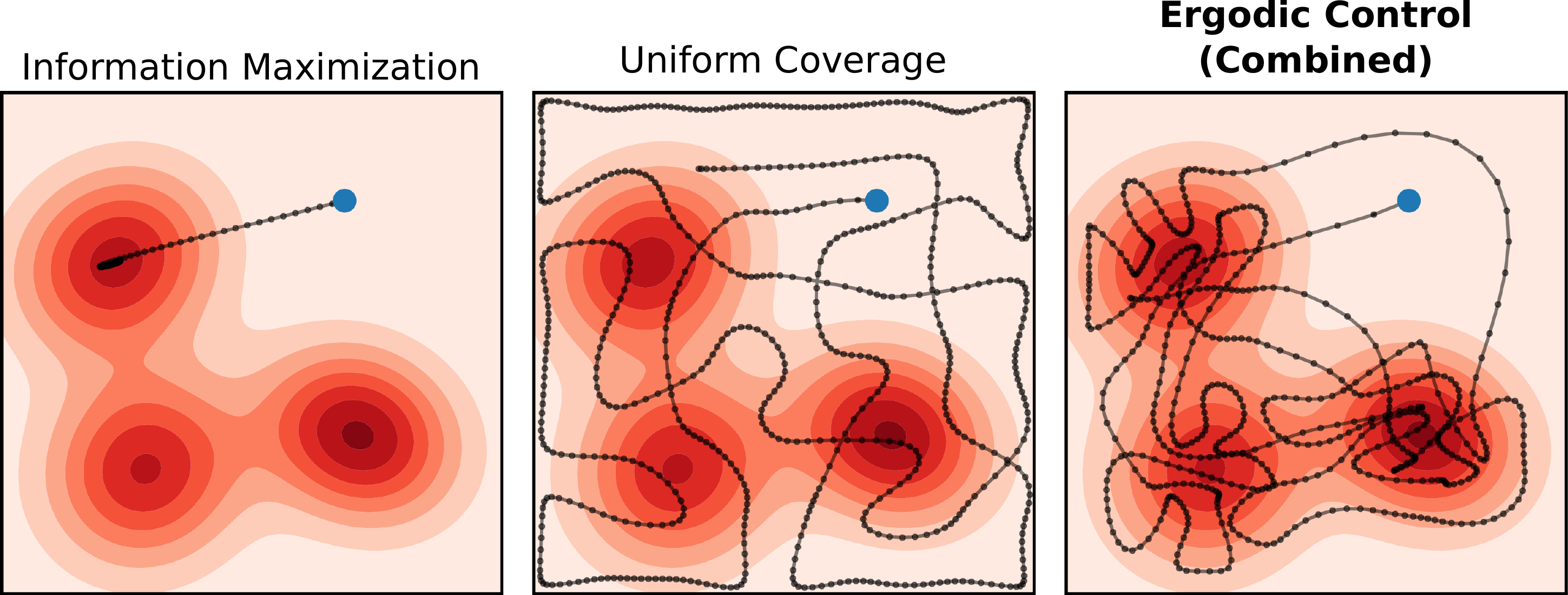}
    \caption{Kernel ergodic control effectively combined information maximization behavior with uniform coverage behavior to synthesize optimal ergodic trajectories.}
    \label{fig:kernel_elements}
    \vspace{-1em}
\end{figure}

The two terms in the formulation (\ref{eq:kernel_metric}) each offers an unique aspect of ergodic control. Minimizing the first term alone within a bounded task space leads to uniform coverage of the task space, which is equivalent to maximum entropy exploration. Minimizing the second term along is equivalent to maximum likelihood estimation of the trajectory within the target distribution, with the resulting trajectory concentrating at the maximums of the target distribution (see Figure~\ref{fig:kernel_elements}).

The kernelized ergodic metric provides a smooth metric formulation through the use of smooth kernel functions. While the Fourier ergodic metric requires spatial integral for computing the Fourier coefficients, which scales exponentially with the task space dimension, the kernelized ergodic metric eliminates the need for spatial integral, and instead replaces it with a double time integral of the system trajectory over the kernel function, which reduces the computational complexity from exponential to linear with respect to the task space dimension, while increasing the computational complexity of trajectory time horizon from linear to quadratic. 

The kernelized ergodic metric does not require pre-processing of the target distribution, such as computing the Fourier coefficients, as the metric directly evaluates the probability density function of the target distribution. While this feature makes the formulation more efficient in respond to evolving target distributions in online planning settings, it also introduces two major limitations compared to the Fourier ergodic metric. First, the formulation requires access to a differentiable and normalized probability density function of the target distribution, which is not necessarily available in practice, while the Fourier ergodic metric is compatible with all three representations of the target distributions. Second, the direct evaluation of the trajectory within the target distribution makes the method more susceptible to local maximums of the target distribution.

Lastly, one of the most important advantages of kernel function compared to the Fourier basis functions is the flexible representation in non-Euclidean spaces. While the Fourier basis function can be defined in non-Euclidean space, such as the special Euclidean group SE(2), such formulations requires a case-by-case derivation. On the other hand, kernel functions provides the much needed generalizability across different potential topological space for the ergodic control problem, including the special orthogonal group SO(3), special Euclidean space SE(3), with an unified derivation.   

\subsection{Maximum Mean Discrepancy} 

\subsubsection{Overview} Maximum mean discrepancy (MMD)~\citep{gretton_kernel_2012} is a  kernel-based statistical discrepancy measure between two probability distributions formulated as the squared $L^2$ norm in a reproducing kernel Hilbert space (RKHS). In practice, MMD is widely used as a statistical test between empirical distributions with tractable computation. While kernel functions are used in the kernelized ergodic metric as a relaxation of the Dirac delta function for approximating the $L^2$ distance, the use of kernel function in MMD has a similar intuition to Fourier ergodic metric. Kernel functions in MMD transform a distribution (including an empirical distribution) into the corresponding RKHS through the mean embedding, where the $L^2$ distance is well defined and can be computed efficiently in practice based on samples drawn from the distribution. MMD-based ergodic metric~\citep{hughes_ergodic_2024} essentially generalizes the kernel ergodic metric with sample-based target distribution representations, while preserving the advantages of kernel functions, such as the generalizability across topological spaces and computational efficiency. 

\subsubsection{Reproducing Kernel Hilbert Space}

Given a positive-definite kernel function $k:\mathcal{X}\times\mathcal{X}\mapsto\mathbb{R}_0^+$, we can define a reproducing kernel Hilbert space (RKHS) $\mathcal{H}$---a Hilbert space of functions where the evaluation of the functions at any point can be expressed as an inner product based on the kernel function. Note that, the inner product within a RKHS is not necessarily the same as an inner product in an $L^p$ space. In particular, the inner product between two kernel functions $K(x,\cdot)$ and $K(x^\prime,\cdot)$ in a RKHS satisfies the following:
\begin{align}
    \langle K(x,\cdot), K(x^\prime,\cdot) \rangle_{\mathcal{H}} = K(x, x^\prime). \label{eq:rkhs_kernel} 
\end{align} An example of the RKHS is the space of functions that are finite summations of the kernel functions:
\begin{align}
   \mathcal{H} = \left\{ f:\mathcal{X}\mapsto\mathbb{R}_0^+ \Big\vert f(\cdot) = \sum_{i=1}^{n} a_i K(\cdot, x_i) \right\}.  
\end{align} We can verify that the evaluation of a function $f(x)$ from the RKHS at any point $x\in\mathcal{X}$ can be expressed as the inner product between the function and the kernel function:
\begin{align}
    f(x) = \langle f(\cdot), K(x, \cdot) \rangle_{\mathcal{H}} = \sum_{i=1}^{n} a_i K(x,x_i), \quad \forall x\in\mathcal{X}.
\end{align}

\subsubsection{Formulation}  Given a positive-definite kernel function $K(x,x^\prime)$, we can construct a RKHS $\mathcal{H}$ as the space of functions that are finite summations of the kernel functions. Given a set of samples $\{q_i\}_{i=1}^N\subset\mathcal{X}$ representing the target distribution $Q$, we can transform the samples to a function $q_{\mathcal{X}}$ within the RKHS through the mean embedding:
\begin{align}
    q_{\mathcal{H}}(\cdot) = \frac{1}{N} \sum_{i=1}^{N} K(\cdot, q_i)
\end{align} Similarly, we can transform the robot trajectory $s(t)$ to be another function $p_{\mathcal{H}}$ within the RKHS:
\begin{align}
    p_{\mathcal{H}}(\cdot) = \frac{1}{T} \int_0^T K(\cdot, s(t)) dt.
\end{align} The maximum mean discrepancy between the target distribution $Q$ and the trajectory empirical distribution $P_{s}$ is defined as the squared norm of the difference between the corresponding RKHS embeddings of the two distributions:
\begin{align}
    MMD(P_{s}, Q) & = \Vert p_{\mathcal{H}} - q_{\mathcal{H}} \Vert^2_{\mathcal{H}} \nonumber \\
        & = \langle p_{\mathcal{H}}, p_{\mathcal{H}} \rangle_{\mathcal{H}} - 2 \langle p_{\mathcal{H}}, q_{\mathcal{H}} \rangle_{\mathcal{H}} + \langle q_{\mathcal{H}}, q_{\mathcal{H}} \rangle_{\mathcal{H}}.
\end{align} In the above formulation, the third term is a constant given the set of samples from the target distribution, thus can be ignored during optimization. The first term can be expanded as follow based on the RKHS inner product between kernel functions shown in (\ref{eq:rkhs_kernel}):
\begin{align}
    \langle p_{\mathcal{H}}, p_{\mathcal{H}} \rangle_{\mathcal{H}} & = \frac{1}{T^2} {\int_0^T\int_0^T} k\Big( s(t), s(t^\prime) \Big) dt dt^\prime. 
\end{align} The second term can be similarly expanded as follow:
\begin{align}
    \langle p_{\mathcal{H}}, q_{\mathcal{H}} \rangle_{\mathcal{H}} & = \frac{1}{T} \int_0^T \left[ \frac{1}{N} \sum_{i=1}^{N} k\Big(s(t), q_i\Big) \right] dt.
\end{align} Hence, the MMD-based ergodic metric is formulated as:
\begin{align}
    D(P_{s}, Q) & = \frac{1}{T^2} {\int_0^T\int_0^T} k\Big( s(t), s(t^\prime) \Big) dt dt^\prime \nonumber \\
        & \quad - \frac{2}{T} \int_0^T \left[ \frac{1}{N} \sum_{i=1}^{N} k\Big(s(t), q_i\Big) \right] dt. \label{eq:mmd_metric}
\end{align} Note that, if the kernel function $K(\cdot,\cdot)$ satisfies the following condition:
\begin{align}
    \int_{\mathcal{X}} K(x, x^\prime) dx^\prime = 1, \quad \forall x \in \mathcal{X},  
\end{align} such as in the case of a Gaussian kernel or kernel density estimation (KDE), the MMD ergodic metric (\ref{eq:mmd_metric}) is equivalent to the kernel ergodic metric (\ref{eq:kernel_metric}), where the density function of the target distribution is:
\begin{align}
    q(x) = \frac{1}{N} \sum_{i=1}^{N} K(x, q_i). 
\end{align}

\subsubsection{Properties}

While having separate derivations, the MMD ergodic metric (\ref{eq:mmd_metric}) and the kernel ergodic metric (\ref{eq:kernel_metric}) have similar formulations in terms of the use of kernel functions. The MMD ergodic metric can be considered a generalization of the kernel ergodic metric with sample-based target distribution representations, and the kernel ergodic metric can be considered an extension of the MMD ergodic metric where the target distributions are represented as continuous density functions. 

Due to the similarities in the formulations, the MMD ergodic metric and kernel ergodic metric have similar numerical properties as well. 

\subsection{Kullback-Leibler Divergence} 

\subsubsection{Overview}

The Kullback-Leibler (KL) divergence is one of the most commonly used statistical discrepancy measures. Given two probability distributions $P$ and $Q$ with the corresponding probablity density functions $p(x)$ and $q(x)$, the KL divergence between the two is defined as follow:
\begin{align}
    D_{KL}(P\Vert Q) = \mathbb{E}_{p(x)}\left[ \log\left(\frac{p(x)}{q(x)}\right) \right] = \int_{\mathcal{X}} p(x) \log\left(\frac{p(x)}{q(x)}\right) dx.
\end{align} Note that, unlike the distance-based ergodic metrics above, KL divergence is not a distance metric---it is asymmetric and does not satisfy the triangle inequality. 

The main challenge of applying KL divergence as an ergodic metric is that it cannot be directly evaluated and optimized over empirical distributions. However, given an empirical distribution $P_s$ and a target distribution $Q$, the descent direction of $D_{KL}(P_s\Vert Q)$ with respect to $P_s$ in the function space can be efficiently evaluated using the Stein variational gradient descent (SVGD) formulation, which leads to effective ergodic control algorithms~\citep{sun_flow_2025}. 

\subsubsection{Stein Variational Gradient Descent}

Given a set of samples $\mathbf{x}=\{x_i\}_{i=1}^{N}$, its empirical distribution $p_{\mathbf{x}}(\cdot)$ is defined as:
\begin{align}
    p_{\mathbf{x}}(x) = \frac{1}{N} \sum_{i=1}^{N} \delta (x - x_i). 
\end{align} We consider an incremental transformation $T(x) = x + \epsilon g(x)$ on the samples as a small perturbation in the direction of $g(x)$ with a magnitude of $\epsilon$. The empirical distribution after the incremental transformation is denoted as $p_{[T]}(x)$:
\begin{align}
    p_{[T]}(x) = \frac{1}{N} \sum_{i=1}^{N} \delta (x - T(x_i)). 
\end{align} Given a perturbation direction $g(x)$ and a target distribution $q(x)$, we now consider the derivative of the KL divergence $D_{KL}(p_{[T]} \Vert q)$ with respect to the perturbation magnitude $\epsilon$ evaluated at $\epsilon=0$. Based on Theorem 3.1 of~\citet{liu_stein_2016}, this derivative can be written in closed-form as follow:
\begin{align}
    \nabla_{\epsilon} D_{KL}(p_{[T]} \Vert q) \big\vert_{\epsilon=0} = -\mathbb{E}_{p_{\mathbf{x}}(x)}\left[ \text{trace}(\mathcal{A}_q g(x)) \right], \label{eq:stein_derivative}
\end{align} where $\mathcal{A}_q g(x)$ is called the Stein operator:
\begin{align}
    \mathcal{A}_q g(x) = \nabla_{x} \log q(x) g(x)^\top + \nabla_{x} g(x). 
\end{align} The derivative (\ref{eq:stein_derivative}) can be understood as the directional derivative of the KL divergence $D_{KL}(p_{\mathbf{x}} \Vert q)$ in the perturbation direction $g(x)$ on the samples $\mathbf{x}$. Therefore, in order to optimizes the samples $\mathbf{x}$ to minimize the KL divergence, we can calculate a perturbation $g(x)$ that is the descent direction of the KL divergence. One of such descent direction is formulated in Lemma 3.2 of~\citet{liu_stein_2016} as follow:
\begin{align} \label{eq:svgd}
    g(\cdot) & = \mathbb{E}_{p_{\mathbf{x}}(x)}\left[ K(x,\cdot) \nabla_x \log q(x) + \nabla_{x} K(x,\cdot) \right] \\
    & = \frac{1}{N} \sum_{i=1}^{N} \Big[ K(x_i,\cdot) \nabla_x \log q(x_i) + \nabla_{x} K(x_i,\cdot) \Big] , \label{eq:stein_grad}
\end{align} where $K(\cdot,\cdot)$ is a positive-definite kernel function. Based on Lemma 3.2 of~\citet{liu_stein_2016}, the formulation (\ref{eq:stein_grad}) is the steepest descent direction of the KL divergence within a ball $\mathcal{B}$ in a reproducing kernel Hilbert space (RKHS) $\mathcal{H}$ induced by the kernel function:
\begin{align}
    \mathcal{B} = \{ g \in \mathcal{H} : \Vert g \Vert_{\mathcal{H}}^2 \leq \mathbb{S}(p_{\mathbf{x}},q) \}, 
\end{align} where $\mathbb{S}(p_{\mathbf{x}},q)$ is the kernelized Stein discrepancy:
\begin{align}
    \mathbb{S}(p_{\mathbf{x}},q) = \max_{g} \left\{ \left[ \mathbb{E}_{p_{\mathbf{x}}(x)}[ \text{trace}(\mathcal{A}_q g(x)) ] \right]^2 \right\}.
\end{align} In Stein variational gradient descent (SVGD), the descent direction in (\ref{eq:stein_grad}) can be applied with a small $\epsilon$ iteratively to the samples, which will continuously transform the samples from the original distribution toward the target distribution $q(x)$. Similarly, the descent direction can be adapted for trajectory empirical distribution $P_{s}$ given a robot trajectory $s(t)$:
\begin{align}
    g(\cdot) & = \frac{1}{T} \int_0^T \Big[ k\Big(s(\tau),\cdot\Big) \nabla_x \log q\Big(s(\tau)\Big) + \nabla_{x} k\Big(s(\tau),\cdot\Big) \Big] d\tau \label{eq:stein_grad_traj} \\
    & = \frac{1}{T} \int_0^T \Big[ k\Big(s(\tau),\cdot\Big) \nabla_x \log q\Big(s(\tau)\Big) \Big] d\tau \nonumber \\
    & \quad + \frac{1}{T} \int_0^T \Big[ \nabla_{x} k\Big(s(\tau),\cdot\Big) \Big] d\tau  
\end{align} The descent direction above can be evaluated on each time step of the trajectory, leading to a descent direction $a(t)$ on the trajectory $a(t)=g(s(t))$:
\begin{align}
    a(t) & = \frac{1}{T} \int_0^T \Big[ k\Big(s(\tau),s(t)\Big) \nabla_x \log q\Big(s(\tau)\Big) \Big] d\tau \nonumber \\
    & \quad + \frac{1}{T} \int_0^T \Big[ \nabla_{x} k\Big(s(\tau),s(t)\Big) \Big] d\tau \label{eq:stein_descent_traj}
\end{align} As we will show later, this descent trajectory $a(t)$ can be used for ergodic control using algorithms such as iterative linear quadratic regulator (iLQR). 

\subsubsection{Properties}

The most unique part of using KL divergence as the ergodic metric is that, the metric itself cannot be directly evaluated due to the trajectory empirical distribution, but an effective descent direction on the robot trajectory can be formulated in closed form (\ref{eq:stein_grad_traj}). 

Similar to the kernel ergodic metric (\ref{eq:kernel_metric}) and the MMD ergodic metric (\ref{eq:mmd_metric}), the SVGD-based descent direction formulation (\ref{eq:stein_descent_traj}) also uses a kernel function during evaluation. In some sense, the SVGD-based formula combines features from both kernel ergodic metric and MMD ergodic metric---similar to kernel ergodic metric, the SVGD formula uses a continuous representation of the target distribution, and similar to the MMD ergodic metric, the SVGD formulation is based on the reproducing kernel Hilbert space. 

Lastly, the evaluation of the SVGD-based descent direction (\ref{eq:stein_descent_traj}) does not require the evaluation of the probability density function of the target distribution $q(x)$, but instead the derivative of the log-likelihood $\nabla_x \log q(x)$, which is also called the \emph{score function}. The most important property of the score function is that it can be derived from \emph{unnormalized} density functions instead of a probability density function. This property makes the score function an ideal representation of the target distribution in practice, especially with learning-based target distribution representations, where enforcing the normalization to the learning pipeline is often challenging. 

\subsection{Optimal Transport} 

\subsubsection{Overview} 

Compared to conventional norm-based metrics for comparing probability distributions, the theory of optimal transport (OT) provides alternative formulations with various advantages~\citep{peyre_computational_2019}, including numerical stability, robustness to noise, and sensitivity to the underlying geometry of the distributions. OT has been widely used in machine learning in recent years, often to construct advanced loss functions to improve the training of neural networks~\citep{frogner_learning_2015, kusner_word_2015, genevay_learning_2018}. Furthermore, OT has played a significant role in the advances of flow matching~\citep{lipman_flow_2023}, a state-of-the-art technique for training generative models.  

Wasserstein distance, also known as the Earth mover's distance, is one of the most commonly used OT metrics. While versatile and robust, the use of Wasserstein distance in practice, including as an ergodic metric, is often limited by its heavy computational cost. Therefore, Sinkhorn divergence is introduced as an entropic regularized formulation for Wasserstein distance, which can be efficiently evaluated and differentiated using the Sinkhorn algorithm~\citep{cuturi_sinkhorn_2013}. Interestingly, Sinkhorn divergence is equivalent to an interpolation between Wasserstein distance and the maximum mean discrepancy (MMD), essentially providing a tunable middle ground between MMD-based ergodic metric and Wasserstein distance-based ergodic metric~\citep{feydy_interpolating_2019}. 

\subsubsection{Preliminaries}

Given two probability distributions $P$ and $Q$ with the corresponding probability density functions $p(x)$ and $q(x)$, the optimal transport (OT) problems measures the difference between the two distributions as follow, based on the Kantorovich formulation:
\begin{gather}
    \min_{\gamma(x,x^\prime)} \int_{\mathcal{X}} \int_{\mathcal{X}} c(x, x^\prime) \gamma(x,x^\prime) dx dx^\prime, \label{eq:general_ot} \\
    \text{s.t., } \int_{\mathcal{X}} \gamma(x,x^\prime) dx = q(x^\prime) \text{ and } \int_{\mathcal{X}} \gamma(x,x^\prime) dx^\prime = p(x).
\end{gather} In the above formulation, $\gamma(x,x^\prime)$ is a joint distribution whose marginal distributions are equivalent to $p(x)$ and $q(x)$, and $c(x,x^\prime)$ is a cost function that varies between specific OT formulations. The intuition behind the formulation (\ref{eq:general_ot}) is that the joint distribution $\gamma(x,x^\prime)$, often also called the transportation plan, describes how to transport the masses of the distribution $p(x)$---think of it as a pile of sand---toward the another distribution $q(x)$---think of it as a shaped hole on the ground, in order to match the two. The cost function $c(x,x^\prime)$ describes the cost of transportation between the two locations $x$ and $x^\prime$, and solving the minimization problem in (\ref{eq:general_ot}) leads to the minimal cost of such a transportation.

\subsubsection{Formulation}

The Wasserstein distance specifies the cost function as the squared $L^d$ distance, which leads to the formulation:
\begin{gather}
    OT(P,Q) = \min_{\gamma(x,x^\prime)} \int_{\mathcal{X}} \int_{\mathcal{X}} \Vert x-x^\prime \Vert^d \gamma(x,x^\prime) dx dx^\prime, \label{eq:continuous_wasserstein} \\
    \text{s.t., } \int_{\mathcal{X}} \gamma(x,x^\prime) dx = q(x^\prime) \text{ and } \int_{\mathcal{X}} \gamma(x,x^\prime) dx^\prime = p(x).
\end{gather} More often in practice, the Wasserstein distance is formulated between two empirical distributions from two sets of samples $\mathbf{x}=\{x_i\}_{i=1}^{M}$ and $\mathbf{x}^\prime=\{x_j^\prime\}_{j=1}^{N}$:
\begin{gather}
   OT(P,Q) = \min_{\gamma\in\mathbb{R}^{MN}} \frac{1}{MN} \sum_{i=1}^{M} \sum_{j=1}^{N} \gamma_{ij} \cdot \Vert x_i - x_j^\prime \Vert^d, \label{eq:discrete_wasserstein} \\
    \text{s.t., } \frac{1}{M}\sum_{i=1}^{M} \gamma_{ij} = 1, \forall j \text{ and } \frac{1}{N}\sum_{j=1}^{N} \gamma_{ij} = 1, \forall i.
\end{gather} The optimization problem for Wasserstein distance, in both continuous and discrete formulations, is a convex optimization problem. In particular, the discrete formulation (\ref{eq:discrete_wasserstein}) is a standard linear programming (LP) and can be solved through LP algorithms such as the simplex algorithm or the interior point algorithm. On the other hand, while LP solvers can evaluate the exact the Wasserstein distance (\ref{eq:discrete_wasserstein}), the computational efficiency is limited for large numbers of samples. Motivated by this computational bottleneck, the Sinkhorn divergence is introduced to approximate the Wasserstein distance through an entropic regularized formulation, which significantly improves the scalability using the Sinkhorn algorithm.

For two continuous probability density functions $p(x)$ and $q(x)$, the entropic Wasserstein distance $OT_{\epsilon}(P,Q)$ is formulated as:
\begin{gather}
     \min_{\gamma(x,x^\prime)} \int_{\mathcal{X}} \int_{\mathcal{X}} \Vert x-x^\prime \Vert^d \gamma(x,x^\prime) dx dx^\prime + \epsilon \cdot D_{KL}(\gamma \Vert p \otimes q), \label{eq:continuous_sinkhorn} \\
    \text{s.t., } \int_{\mathcal{X}} \gamma(x,x^\prime) dx = q(x^\prime) \text{ and } \int_{\mathcal{X}} \gamma(x,x^\prime) dx^\prime = p(x),
\end{gather} where $\epsilon$ is the entropic regularization weight and the KL divergence regularization term can be expanded as:
\begin{align}
    D_{KL}(\gamma \Vert p \otimes q) = \int_{\mathcal{X}} \int_{\mathcal{X}} \gamma(x,x^\prime) \log\left( \frac{\gamma(x,x^\prime)}{p(x) q(x^\prime)} \right) dx dx^\prime.
\end{align} This KL divergence term regularizes the deviation of the joint distribution $\gamma(x,x^\prime)$ from the joint distribution $p(x)q(x^\prime)$, the latter of which indicates that $x$ and $x^\prime$ are independent. Similarly to (\ref{eq:discrete_wasserstein}), the entropic Wasserstein distance can also be formulated for empirical distributions:
\begin{gather}
    \min_{\gamma\in\mathbb{R}^{MN}} \frac{1}{MN} \sum_{i=1}^{M} \sum_{j=1}^{N} \gamma_{ij} \cdot \Vert x_i - x_j^\prime \Vert^d + \frac{\epsilon}{MN} \sum_{i=1}^{M} \sum_{j=1}^{N} \gamma_{ij} \log(\gamma_{ij}), \label{eq:discrete_entropic_wasserstein} \\
    \text{s.t., } \frac{1}{M}\sum_{i=1}^{M} \gamma_{ij} = 1, \forall j \text{ and } \frac{1}{N}\sum_{j=1}^{N} \gamma_{ij} = 1, \forall i.
\end{gather} Note that entropic Wasserstein distance (\ref{eq:discrete_entropic_wasserstein}) still solves a convex optimization problem, but the problem in the discrete formulation can no longer be solved as linear programming.

While the entropic Wasserstein distance is more computationally efficient compared to the standard Wasserstein distance, the KL divergence regularization term makes it asymmetric. To solve this issue, the Sinkhorn- divergence is formulated based on the entropic Wasserstein distance (\ref{eq:discrete_entropic_wasserstein}) as follow:
\begin{align}
    D(P,Q) = OT_{\epsilon}(P,Q) - \frac{1}{2} OT_{\epsilon}(P,P)  - \frac{1}{2} OT_{\epsilon}(Q,Q).
\end{align} To be used as a metric for ergodicity, the system trajectory will be discretized at a set of discretized time steps and evaluated as a set of samples. 

\subsubsection{Properties}

The Sinkhorn divergence ergodic metric is used in practice with sample-based representation of the target distribution~\citep{sun_flow_2025}. The Wassertein distance component of the metric makes it particularly suitable for non-smooth target distributions with irregular or disjoint supports. The Sinkhorn divergence interpolates between the Wasserstein distance and the maximum mean discrepancy (MMD) based on the KL divergence regulartization term with the weight parameter $\epsilon$, which provides a tuning mechanism to balance the fidelity and computational efficiency of the metric in response to the computational resource of the robot. Lastly, another unique feature of optimal transport-based ergodic metrics is that the cost function $c(x,x^\prime)$ can be specified differently based on the task, including common choices such as the $L^1$ norm and the squared $L^2$ norm.

\section{Control Optimization}

After specifying a formulation of the ergodic metric, the last step in ergodic control is to synthesis the control of the system to optimize the ergodic metric wit the system trajectory under the dynamics constraints. General control optimization methods view ergodic control as a standard optimization or optimal control problem, leading to algorithms that apply to a wide range of system dynamics and ergodic metrics. General control optimization algorithms can be applied for long-horizon trajectory optimization and online model-predictive control (MPC). On the other hand, various specialized control optimization methods have also been proposed to leverage the structure of the ergodic metric or the control problem.

\subsection{Nonlinear Programming (NLP)}

Ergodic control can be viewed as a finite-dimensional, nonlinear constrained optimization problem. The trajectory $s(t)$ is discretized as a set of time steps $\mathbf{s}=[s_1, \dots, s_t, \dots, s_T]^\top$, with discrete-time dynamics $s_{t+1} = f(s_t, u_t)$. The nonlinear programming is formulated as follow:
\begin{gather}
    \mathbf{s}^*, \mathbf{u}^* = \argmin_{\mathbf{s}, \mathbf{u}} D(P_\mathcal{\mathbf{s}}, Q), \\
    \text{s.t., } s_{t+1} = f(s_t, u_t), \quad t = 1, \dots, T. \nonumber
\end{gather} The NLP can be sovled using methods such as the trust-region method, sequential least squares programming (SLSQP), and augmented Lagrange multiplier~\citep{nocedal_numerical_2006}.

\subsection{Iterative Linear Quadratic Regulator (iLQR)}

\subsubsection{Overview}

A wide range of ergodic control formulations can also be solved as an optimal control problem through the iterative linear quadratic regulator (iLQR) algorithm. The iLQR algorithm can derived for both discrete-time and continuous-time systems with the same principle, which is to iteratively calculate a descent direction for the controls based on the local linearization of the system dynamics and the second-order approximation of the cost function. We will show the continuous-time derivation here with notes on the differences to the discrete-time derivation. 

\subsubsection{Preliminaries}

The iLQR is based on the local linearization of the system dynamics. We assume the system dynamics $\dot{s}(t)=f(s(t), u(t))$, the current control $u(t)$, and the resulting trajectory $s(t)$ with the initial state $s_0$. Given a local perturbation $v(t)$ to the control $u(t)$, the resulting perturbation $z(t)$ on the state trajectory $s(t)$ is subject to linear dynamics:
\begin{gather}
    \dot{z}(t) = A(t) z(t) + B(t) v(t), \quad z(0) = \mathbf{0}, \\
    A(t) = \nabla_{s} f(s(t), u(t)), \quad B(t) = \nabla_{u} f(s(t), u(t)). \label{eq:linearized_dynamics}
\end{gather} With the standard cost function formulation below:
\begin{align}
    J(s(\cdot), u(\cdot)) = \int_0^T l(s(t), u(t)) dt, \label{eq:ilqr_overall_cost}
\end{align} the directional derivative of the cost function with respect to the state and control perturbation is:
\begin{align}
    DJ_{z(t)}(s(t), u(t)) & = \nabla_{s} l(s(t), u(t))^\top z(t) = a(t)^\top z(t), \\
    DJ_{v(t)}(s(t), u(t)) & = \nabla_{u} l(s(t), u(t))^\top v(t) = b(t)^\top v(t).
\end{align} The iLQR algorithm iteratively constructs and solves the following linear quadratic regulator (LQR) problem:
\begin{gather}
    v(\cdot)^* {=} \argmin_{v(\cdot)} {\int_0^T} a(t)^\top z(t) + b(t)^\top v(t) + \Vert z(t) \Vert_{Q}^2 + \Vert v(t) \Vert_{R}^2 dt, \label{eq:standard_lqr} \\
    \text{s.t., } \dot{z}(t) = A(t) z(t) + B(t) v(t), \quad z(0) = \mathbf{0}. \nonumber
\end{gather} In the LQR formulation above (\ref{eq:standard_lqr}), the first two terms $a(t)^\top z(t) + b(t)^\top v(t)$ is the directional derivative of the nonlinear cost function in (\ref{eq:ilqr_overall_cost}) under the state and control perturbation of $z(t)$ and $v(t)$, and the second two terms $\Vert z(t) \Vert_{Q}^2 + \Vert v(t) \Vert_{R}^2$ can be viewed quadratic regulations on the perturbations, with the positive-definite matrices $Q$ and $R$ being user-specified parameters. The LQR cost (\ref{eq:standard_lqr}) is a local quadratic approximation of the nonlinear cost function. In the special case where the matrices are time-varying and $Q(t)=\nabla_{s}^2 l(s(t), u(t))$ and $R(t)=\nabla_{u}^2 l(s(t), u(t))$, the LQR cost function becomes the second-order Taylor expansion of the nonlinear cost function. 

The LQR problem (\ref{eq:standard_lqr}) can be solved efficiently in closed-form through the Riccati equation, the solution of which is the optimal descent direction on the current control $u(t)$ for the nonlinear cost (\ref{eq:ilqr_overall_cost}). The iLQR algorithm iteratively constructs the LQR problem (\ref{eq:standard_lqr}), computes the descent direction of control, and update the control following the descent direction, with either a fixed and sufficiently small step size, or adaptive step size through algorithms such as backtracking line search. The iLQR algorithm is essentially a gradient descnet
The LQR problem (\ref{eq:standard_lqr}) can be solved efficiently in closed-form through the Riccati equation, the solution of which is the optimal descent direction on the current control $u(t)$ for the nonlinear cost (\ref{eq:ilqr_overall_cost}). The iLQR algorithm iteratively constructs the LQR problem (\ref{eq:standard_lqr}), computes the descent direction of control, and update the control following the descent direction, with either a fixed and sufficiently small step size, or adaptive step size through algorithms such as backtracking line search. The iLQR algorithm essentially acts as a gradient descent algorithm but with the robot's nonlinear dynamics taken into account. 

\subsubsection{Ergodic Control with iLQR}

The main challenge of applying iLQR to ergodic control is that, the cost function of ergodic control—the ergodic metric—does not have the same runtime cost structure as in the standard iLQR formulation (\ref{eq:ilqr_overall_cost}). In other words, the ergodic metric cannot be formulated as a time integral of a nonlinear cost function over the trajectory. However, for several ergodic metric formulations, we can still derive the directional derivative of the ergodic metric at the current trajectory $s(t)$ in the direction of state perturbation $z(t)$~\citep{miller_ergodic_2016}.

\noindent\textbf{(Fourier ergodic metric) } Based on the formulation of the Fourier ergodic metric in (\ref{eq:fourier_metric}), we have
\begin{align} 
    \nabla D(P_{\mathbf{s}}, Q) \cdot z(t) & = 2 \sum_{\mathbf{k}\in\mathcal{K}} \lambda_{\mathbf{k}} (\phi_{\mathbf{k}} - c_{\mathbf{k}}) \cdot (\nabla \phi_{\mathbf{k}} \cdot z(t)) \nonumber \\ 
    & = 2 \sum_{\mathbf{k}\in\mathcal{K}} \lambda_{\mathbf{k}} (\phi_{\mathbf{k}} - c_{\mathbf{k}}) \cdot \left( \frac{1}{T} \nabla f_k(s(t))  \cdot z(t) \right) \nonumber \\
    & = \frac{2}{T} \left[ \sum_{\mathbf{k}\in\mathcal{K}} \lambda_{\mathbf{k}} (\phi_{\mathbf{k}} - c_{\mathbf{k}}) \nabla f_k(s(t))  \right] \cdot z(t).
\end{align} Therefore, we can still construct the LQR problem in (\ref{eq:standard_lqr}) with the $a(t)$ being:
\begin{align}
    a(t) = \frac{2}{T} \sum_{\mathbf{k}\in\mathcal{K}} \lambda_{\mathbf{k}} (\phi_{\mathbf{k}} - c_{\mathbf{k}}) \nabla f_k(s(t)) .
\end{align} \\

\noindent\textbf{(Kernel ergodic metric) } Based on the formulation of the kernel ergodic metric in (\ref{eq:kernel_metric}), we have:
\begin{align}
    & \nabla D(P_{\mathbf{s}}, Q) \cdot z(t) \nonumber \\
    & = \frac{2}{T^2} {\int_0^T\int_0^T} \nabla_1 K\Big( s(t^\prime),  s(t) \Big)   z(t) dt^\prime dt \nonumber \\
    & \quad - \frac{2}{T} \int_0^T \nabla q(s(t))  z(t) dt \nonumber \\
    & = a(t) \cdot z(t) \nonumber, 
\end{align} where
\begin{align}
    a(t) = \frac{2}{T} \left[\frac{1}{T} \int_0^T \nabla_1 K\Big( s(t^\prime),  s(t) \Big) dt^\prime - \nabla q(s(t)) \right] 
\end{align} can be substituted into the LQR formulation in (\ref{eq:standard_lqr}). \\

\noindent\textbf{(MMD ergodic metric) } Similar to the kernel ergodic metric, we can iteratively formulate the LQR problem (\ref{eq:standard_lqr}) for the MMD ergodic metric (\ref{eq:mmd_metric}) with $a(t)$ substituted as:
\begin{align}
    a(t) = \frac{2}{T} \Bigg[ \frac{1}{T} \int_0^T & \nabla_1 K\Big( s(t^\prime),  s(t) \Big) dt^\prime - \frac{1}{N} \sum_{i=1}^{N} \nabla_1 K(s(t), q_i) \Bigg] .
\end{align}

\noindent\textbf{(KL divergence) } Applying iLQR to KL divergence-based ergodic metric is different from the derivations above. As mention in Section (KL divergence number), KL divergence cannot be directly evaluated over the trajectory empirical distribution. Instead, the descent direction of the KL divergence with respect to a given state trajectory $s(t)$ can be derived using the Stein variational gradient descent (\ref{eq:stein_grad_traj}), which can serve as $a(t)$ in (\ref{eq:standard_lqr}):
\begin{align}
    a(t) = \Bigg[ \frac{1}{T} \int_0^T & \Bigg( K\Big(s(\tau),s(t)\Big) \nabla \log q\Big(s(\tau) \Big) + \nabla_1 K\Big(s(\tau),s(t)\Big) \Bigg) d\tau \Bigg] .
\end{align}

\noindent\textbf{(Sinkhorn divergence) } There is no analytical derivation of the iterative LQR formulation for Sinkhorn divergence as the ergodic metric. However, when evaluating the Sinkhorn divergence through the Sinkhorn algorithm, the derivative of the metric with respect to a discrete system trajectory can be obtained numerically through auto-differentiation, which can serve as $a(t)$ in (\ref{eq:standard_lqr}) for applying the iLQR algorithm.

\subsubsection{Flow Matching and iLQR}

\begin{figure}[t]
    \centering
    \includegraphics[width=1.0\linewidth]{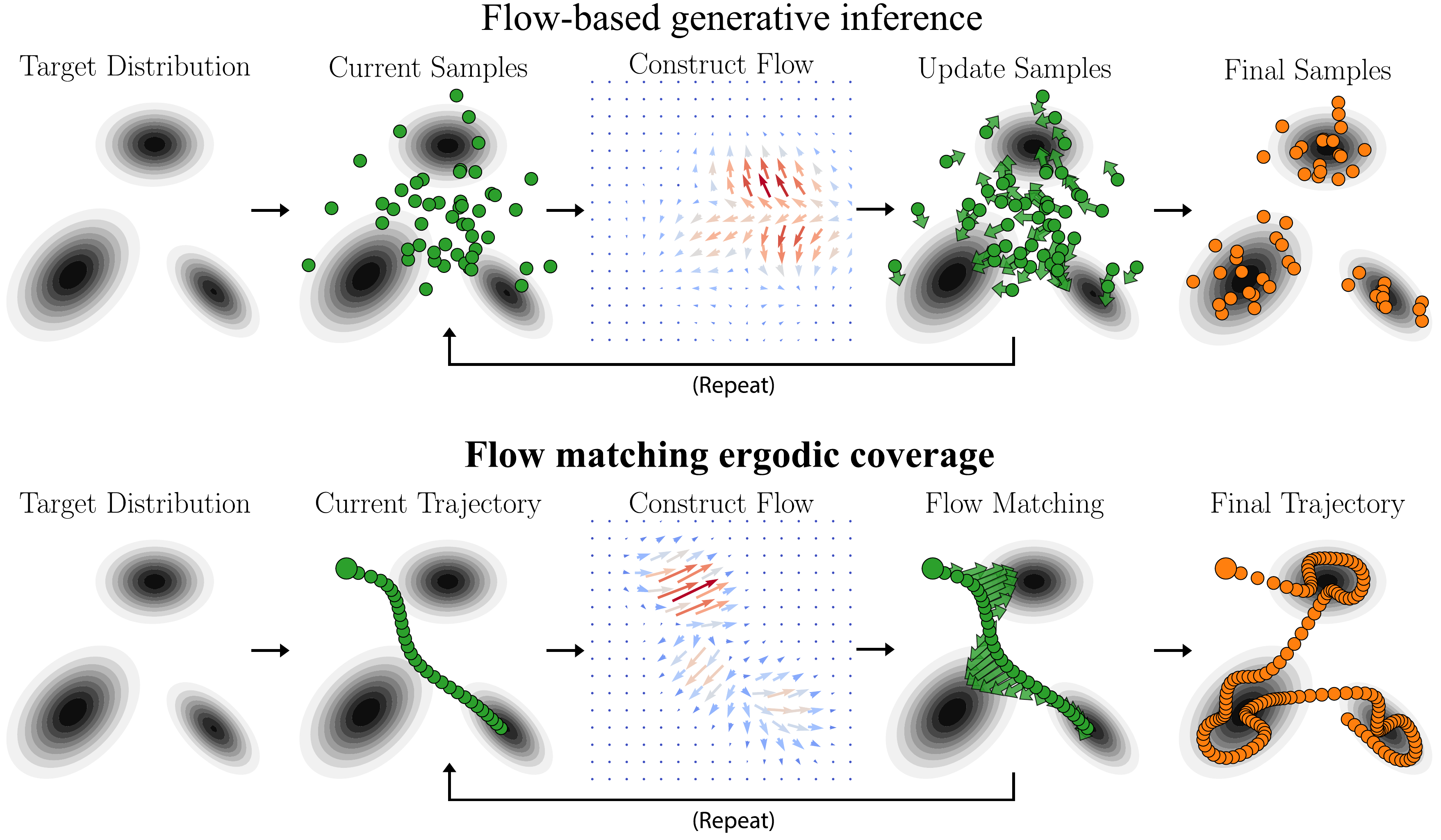}
    \caption{The iterative LQR process is shown to be equivalent to the flow matching~\citep{lipman_flow_2022} framework in generative models subject to dynamics constraints~\citep{sun_flow_2025}.}
    \label{fig:flow_ergodic_overview}
    \vspace{-1em}
\end{figure}

Note that the LQR problem in (\ref{eq:standard_lqr}) is not the only formulation for applying iLQR to ergodic control. The iterative LQR formulation above can also be interpreted from the perspective of flow matching, a method widely used for generative inference~\citep{lipman_flow_2023} (see Figure~\ref{fig:flow_ergodic_overview} for overview). To introduce this connection, we first define a \emph{trajectory path} as $s(\tau,t)$, in which there are two temporal variables: the \emph{system time} $t\in[0,T]$ is associated with the dynamic system, while the \emph{flow time} $\tau$ is associated with the evolution of the entire trajectory during optimization. In other words, $\tau$ is a continuous generalization of the discrete iteration number in iterative optimization. Similar to the definition of trajectory empirical distribution (\ref{eq:emp_distr}), the trajectory path induces an empirical distribution path:
\begin{align} 
    p_{[s]}(\tau,x)=\frac{1}{T}\int_0^T\delta(x-s(\tau,t))dt.
\end{align} Similar to a \emph{probability density path} in flow-based sampling~\citep{liu_stein_2016, lipman_flow_2023}, the evolution of $p_{[s]}(\tau,x)$ alongside $\tau$ is governed by the Fokker-Planck equation:
\begin{align}
    \frac{d}{d\tau}p_{[s]}(\tau,x)=\nabla\cdot\left(p_{[s]}(\tau,x)z(\tau,x)\right),
\end{align} where $z(\tau,x) = \frac{d}{d\tau} p_{[s]}(\tau,x)$ is the flow vector field of the empirical distribution path. More specifically, since $p_{[s]}(\tau,x)$ is an empirical distribution path over the system trajectory path $s(\tau,t)$, the flow vector field evaluated along the trajectory satisfies
\begin{align} \frac{d}{d\tau}s(\tau,t)=z(\tau,s(\tau,t))\triangleq \tilde{z}(\tau,t), \end{align}
where $\tilde{z}(\tau,t)$ describes the change of the system state at system time $t$ as the entire trajectory evolves along the flow time $\tau$. 

Similarly, we define a control sequence path $u(\tau,t)$ associated with the system trajectory path $s(\tau,t)$, and define its flow along $\tau$ as
\begin{align} \frac{d}{d\tau}u(\tau,t)=v(\tau,t), \end{align}
where $v(\tau,t)$ is the \emph{control sequence flow}. At any fixed flow time $\tau$, the system trajectory and control sequence are constrained by the original system dynamics $\dot{s}(\tau,t)=f(s(\tau,t),u(\tau,t))$. The relationship between the empirical distribution flow $\tilde{z}(\tau,t)$ and the control sequence flow $v(\tau,t)$ can then be obtained by differentiating the system dynamics with respect to the flow time $\tau$. This gives
\begin{align} \dot{\tilde{z}}(\tau,t)=A(\tau,t)\tilde{z}(\tau,t)+B(\tau,t)v(\tau,t),\qquad \tilde{z}(\tau,0)=0, \end{align}
where
\begin{align} A(\tau,t)=\nabla_s f(s(\tau,t),u(\tau,t)),\qquad B(\tau,t)=\nabla_u f(s(\tau,t),u(\tau,t)). \end{align}
Therefore, at any fixed flow time $\tau$, the empirical distribution flow evaluated along the system trajectory and the control sequence flow are governed by a time-varying linear system corresponding to the local linearization of the original dynamics. This is the same linearized dynamics used in iLQR (\ref{eq:linearized_dynamics}), but now interpreted as describing how the entire system trajectory evolves along the flow time $\tau$.

In standard flow matching~\citep{lipman_flow_2023}, given a reference flow vector field $h(\tau,x)$ that generates a probability density path toward the target distribution $q(x)$, a parameterized flow vector field $g(\tau,x;\theta)$ is optimized as:
\begin{align} 
    \theta^*=\arg\min_\theta \mathbb{E}_{p(\tau,x)}\left[\left\|h(\tau,x)-g(\tau,x;\theta)\right\|^2\right]. 
\end{align} In other words, instead of directly minimizing a discrepancy between the current and target distributions, flow matching optimizes a flow vector field to match a reference flow toward the target distribution.

For ergodic control, however, the state flow $z(\tau,x)$ cannot be directly optimized, since the evolution of the system trajectory is constrained by the robot dynamics. Instead, we optimize the control sequence flow $v(\tau,t)$, which induces the state flow $\tilde{z}(\tau,t)=z(\tau,s(\tau,t))$ through the linearized dynamics derived above. Therefore, the corresponding flow matching problem is:
\begin{align} 
    v^*(\tau,\cdot)=\arg\min_{v(\tau,\cdot)}\mathbb{E}_{p_{[s]}(\tau,x)}\left[\left\|h(\tau,x)-z(\tau,x)\right\|_Q^2\right]+\int_0^T\|v(\tau,t)\|_R^2dt,
\end{align} where the optimization variable is $v(\tau,t)$, while $z(\tau,x)$ is implicitly determined by the system dynamics.

Using the definition of the trajectory empirical distribution and the inner product property of Dirac delta function~\citep{sun_flow_2025}, the first term can be written as:
\begin{align} 
    \mathbb{E}_{p_{[s]}(\tau,x)}\left[\left\|h(\tau,x)-z(\tau,x)\right\|_Q^2\right]=\frac{1}{T}\int_0^T\left\|\tilde{h}(\tau,t)-\tilde{z}(\tau,t)\right\|_Q^2dt, 
\end{align} where $\tilde{h}(\tau,t)=h(\tau,s(\tau,t))$. Therefore, at any fixed flow time $\tau$, together with the linear dynamics of $\tilde{z}(\tau,t)$ derived above, flow matching for ergodic control has the linear quadratic form:
\begin{gather} 
    v(\cdot)^*=\arg\min_{v(\cdot)}\int_0^T\|\tilde{z}(t)-\tilde{h}(t)\|_Q^2+\|v(t)\|_R^2dt,\qquad \\
    \text{s.t.}\quad \dot{\tilde{z}}(t)=A(t)\tilde{z}(t)+B(t)v(t),\quad \tilde{z}(0)=0. 
\end{gather} This formula has the same LQR structure as the problem solved at each iteration of standard iLQR, with the reference flow evaluated along the current trajectory playing the role of the desired state-trajectory perturbation. Therefore, the iterative LQR formulation can be interpreted as flow matching subject to the dynamic constraints of the robot, and the same techniques used to solve the LQR subproblem in iLQR, such as the Riccati equation, can be applied to solve the flow matching ergodic control problem.

Several choices of the reference flow $\tilde{h}(\tau,t)=h(\tau,s(\tau,t))$ can be used depending on the discrepancy measure of interest~\citep{sun_flow_2025}. For example, the Fourier ergodic metric induces a reference flow that drives the trajectory empirical distribution toward matching the target Fourier coefficients. The Stein variational gradient flow can be used when the discrepancy is based on KL divergence, with the reference flow determined by the score function of the target distribution. The Sinkhorn divergence provides another reference flow based on optimal transport, which is particularly useful for target distributions with irregular or non-smooth supports. Importantly, these different reference flows can all be incorporated into the same LQR-based flow matching formulation without changing the underlying control optimization structure.

\subsection{Spectral Multiscale Coverage (SMC)}

\subsubsection{Overview}

The spectral multiscale coverage (SMC) algorithm~\citep{mathew_metrics_2011} is a specialized control optimization method specifically for the Fourier ergodic metric and for first-order and second-order integrator systems. SMC is a reactive control optimization methods that directly leads to a closed-loop feedback law. 

\subsubsection{Formulation}

At a given time $t$, the cost function $C(t+\Delta t)$ for SMC is the first-order time derivative of the Fourier ergodic metric at the end of the time horizon $[t, t+\Delta t]$:
\begin{align}
    C(t+\Delta t) & = \frac{d}{d\tau} \left[ \frac{1}{2} \sum_{\mathbf{k}\in\mathcal{K}} \lambda_{\mathbf{k}} \vert \phi_{\mathbf{k}}(\tau) - c_{\mathbf{k}} \vert^2 \right]_{\tau=t+\Delta t}, \nonumber \\
    & = \sum_{\mathbf{k}\in\mathcal{K}} \lambda_{\mathbf{k}} \Big( \phi_{\mathbf{k}}(t+\Delta t) - c_{\mathbf{k}} \Big) f_{\mathbf{k}}(s(t+\Delta t)) \label{eq:smc_first_cost} \\
    \phi_{\mathbf{k}}(t) & = \frac{1}{t} \int_0^{t} f_{\mathbf{k}}(s(\tau)) d\tau , 
\end{align} with the controls subject to the constraint $\Vert u(\cdot) \Vert_2 \leq u_{max}$. For first-order integrator systems $\dot{x}(t)=u(t)$, the optimal control solution at the limit of $\Delta t\rightarrow 0$ is:
\begin{gather}
    u^*(t) = -u_{max} \frac{B(t)}{\Vert B(t) \Vert_2}, \label{eq:smc_first} \\
    B(t) = \sum_{\mathbf{k}\in\mathcal{K}} \lambda_{\mathbf{k}} \Big( \phi_{\mathbf{k}}(s(t)) - c_{\mathbf{k}} \Big) \nabla f_{\mathbf{k}}(s(t)). \label{eq:smc_B_term}
\end{gather} For second-order integrator systems $\ddot{x}(t)=u(t)$, the cost function $C^\prime(t+\Delta t)$ is as follow:
\begin{gather}
    C^\prime(t+\Delta t) = C(t+\Delta t) + \frac{c}{2} \int_t^{t+\Delta t} \Vert v(\tau)\Vert^2 d\tau, 
\end{gather} where $c$ is the regulation weight parameter for the kinetic energy cost and $v(t)$ is the velocity of the system at time $t$. The optimal control solution at the limit of $\Delta t\rightarrow 0$ is:
\begin{align}
    u^*(t) = -u_{max} \frac{c\cdot v(t) + B(t)}{\Vert c\cdot v(t) + B(t) \Vert_2}, \label{eq:smc_second}
\end{align} where $B(t)$ is the same as in (\ref{eq:smc_B_term}). 

In practice, $\phi_{\mathbf{k}}(t)$ can be updated recursively without recomputing the full integral:
\begin{align}
\phi_{\mathbf{k}}(t+\Delta t) \approx \frac{t \cdot \phi_{\mathbf{k}}(t)+\Delta t \cdot f_{\mathbf{k}}(s(t+\Delta t))} {t+\Delta t}.
\end{align} Thus, the computational cost is constant with respect to the elapsed time and linear in the number of Fourier coefficients. Since SMC directly computes the control from the current state and accumulated coverage statistics, it avoids full-horizon trajectory optimization and is well suited for real-time closed-loop control.

\subsubsection{Interpretation and Extension}

An alternative interpretation of the SMC algorithm is that, the optimal control solution minimizes the second-order time derivative of the Fourier ergodic metric at a given time $t$\footnote{Recall that, the cost function for the SMC algorithm (\ref{eq:smc_first_cost}) is the first-order time derivative \emph{at the end} of the planning horizon, and the optimal control is derived as the limited with an infinitesimally small horizon $\Delta t$.}. We start with the first-order time derivative:
\begin{align}
    & \frac{d}{dt} \left[ \frac{1}{2} \sum_{\mathbf{k}\in\mathcal{K}} \lambda_{\mathbf{k}} \vert \phi_{\mathbf{k}}(t) - c_{\mathbf{k}} \vert^2 \right] = \sum_{\mathbf{k}\in\mathcal{K}} \lambda_{\mathbf{k}} \Big( \phi_{\mathbf{k}}(t) - c_{\mathbf{k}} \Big) \dot{\phi}_{\mathbf{k}}(t), \nonumber
\end{align} where
\begin{align}
    \dot{\phi}_{\mathbf{k}}(t) & = \frac{d}{dt} \left[ \frac{1}{t} \int_0^t f_{\mathbf{k}}(s(\tau)) d\tau \right] = -\frac{1}{t} \phi_{\mathbf{k}}(t) + \frac{1}{t} f_{\mathbf{k}}(s(t)). \nonumber
\end{align} The second-order time derivative can be derived as:
\begin{align}
    \frac{d^2}{dt^2} \left[ \frac{1}{2} \sum_{\mathbf{k}\in\mathcal{K}} \lambda_{\mathbf{k}} \vert \phi_{\mathbf{k}}(t) - c_{\mathbf{k}} \vert^2 \right] & = \frac{d}{dt} \left[ \sum_{\mathbf{k}\in\mathcal{K}} \lambda_{\mathbf{k}} \Big( \phi_{\mathbf{k}}(t) - c_{\mathbf{k}} \Big) \dot{\phi}_{\mathbf{k}}(t) \right] \nonumber \\
    & = \sum_{\mathbf{k}\in\mathcal{K}} \lambda_{\mathbf{k}} \dot{\phi}_{\mathbf{k}}(t)^2 + \sum_{\mathbf{k}\in\mathcal{K}} \lambda_{\mathbf{k}} \Big( \phi_{\mathbf{k}}(t) - c_{\mathbf{k}} \Big) \ddot{\phi}_{\mathbf{k}}(t), \nonumber
\end{align} where
\begin{align}
    \ddot{\phi}_{\mathbf{k}}(t) & = \frac{d}{dt} \left[ -\frac{1}{t} \phi_{\mathbf{k}}(t) + \frac{1}{t} f_{\mathbf{k}}(s(t)) \right] \nonumber \\
    & = \frac{1}{t^2} \phi_{\mathbf{k}}(t) - \frac{1}{t} \dot{\phi}_{\mathbf{k}}(t) - \frac{1}{t^2} f_{\mathbf{k}}(s(t)) + \frac{1}{t} \nabla f_{\mathbf{k}}(s(t)) \dot{s}(t). \nonumber
\end{align} The only term in the second-order time derivative that depends on the system dynamics $\dot{x}(t)$ is:
\begin{align}
    \sum_{\mathbf{k}\in\mathcal{K}} \lambda_{\mathbf{k}} \Big( \phi_{\mathbf{k}}(t) - c_{\mathbf{k}} \Big) \cdot \left( \frac{1}{t} \nabla f_{\mathbf{k}}(s(t)) \dot{s}(t) \right) = \frac{1}{t} B(t) \dot{s}(t). \nonumber
\end{align} Therefore, the optimal control in the SMC algorithm minimizes the second-order time derivative of the Fourier ergodic metric at any given time subject to the control constraints. 

More importantly, following this interpretation, we can extend the derivative of the SMC algorithm to other ergodic metrics, such as the kernel ergodic metric. Denote the kernel ergodic metric as a function of time $\mathcal{E}(t)$:
\begin{align}
    \mathcal{E}(t) = \frac{1}{t^2} \int_0^t \int_0^t K\Big( s(\tau), s(\tau^\prime) \Big) d\tau d\tau^\prime - \frac{2}{t} \int_0^t q(s(\tau)) d\tau,  \nonumber
\end{align} the first-order time derivative can be derived as:
\begin{align}
    \dot{\mathcal{E}}(t) & = -\frac{2}{t^3} \int_0^t \int_0^t K\Big( s(\tau), s(\tau^\prime) \Big) d\tau d\tau^\prime + \frac{2}{t^2} \int_0^t K(s(t), s(\tau)) d\tau \nonumber \\
    & \quad + \frac{2}{t^2} \int_0^t q(s(\tau)) d\tau - \frac{2}{t} q(s(t)).
\end{align} In the second-order time derivative, the only terms that depend on the system dynamics would be:
\begin{align}
    & \frac{2}{t^2} \int_0^t \nabla_1 K(s(t), s(\tau)) \dot{s}(t) d\tau - \frac{2}{t} \nabla q(s(t)) \dot{s}(t) \nonumber \\
    & = \frac{2}{t} \left[ \frac{1}{t} \int_0^t \nabla_1 K(s(t), s(\tau)) d\tau - \nabla q(s(t)) \right] \dot{s}(t). 
\end{align} Therefore, we can write down the optimal control solutions for the kernel ergodic metric for first- and second-order integrator systems as in (\ref{eq:smc_first}) and (\ref{eq:smc_second}) for the SMC method, with the only difference being the $B(t)$ formulation:
\begin{align}
    B(t) = \frac{1}{t} \int_0^t \nabla_1 K(s(t), s(\tau)) d\tau - \nabla q(s(t)).
\end{align}

\subsection{Heat Equation-Driven Area Coverage (HEDAC)}

\subsubsection{Overview}
The Heat Equation-Driven Area Coverage (HEDAC) algorithm~\citep{ivic_ergodicity-based_2017} is a closed-loop control framework that minimizes a smoothed spatial coverage error by solving a coupled system of differential equations. Similarly to SMC, HEDAC offers a spectral and multiscale coverage strategy. However, it is not limited to Euclidean domains with rectangular boundaries, allows the multiscale behavior to be modulated via a single parameter, and avoids the emergence of spurious repulsive artifacts

\subsubsection{Formulation}
HEDAC aims to minimize a spatial coverage error over time to achieve ergodicity:
\begin{align}
\lim_{t \rightarrow \infty} \|e(x, t)\|_2 = 0.
\end{align}
However, as mentioned with the kernel ergodic metric, the spatial coverage error cannot be directly expressed as the $L^2$ norm between the target density $q(x)$ and the empirical trajectory distribution $p_s(x)$, since these distributions are not in $L^2(\Omega)$, and their difference is not square-integrable. To address this, HEDAC smooths both distributions using a kernel function $K_\sigma$ typically chosen as a Gaussian radial basis function (RBF) kernel
\begin{align}
    K_\sigma\left(x-x^{\prime}\right)=\frac{1}{\left(2 \pi \sigma^2\right)^{n / 2}} \exp \left(-\frac{\left\|x-x^{\prime}\right\|^2}{2 \sigma^2}\right)
\end{align}
The spatial coverage error is then written as:
\begin{align}
e(x, t) = \left(K_\sigma * q\right)(x) - c_\sigma(x, t),
\end{align}
where
\begin{align}
\left(K_\sigma * q\right)(x) = \int_\Omega K_\sigma(x - x') q(x') \, dx'
\end{align}
is the convolution of the target density with the smoothing kernel and the smoothed empirical coverage is defined as:
\begin{align}
c_\sigma(x, t) = \frac{1}{N t} \sum_{i=1}^N \int_0^t K_\sigma\left(x - s_i(\tau)\right) \, d\tau.
\end{align}

HEDAC minimizes this smoothed coverage error iteratively by solving a coupled system consisting of:
\begin{itemize}
\item \textbf{Ordinary differential equations (ODEs)} modeling the dynamics of the coverage agents, and
\item A \textbf{partial differential equation (PDE)} governing the evolution of a potential field that encodes the smoothened spatial coverage error.
\end{itemize}
In the simplest case, each agent follows first-order integrator dynamics:
\begin{align}
\dot{s}_i(t) = -\nabla \psi(s_i(t), t),
\end{align}
where the control input is given by the negative gradient of a potential field $\psi(x, t)$. This potential encodes the smoothed coverage error and is governed by a second-order elliptic PDE:
\begin{align}
k \Delta \psi(x, t) - \psi(x, t) + \mu(x, t) = 0,
\label{eq:pde}
\end{align}
where $k$ is the diffusion coefficient adjusting the global/local coverage behavior, $\Delta$ is the Laplacian operator and $\mu(x, t)$ is a non-negative source term representing the local mismatch between target and empirical coverage. Two common choices for the source term $\mu(x, t)$ are:

1. Squared ReLU of the coverage error:
\begin{align}
\mu(x, t) = \max(e(x, t), 0)^2,
\end{align}
which suppresses spurious repulsive effects in over-covered regions.

2. An exponential decay formulation:
\begin{align}
\mu(x, t) = q(x) \exp(-c_\sigma(x, t)),
\end{align}
which removes the need for normalization. Both variants ensure that agents continue to be attracted to under-covered regions without being repelled from over-covered ones, addressing a key limitation of spectral multiscale coverage (SMC), which can produce spurious repulsive forces when $e(x, t) < 0$.

The PDE in \eqref{eq:pde} is solved in the interior of the domain $\Omega$, and its boundary behavior is governed by a zero-Neumann condition:
\begin{align}
\mathbf{n} \nabla \psi(x, t) = 0 \quad \forall x \in \partial \Omega,
\label{eq:hedac_bc}
\end{align}
where $\mathbf{n}$ is the outward unit normal vector to the boundary $\partial \Omega$. This boundary condition ensures that the potential field does not induce motion across the domain boundaries. Notably, SMC also enforces a zero-Neumann boundary condition by employing a cosine Fourier basis, which prevents agents from leaving the domain but inherently restricts the method to rectangular Euclidean boundaries. The connection between HEDAC and SMC becomes more evident from a spectral viewpoint: the Fourier basis functions used in SMC are eigenfunctions of the Laplacian operator on Euclidean domains with rectangular boundary. We refer the reader to \citet{bilaloglu_tactile_2025} for a detailed discussion of this connection.

\subsubsection{Properties and advanced techniques}

\subsubsection{Finite difference implementation} 

Probably the simplest way to solve \eqref{eq:pde} is using the Finite Difference Method (FDM). For the domain discretization using a rectilinear numerical grid, we can define the $d$-dimensional discrete Laplacian
\begin{align}
\Delta_h \psi_{\mathbf{i}} = \sum_{d=1}^{n} \frac{\psi_{\mathbf{i}-\mathbf{e}_d} - 2 \psi_{\mathbf{i}} + \psi_{\mathbf{i}+\mathbf{e}_d}}{h_d^2}
\end{align}
where $\mathbf{i}$ is d-dimensional vector of node indices, and $\mathbf{e}_d$ is single cell offset in direction of dimension $d$. The partial differential equation \eqref{eq:pde} can be approximated in each internal node $\mathbf{i}$ with linear algebraic equation
\begin{align}
k \, \Delta_h \psi_{\mathbf{i}} - \psi_{\mathbf{i}} = -\mu_{\mathbf{i}}.
\label{eq:fdm_stencil}
\end{align}
For domain boundary nodes, the zero-Neumann condition is implemented with
\begin{align}
\dfrac{\psi_{\mathbf{i}} - \psi_{\mathbf{i}-\mathbf{n}_d}}{\mathbf{n}_d}
 = 0
\label{eq:fdm_neumann_bc}
\end{align}
where $\mathbf{n}_d$ is the unit offset in the direction of outward normal to the domain boundary. Note that, although written as in vector form, the equation applies only to the dimension $d$ corresponding to the direction of normal, hence $\mathbf{n}_d$ is interpreted as a scalar equal to $\pm 1$. 

Combining linear equations \eqref{eq:fdm_stencil} and \eqref{eq:fdm_neumann_bc}, for all internal and boundary nodes respectively, yields a system of linear equations $\mathbf{A}\mathbf{\Psi}=\mathbf{b}$. 
The system matrix $\mathbf{A}$ is unchanging, since it represents the topology of the numerical grid and the finite difference approximation of the PDE and boundary condition. However, the LHS vector $\mathbf{b}$ updates according to $\mu$ alterations.  Hence, in a control time loop, the solution vector $\mathbf{\Psi}$ only depends on the updated vector $\mathbf{b}$.
This property allows for a substantial speed-up by solving $\mathbf{A}^{-1}$ once in the initialization and later obtaining $\mathbf{\Psi}=\mathbf{A}^{-1}\mathbf{b}$. However, using the explicit inversion is still inefficient, but can be improved by using matrix factorization such as LU decomposition ($\mathbf{A}=\mathbf{L}\mathbf{U}$). With two triangular matrices, a forward substitution can be used to solve for $\mathbf{y}$:
\begin{align}
\mathbf{L}\mathbf{y}=\mathbf{b}
\end{align}
and then backward substitution for $\mathbf{\Psi}$:
\begin{align}
\mathbf{U}\mathbf{\Psi}=\mathbf{y}.
\end{align}
Both of these substitution procedures cost $\mathcal{O}(n^2)$ for dense matrices $\mathbf{A}$. Since the matrix $\mathbf{A}$ is sparse, using frameworks for sparse linear algebra can further improve computational efficiency.

\subsubsection{Finite element implementation}

Formulating the ergodic control problem using PDEs allows for the use of irregularly shaped domains as well as the definition of inner boundaries within the domain. As with the outer boundary, a zero-Neumann condition is applied to all inner boundaries; thus, the boundary condition \eqref{eq:hedac_bc} remains valid, with the note that $\partial \Omega$ now includes both the outer and any inner domain boundaries. Since this boundary condition prevents motion across boundaries, cut-out regions of the domain—the holes—naturally act as obstacles that trajectories avoid. Although this behavior arises intrinsically from the combination of the PDE boundary condition and the gradient-based motion law, it necessitates PDE solving techniques more advanced than the FDM. Fortunately, a wide range of robust and efficient alternatives exists, many of which have been well-established and refined in engineering applications, such as computational fluid dynamics and computational structural analysis.

\citet{ivic_constrained_2022} justify the use of Finite Element Method (FEM) for solving \eqref{eq:pde}, due to its inherent properties to provide interpolation of the solution $\psi(x)$ as well as the gradient $\nabla\psi(x)$.

FEM requires the weak formulation of PDE, which is obtained by multiplying the PDE by a smooth test function $v\in H^1(\Omega)$ and integrating over the domain. Using the integration by parts, the following equation is acquired:
\begin{align}
k \int_{\Omega} \nabla \psi \cdot \nabla v \, dx 
+ \int_{\Omega} \psi \, v \, dx
= \int_{\Omega} \mu \, v \, dx 
+ k \int_{\partial \Omega} \frac{\partial \psi}{\partial n} \, v \, ds,
\quad \forall v \in H_1(\Omega).
\label{eq:weak_formulation_general}
\end{align}
For homogeneous Neumann boundary conditions (zero-Neumann condition on all boundaries) the last term in \eqref{eq:weak_formulation_general} vanishes and weak form is simplified to 
\begin{align}
k \int_{\Omega} \nabla \psi \cdot \nabla v \, dx
+ \int_{\Omega} \psi\, v \, dx
= \int_{\Omega} \mu\, v \, dx,
\quad \forall v \in H_1(\Omega).
\label{eq:weak_neumann}
\end{align}

Let $\Omega \subset \mathbb{R}^d$ be a bounded domain with boundary $\partial \Omega$, and let $\mathcal{T}_h$ be a triangulation of $\Omega$. 
Define the finite-dimensional subspace
\begin{align}
V_h := \operatorname{span} \{\varphi_1, \varphi_2, \dots, \varphi_N\} \subset H^1(\Omega),
\end{align}
where $\{\varphi_j\}$ are standard nodal basis functions.

The finite element approximation $\psi_h \in V_h$ of the solution $\psi$ of \eqref{eq:pde} is defined by
\begin{align}
\psi_h(x) = \sum_{j=1}^{N} \Psi_j \, \varphi_j(x),
\end{align}
where the coefficients $\Psi_j$ satisfy the discrete weak form:
\begin{align}
k \int_{\Omega} \nabla \psi_h \cdot \nabla \varphi_i \, dx 
+ \int_{\Omega} \psi_h \, \varphi_i \, dx 
= \int_{\Omega} \mu \, \varphi_i \, dx,
\qquad i = 1, \dots, N.
\label{eq:weak_discrete}
\end{align}

The resulting system of linear equations can be written in matrix form as
\begin{align}
(K + M) \, \Psi = F,
\label{eq:fem_lin_sys}
\end{align}
where
\begin{align*}
K_{ij} = k \int_{\Omega} \nabla \varphi_j \cdot \nabla \varphi_i \, dx, 
\quad 
M_{ij} = \int_{\Omega} \varphi_j \, \varphi_i \, dx, 
\quad 
F_i = \int_{\Omega} \mu \, \varphi_i \, dx,
\end{align*}
and $\Psi = (\Psi_1, \dots, \Psi_N)^T$ is the vector of unknown nodal values. FEM-derived linear system \eqref{eq:fem_lin_sys} allows for the same speedups as for solving FDM's linear system. 

\begin{figure}[t]
    \centering
    \includegraphics[width=0.4\linewidth]{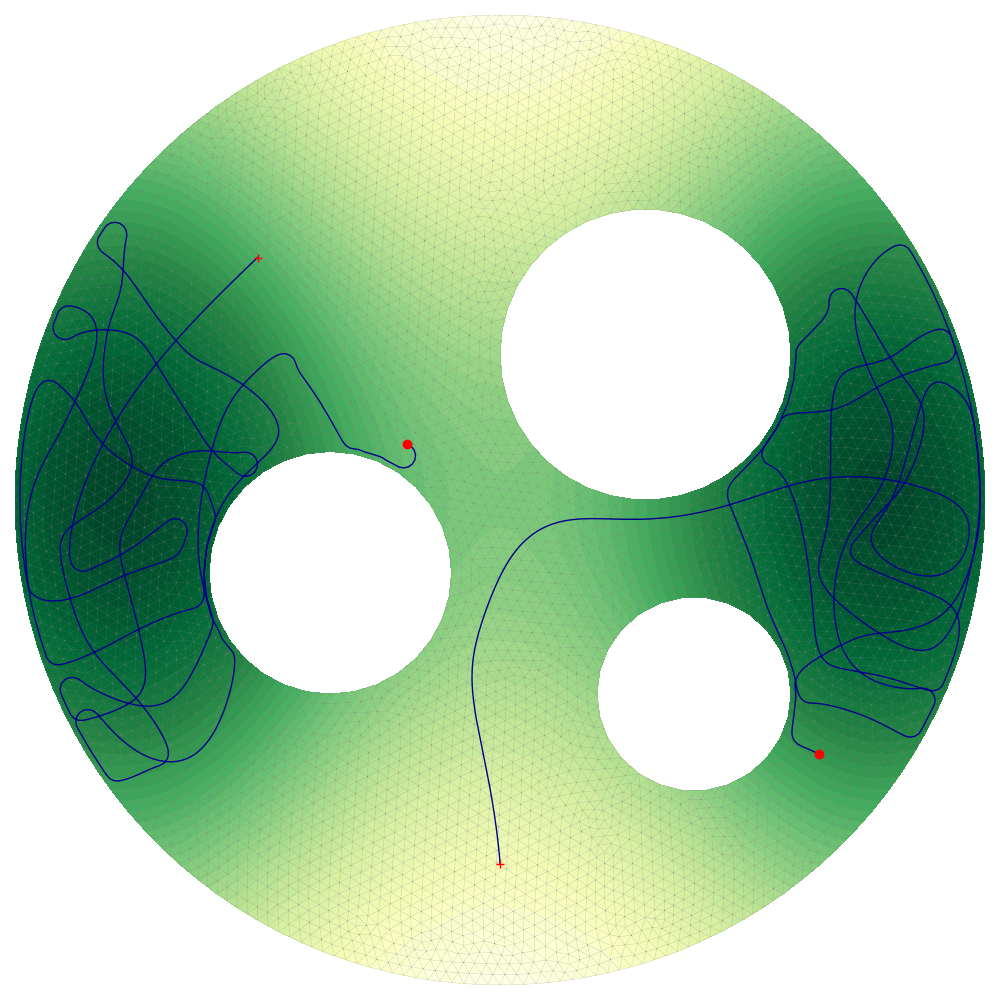}
    \includegraphics[width=0.59\linewidth]{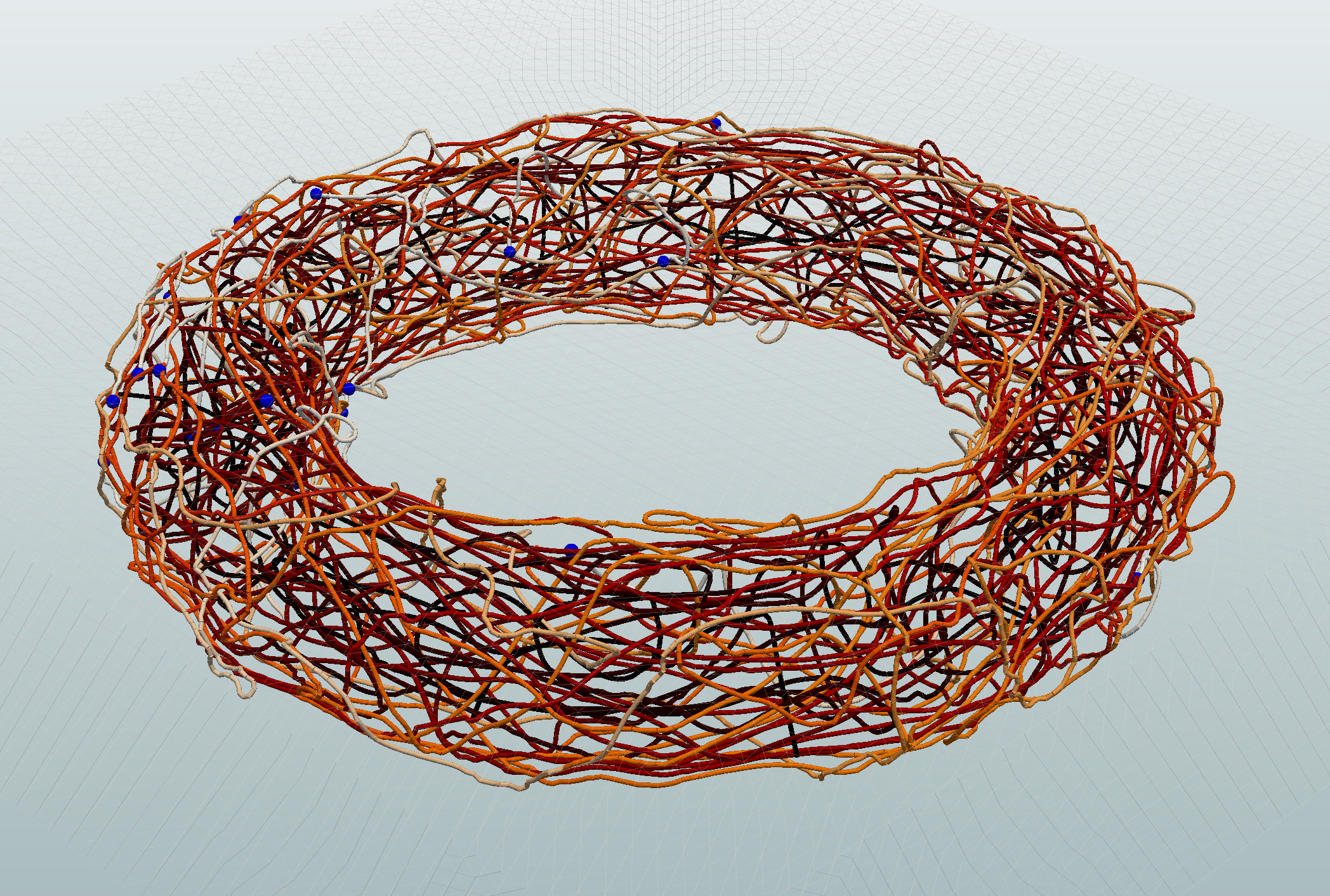}
    \caption{(a) Example of ergodic exploration using HEDAC FEM method presented in \citet{ivic_constrained_2022}. The triangular discretization allows for curved outer and inner boundaries of the domain.
    (b) 3D trajectories filling a torus with HEDAC FEM implementation from \citet{ivic_multi-uav_2023}.}
    \label{fig:hedac_fem}
\end{figure}

Demonstration of FEM implementation is given in Figure~\ref{fig:hedac_fem}:
an example of two agents exploring a circular domain with three obstacles using HEDAC FEM 2D implementation, and covering a 3D body using HEDAC FEM 3D.

\subsection{Choosing a Control Optimization Method}

\noindent\textbf{Nonlinear programming (NLP).}
NLP is generally the most stable and flexible choice. It explicitly formulates the dynamics and other constraints (e.g., obstacle avoidance), is compatible with most ergodic metrics. It also enables alternative formulations in which ergodicity is imposed as a constraint while another objective is optimized, such as execution time~\citep{dong_time-optimal_2024} and energy consumption~\citep{seewald_energy-aware_2024}. Compared with iLQR, NLP is generally better suited to highly nonlinear systems and is less structurally dependent on smooth, differentiable dynamics, although this depends on the chosen solver. Its main disadvantages are high computational cost, sensitivity to initialization, and susceptibility to local optima, making it more suitable for offline computation or relatively slow replanning.

\noindent\textbf{Iterative linear quadratic regulator (iLQR).}
Like NLP, iLQR solves a full-horizon trajectory optimization problem, but the dynamics are incorporated implicitly through trajectory rollout rather than imposed as explicit constraints. This maintains dynamic feasibility with respect to the modeled system and generally provides greater computational efficiency, since each local LQR problem can be solved through a Riccati recursion and the method often requires relatively few iterations, particularly with techniques such as backtracking line search. Its structure also enables specialized integration with certain ergodic metrics, including flow-matching formulations~\citep{sun_flow_2025}. However, iLQR generally requires smooth, differentiable dynamics and objectives, is less effective for highly nonlinear systems, and handles hard constraints less naturally than NLP.

\noindent\textbf{Spectral multiscale coverage (SMC).}
Unlike NLP and iLQR, SMC does not optimize a complete finite-horizon trajectory, but instead generates the trajectory incrementally one time step at a time. It can therefore be orders of magnitude more computationally efficient, making it particularly suitable for fast online control with limited computational resources. The tradeoff is that SMC does not optimize performance over a fixed future horizon and may be suboptimal for tasks with strict time budgets or terminal requirements. It is also restricted to relatively simple dynamics, primarily first- and second-order integrator systems, with additional techniques required for more complex robot dynamics.

\noindent\textbf{Heat Equation-Driven Area Coverage (HEDAC).}
Like SMC, HEDAC incrementally generates controls rather than solving a full-horizon optimization problem, making it suitable for online coverage, including coverage of dynamic target distributions~\citep{lanca_ergodic_2025}. Compared with SMC, it naturally supports irregular domain boundaries and internal obstacles, allows its multiscale behavior to be adjusted through a single parameter, and avoids some spurious repulsive artifacts associated with truncated Fourier representations. However, its computational cost depends on the resolution and dimension of the discretized PDE domain. HEDAC is also most directly compatible with first-order dynamics, and additional planning, tracking, or control techniques are generally required for more complex robot systems~\citep{lanca_probabilistic_2026}.

\section{Problems Related to Ergodic Control}

\subsection{Coverage Motion Planning}

A topic closely related to ergodic control is coverage motion planning~\citep{choset_coverage_2001, galceran_survey_2013}, commonly referred to as coverage path planning. Coverage motion planning considers the problem of synthesizing a path such that the footprint of a robot, sensor, or end effector passes over all points within an area or volume of interest while avoiding obstacles. In addition to achieving complete coverage, the generated path is commonly optimized to minimize costs such as path length, execution time, overlap, number of turns, or energy consumption. 

Another closely related paradigm is Voronoi-based coverage control, which addresses the density-aware spatial deployment and coordination of mobile sensing agents~\citep{du_centroidal_1999, cortes_coverage_2004, lekien_nonuniform_2010}. These methods commonly define a density-weighted locational objective that penalizes the distance between each point in the environment and its nearest agent. The resulting objective can be decomposed over the agents' Voronoi cells, and Lloyd-type control laws move each agent toward the density-weighted centroid of its assigned cell~\citep{du_centroidal_1999, cortes_coverage_2004}. At equilibrium, the agents form a centroidal Voronoi configuration that allocates more sensing resources to regions with greater importance. Extensions have considered unknown importance distributions learned online from sensor measurements~\citep{schwager_decentralized_2009}, time-varying importance distributions~\citep{lekien_nonuniform_2010, lee_multirobot_2015}, and heterogeneous sensing capabilities~\citep{pimenta_sensing_2008, arslan_voronoi-based_2016}.

Ergodic control shares with nonuniform coverage control the use of a spatial distribution to represent the relative importance of different regions. The key distinction is what is matched to this distribution. Voronoi-based coverage control primarily optimizes the instantaneous or steady-state spatial configuration of a team of agents, whereas ergodic control optimizes the time-averaged spatial statistics of their trajectories. Consequently, classical Voronoi-based methods typically drive agents toward a stationary deployment, while ergodic control generally produces persistent motion that repeatedly visits regions in proportion to their importance. Multi-agent coordination in Voronoi-based methods is naturally expressed through spatial partitions and local interactions between neighboring agents, whereas ergodic methods commonly coordinate agents through their aggregate visitation statistics. Voronoi-based methods can be sensitive to initialization and may converge to locally optimal partitions, particularly in nonconvex environments, while extensions may be required to accommodate complex dynamics or rapidly changing importance distributions. Ergodic control avoids permanently assigning agents to fixed regions and directly incorporates system dynamics, but its finite-horizon optimization does not generally guarantee that every point will be visited within a prescribed time.

Coverage path planning differs from both paradigms by using a primarily set-based definition of coverage: a point is considered covered once it falls within the footprint of the robot or its sensor, and the objective is typically to cover every point of the target space at least once within a finite task horizon. It therefore focuses on constructing an explicit path that achieves complete geometric coverage as efficiently as possible. In contrast, both nonuniform coverage control and ergodic control are distribution-aware, but the former distributes agents across space while the latter distributes trajectory time across space. 

\subsection{Density Steering}

Another closely related topic to ergodic control is distribution steering~\citep{chen_optimal_2016, caluya_reflected_2021, chen_optimal_2021}, also referred to as density control. A branch of stochastic control, distribution steering considers the problem of controlling the time evolution of a probability distribution over the system's states. Denote a stochastic dynamic system as:
\begin{align}
    d s_t = f(s_t, u_t) dt + G(s_t) dW_t, 
\end{align} where $G(s_t)$ denotes the diffusion matrix and $W_t$ denotes the Wiener process (standard Brownian motion), the temporal evolution of the state probability distribution $p(t,s)$ is governed by the Fokker-Planck equation:
\begin{align}
    \frac{\partial}{\partial t} p(t, s) = - \nabla_s \cdot (f(s, u) p(t, s)) + \frac{1}{2} \nabla_s^2 : (G G^\top p(t,s)). 
\end{align} The problem of distribution steering seeks a control policy that transports the initial state distribution to a desired state distribution while minimizing the control effort or other additional cost. An important variant of the distribution steering problem is the Schrödinger bridge problem~\citep{leonard_survey_2013}, which synthesizes a stochastic process that leads to the desired state distribution while minimizing the Kullback-Leibler divergence with respect to a reference stochastic process. Aside from stochastic control, the Schrödinger bridge problem plays an important role in optimal transport and generative modeling~\citep{bortoli_diffusion_2021}, where it can enable more efficient sampling from complex data distributions. 

Despite both focusing on synthesizing state-related distributions with dynamics constraints, compared to controlled diffusion and ergodic control, distribution steering focuses on synthesizing the instantaneous state distribution that describes the ensemble of possible system states at a particular time. More importantly, distribution steering assumes a fundamentally stochastic system and focuses on controlling the stochasticity of the system at each timestep, while controlled diffusion does not make such assumptions. By viewing the entire trajectory as a batch of samples, controlled diffusion is compatible with both deterministic and stochastic systems, where the exhibited stochasticity is the outcome of the accumulated state evolution over time instead of the instantaneous uncertainty.

Nevertheless, the two problem classes share similar mathematical and computational machinery. The Liouville and Fokker–Planck equations describe the temporal evolution of probability distributions under deterministic and stochastic dynamics, respectively; KL divergence, entropy regularization, and optimal transport provide tools for comparing and transporting distributions or stochastic path measures. Several methods presented in this review can be interpreted through this broader density-control viewpoint. For example, Stein variational gradient descent (SVGD)~\citep{liu_stein_2016}, which is used for ergodic control, can be viewed as a density transport tool relevant to distribution steering, as it constructs a deterministic particle transport that decreases KL divergence. Wasserstein distance and Sinkhorn divergence~\citep{feydy_interpolating_2019} provide optimal-transport-based tools for comparing distributions in both problem classes, while Schrödinger bridges use closely related entropic transport machinery to control an entire stochastic path distribution. In addition, HEDAC~\citep{ivic_ergodicity-based_2017} employs a diffusion PDE, although its PDE evolves an auxiliary potential field encoding the smoothed coverage error rather than the physical probability density of the robot state.


\chapter{Ergodic Control: Recipes for Practitioners}

This chapter will review different aspects of extending ergodic control in response to different task-specific requirements in practice.
In particular, this chapter will consider ergodic control on systems with physical constraints with performance guarantees. 
Such guarantees consist of safety during ergodic control, robustness (on coverage outcome as a function of uncertainty), time and energy optimality guarantees, and distributed consensus (e.g., multi-agent coverage problems). 

\section{Safety Guarantees for Ergodic Control}

Systems that are subject to ergodic control, e.g., robots, may venture into unsafe or dangerous configurations. 
This is a natural consequence of ergodic control as areas underexplored are penalized more highly, yielding control behaviors that drive the system in unpredictable way. 
This kind of unpredictable behavior is especially true in multi-agent settings where ergodic agents may collide with one another, and even more so in data-generation settings where collected data may require highly dynamic changes in robot motion. 
Thus, this section will define safety and present ways to incorporate safety guarantees into the ergodic control problem. 

\subsection{Introduction to Safety Guarantees} 

We first define a notion of safety for an underlying system. 
Let $s_t \in \mathcal{S} \subseteq \mathbb{R}^n$ be the state of a robot at time $t \in \mathbb{Z}_+$ that is constrained by the continuous-time dynamics $\dot{s} = f_c(s_t, u_t)$ for control input $u_t \in \mathcal{U} \subseteq \mathbb{R}^m$. \footnote{We define the dynamics in continuous time to define a notion of instantaneous safety and then extend the definition of safety to the discrete-time setting.}
Given a continuously differentiable function $h : \mathcal{S} \to \mathbb{R}$, we define the set $\mathbb{S} = \{ s \in \mathcal{S} | h(x) \ge 0\}$ where $\partial \mathbb{S} = 0$ is the boundary of the set. 
Then, one can establish a \emph{safety guarantee} over this set $\mathbb{S}$ if $\forall s \in \mathcal{S}$, 
\begin{equation}
    \frac{d}{dt} h(s)  \ge - \gamma \left( h(s) \right)
\end{equation}
for some $\mathcal{K}_\infty$ function $\gamma$. 
Subject to control, the safety guarantee is defined as 
\begin{equation} \label{eq:cbf}
    \exists u\in\mathcal{U} \text{ s.t. } \frac{d}{dt} h(s) = \nabla_s h(s) \cdot f(s,u)  \ge - \gamma \left( h(s) \right)
\end{equation}
for $\nabla_s h(s) \neq 0$. 
If this condition holds and one finds a control $u$, the function $h$ is then known as a \emph{control barrier function}. 

\begin{figure}[h!]
  \begin{minipage}[c]{.5\columnwidth}
    \caption{\textbf{Filtered Ergodic Control.} Shown is an example ergodic trajectory (orange) for a planar drone-like system covering the darker areas in the background. Obstacles to be avoided are shown in green where the non-occupied space defines the safe set for the drone. The red arrow illustrates the magnitude of the ergodic control vector $u_\mathcal{E}$ while the blue arrow is the closest resulting safe controller that avoids the obstacle.    
    } 
    \label{fig:filtered_ergodic_control}
  \end{minipage}
  \begin{minipage}[c]{.5\columnwidth}
    \centering
    \includegraphics[width=0.75\linewidth]{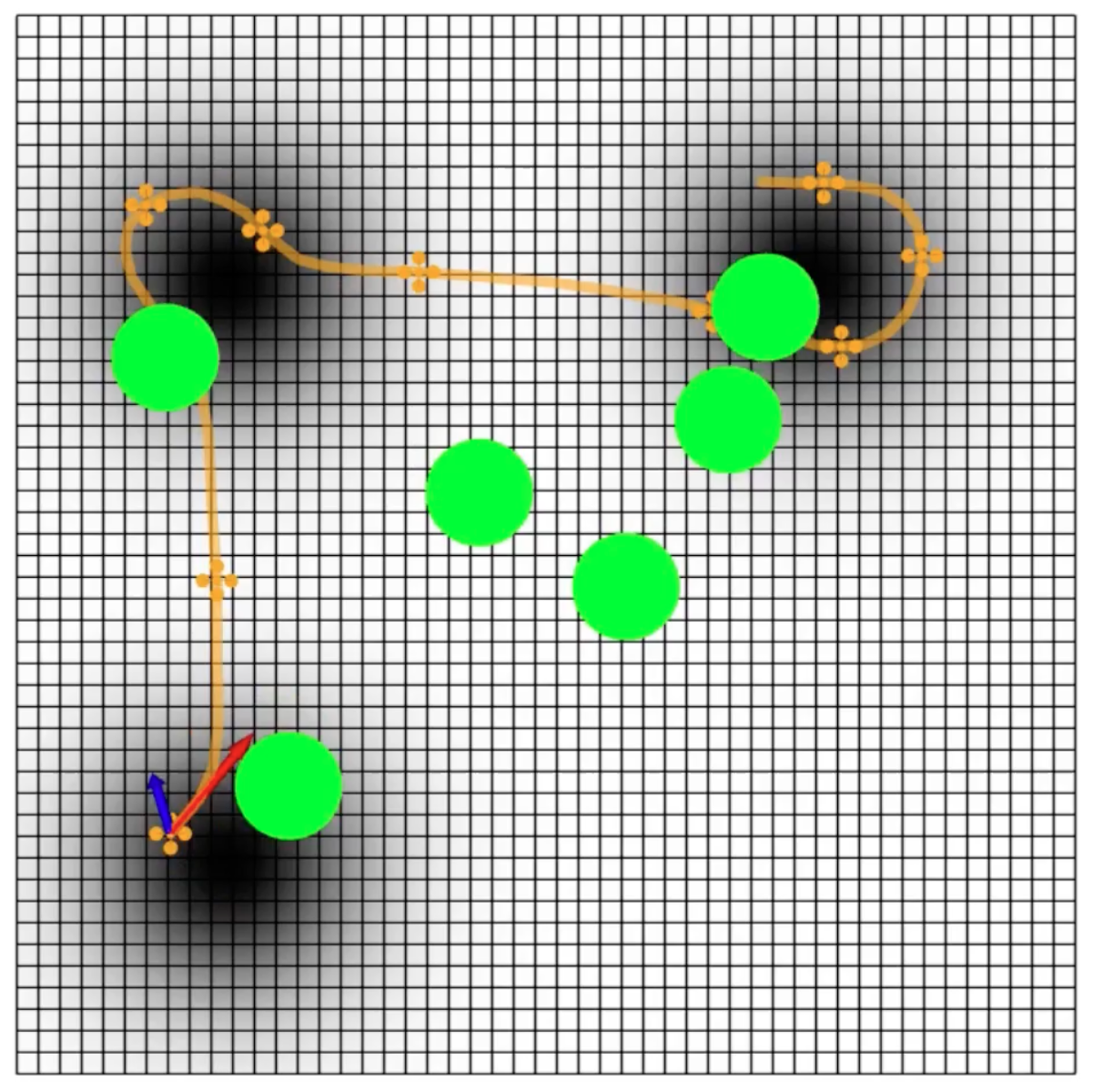}
  \end{minipage}
\end{figure}

In general, satisfying Eq.~\eqref{eq:cbf} can often be accomplished using a safety filter~\citep{ames_control_2017} which searches for the closes control $u$ such that Eq.~\eqref{eq:cbf} is satisfied. 
Specifically, given an ergodic control $u_\mathcal{E}$, the safety-filtered control $u$ is given by the optimization 
\begin{align}
    &\min_u \frac{1}{2}\Vert u - u_\mathcal{E} \Vert_2^2 \\ 
    &\text{subject to } \nabla_s h(s) \cdot f(s, u) \ge - \gamma(h(s)).
\end{align}
One can use the safety filter formulation to filter ergodic controllers as shown in Fig.~\ref{fig:filtered_ergodic_control}.
However, note that following the safety filtered control will lead to worse ergodicity downstream. 
To solve this issue, we include the safety filter constraint equation into the trajectory planning of ergodic control. 

\subsection{Integrated Safety in Ergodic Control} 

In general, safety considerations will impede on effective ergodic coverage. 
As such, safety in ergodic control is best handled in a receding horizon~\citep{mavrommati_real-time_2018}, or trajectory optimization formulation~\citep{miller_trajectory_2013}, where the planner can search for trajectory solutions that are local minima of the ergodic metric. 
Since there exists many local minima with equal levels of ergodicity~\citep{miller_trajectory_2013}, especially at longer time-horizons, the safety guarantees can be maintained throughout. 

\begin{figure}[h!]
  \begin{minipage}[c]{.5\columnwidth}
    \caption{\textbf{Integrated Safety Ergodic Control.} Here, the optimized ergodic trajectory (orange) has the safety barrier constraint satisfied throughout the planning horizon. The blue vector is the commanded ergodic control with integrated safety is that follows the optimized path. Performance of the ergodic controller is preserved while maintaining safety throughout. 
    } 
    \label{fig:integrated_safety_ergodic_control}
  \end{minipage}
  \begin{minipage}[c]{.5\columnwidth}
    \centering
    \includegraphics[width=0.75\linewidth]{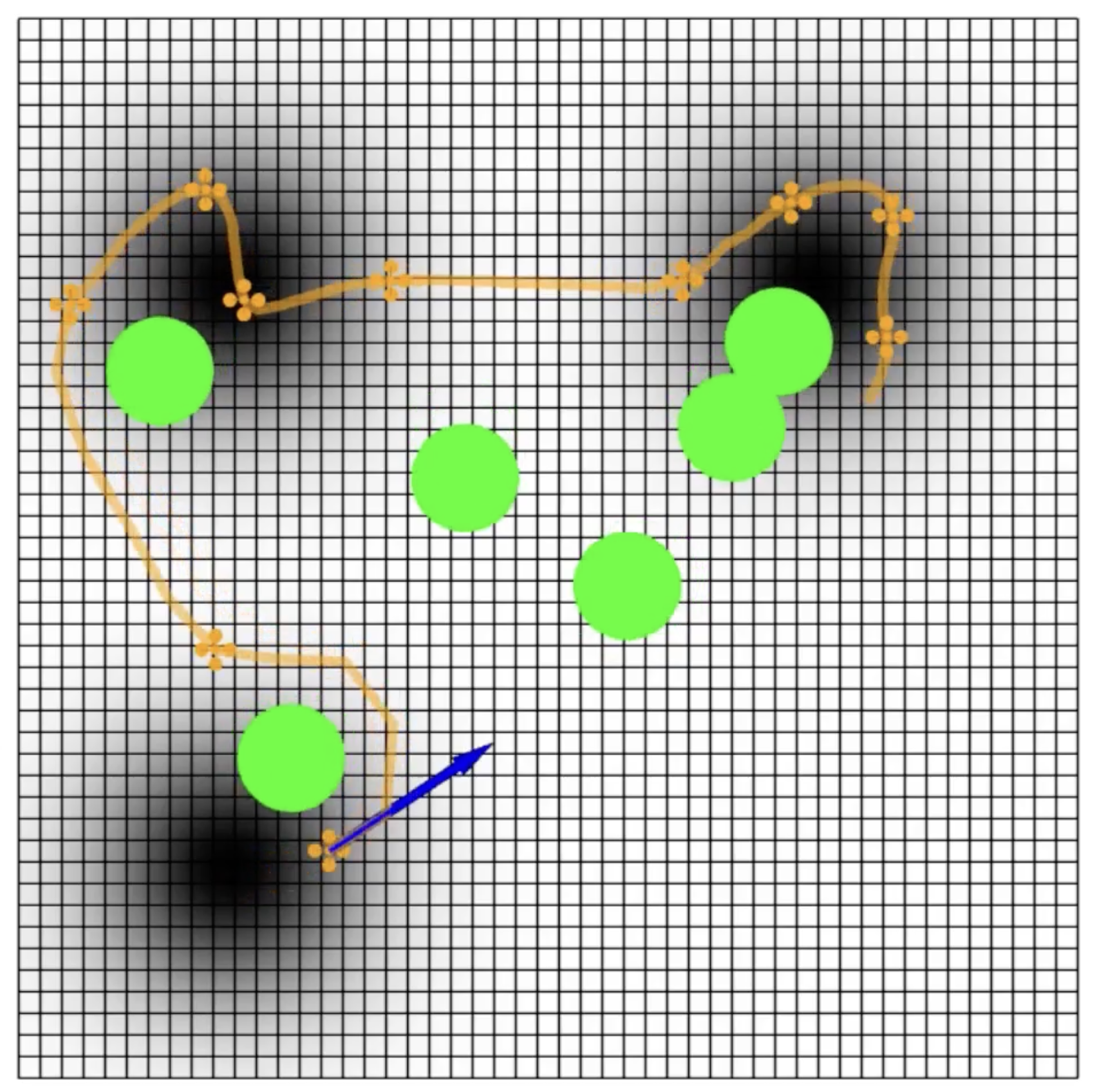}
  \end{minipage}
\end{figure}

We pose the problem of integrated safety in ergodic trajectory optimization via the differential constraints as
\begin{align}
    \begin{split}
    &\min_{\tau} D(p_s, q)\\ 
    &\text{subject to } \\ 
    & \dot{s} = f(s,u), s(t_0) = s_0 \\ 
    &\nabla_s h(s(t)) \cdot f(s(t),u(t)) \ge -\gamma(h(s(t))) \vert_t \forall t\in[t_0, t_f]
    \end{split}
\end{align}
with $\tau = \{s(t)\in \mathcal{S}, u(t) \in \mathcal{U} \forall t \in [t_0,t_f] \}$, and $D(p_s, q)$ is the generic form of the ergodic metric~\ref{def:ergodic_metric}.
In practice, the differential constraints are discretized over $N$ step time length $\Delta t = \frac{t_f-t_0}{N}$ as done in~\citet{lerch_safety-critical_2023} using the differential form of the control barrier function
\begin{equation}
    \Delta h(s_t, u_t) = h(f(s_t, u_t)) - h(s_t) \ge - \gamma(h(s_t))
\end{equation}
where $s_{t+1} = f(s_t, u_t)$ is the overloaded discrete state transition dynamics. 
An example trajectory can be seen in Fig.~\ref{fig:integrated_safety_ergodic_control}. 
This example can be taken further in the multi-agent setting as described in~\citet{lerch_safety-critical_2023} by composing a inter-agent safety constraint between agents and solving a joint optimization for ergodic controls. 

\section{Robustness Guarantees for Ergodic Control}

An open challenge with ergodic control is mitigating external disturbances to ensure \emph{coverage-based} performance guarantees. 
More specifically, systems that experience any form of external disturbance may not achieve effective coverage via ergodic control within a certain time limit. 
In these scenarios, ergodic control is best solved via a recast of the original problem statement where coverage is satisfied according achieve a sub-level of ergodocity, rather than strict minimization. 

\subsection{Hamilton-Jacobi-Isaacs Ergodic Reachability Problem}

Given a dynamical system subject to disturbance $d(t) \in \mathcal{D} \subseteq \mathbb{R}^{n_d}$
\begin{equation} \label{eq:dyn_dist}
    \dot{s}(t) = f(s(t), u(t), d(t)),
\end{equation}
our goal is to find a controller $u_\mathcal{E}$ subject to the worse-case disturbance $d^\star$ that can satisfy the terminal condition
\begin{equation}
    D(p_s, q) \le \varepsilon
\end{equation}
where $p_s : \mathcal{S} \times [0,t_f] \to \mathbb{R}$ takes as argument a \emph{trajectory} subject to dynamics~\eqref{eq:dyn_dist}. 
To solve this problem, we formulate the target set 
\begin{equation}
    \mathcal{L} \equiv \{ s(t) \forall t \in[0,t_f] : D(p_s, q) \le \varepsilon\}
\end{equation}
and the ergodic optimal control problem with disturbances 
\begin{align} \label{eq:ergodic_control_hji}
    \begin{split}
    &\min_{u \in \mathcal{U}} \max_{d \in \mathcal{D}} \hspace{0.5em} h(s(\cdot), t_f) + \int_0^{t_f} u(t)^\top R u(t) dt \\ 
    &\text{subject to } \\
    & \dot{s}(t) = f(s(t), u(t), d(t)), s(0) = s_0 \\ 
    & d(t) = d(s(t), u(t)) \\
    & h(s(\cdot), t_f) = \min(0,D(p_s(s(\cdot)), t_f), q) -\varepsilon) 
    \end{split}
\end{align}
where $R$ is a positive-definite matrix, and the min function satisfies the target set condition $\mathcal{L}$. This optimal control problem is known as the Hamilton-Jacobi-Isaacs (HJI) \citep{bansal_hamilton-jacobi_2020} ergodic reachability problem (\cite{berger_range_2024}). 
The challenge with solving~\eqref{eq:ergodic_control_hji} is that the terminal target set condition takes as argument a trajectory $s(\cdot)$ rather than a terminal state which is not well represented as a value function according to the general form of reachability problem tools \citep{bansal_hamilton-jacobi_2020}. 
This problem can be solved by expanding the ergodic metric into an augmented-state form that encodes continuous state visitation into a differential equation which we describe in the following section. 

\subsection{Augmented-State Formulation}

The HJI ergodic problem can be solved by converting the ergodic metric, specifically the Fourier-based metric \eqref{eq:fourier_metric}, into a differential form. 
\begin{lemma} \citep{de_la_torre_ergodic_2016, berger_range_2024}
    The Fourier-based ergodic metric ~\eqref{eq:fourier_metric} is equivalently defined as 
    \begin{align}
        D(p_s, q) & = \sum_{\mathbf{k}\in\mathcal{K}} \lambda_\mathbf{k}(\phi_\mathbf{k} - c_\mathbf{k})^2 \\ 
        & = \frac{1}{t_f^2} \sum_{\mathbf{k} \in \mathcal{K}}\lambda_\mathbf{k} z_\mathbf{k}(t_f)^2
    \end{align}
    where $z_\mathbf{k}(\cdot)$ is the solution to the augmented-state ergodic visitation differential equation
    \begin{align}
        \dot{z}_\mathbf{k}(t) &= f_\mathbf{k}(s(t)) - \phi_\mathbf{k} \\ 
        z_\mathbf{k}(0) &= 0 \forall \mathbf{k} \in \mathcal{K}.
    \end{align}
\end{lemma}
The variable $z_\mathbf{k}$ can be seen as a continuous visitation variable that is being recorded alongside the trajectory evolution. 
Using the augmented-state ergodic formulation, the HJI ergodic reachability problem is defined as 
\begin{align} \label{eq:ergodic_control_disturbance}
    \begin{split}
    &\min_{u \in \mathcal{U}} \max_{d \in \mathcal{D}} \hspace{0.5em} h(s(\cdot), t_f) + \int_0^{t_f} u(t)^\top R u(t) dt \\ 
    &\text{subject to } \\
    & \dot{s}(t) = f(s(t), u(t), d(t)), s(0) = s_0 \\ 
    & \dot{z}_\mathbf{k}(t) = f_\mathbf{k}(s(t)) - \phi_\mathbf{k}, z_\mathbf{k}(0) = 0 \forall \mathbf{k} \in \mathcal{K}\\
    & d(t) = d(s(t), z(t), u(t)) \\
    & h(z(t_f), t_f) = \min(0,\frac{1}{t_f^2}\sum_{\mathbf{k}\in\mathcal{K}} \lambda_\mathbf{k} z_\mathbf{k}(t_f)^2 -\varepsilon) 
    \end{split}
\end{align}
which has the solution 
\begin{align}
    -\frac{\partial V(s, z,t)}{\partial t} = \min_{u\in\mathcal{U}} \max_{d \in\mathcal{D}} H(s,z,d,u, t)
\end{align}
where 
\begin{align}
    \begin{split}
        & H(s,z,d,u, t) = u(t)^\top R u(t) + \frac{\partial V(s,z,t)}{\partial x}^\top f(s, u, d) \\
        & + \sum_{\mathbf{k} \in \mathcal{K}} \frac{\partial V (s,z,t)}{\partial z_\mathbf{k}}^\top (f_\mathbf{k}(s(t)) - \phi_\mathbf{k})          
    \end{split}
\end{align}
and the value function $V(s,z,t)$ has terminal condition 
\begin{equation}
    V(s,z,t_f) = \min(0,\frac{1}{t_f^2}\sum_{\mathbf{k}\in\mathcal{K}} \lambda_\mathbf{k} z_\mathbf{k}(t_f)^2 -\varepsilon). 
\end{equation}
Here, the optimal control $u_\mathcal{E}$ and disturbance $d^\star$ are given by 
\begin{align}
    u_\mathcal{E} &= \arg \min_{u \in \mathcal{U}}\max_{d \in \mathcal{D}} H(s,z,d,u,t) \\ 
    d^\star &= \arg \max_{d \in \mathcal{D}} H(s,z,d,u_\mathcal{E},t).
\end{align}
    
\subsection{Performance and Discussion}

\begin{figure*}[h!]
    \centering
    \includegraphics[width=\linewidth]{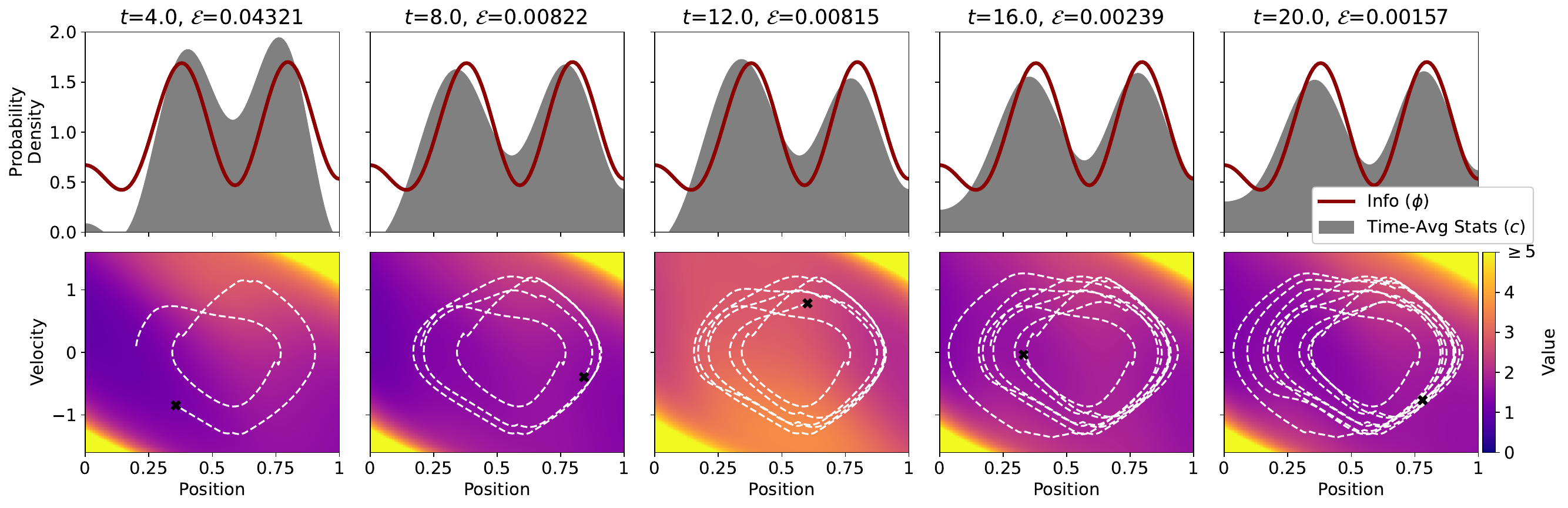}
    \caption{\textbf{Time-Evolution of a 2D drone following a robust ergodic controller.} Top row: the time-averaged spatial distribution and the information distribution. As time increases, the time-averaged trajectory statistics approach the information distribution, which is the goal of ergodic exploration. Bottom: evolution of the trajectory through phase space, superimposed on a cross-section of the value function with respect to position and velocity, where $z_k$ are fixed to their value at that instant in the trajectory. }
    \label{fig:hji_ergodic_result}
\end{figure*}

The resulting controller is considered a robust controller in the ergodic sense. 
That is, under the disturbance model, the ergodic controller will satisfy ergodicity up to the level $\varepsilon$. 
In the example Figure~\ref{fig:hji_ergodic_result} is a case where the ergodic controller is solved for a 2D drone coverage problem (see~\citet{berger_range_2024} for additional detail). 
The goal for the drone is cover an area with a target probability density given by $\phi$. 
The resulting value function from solving the HJI ergodic problem is shown below for different ergodicity $\mathcal{E}$ levels in Fig.~\ref{fig:hji_ergodic_result}. 

While the results show promise, there are limitations regarding computing robust ergodic controllers at scale. 
Specifically, as the complexity of the value function increases, so does the computation time to compute the value function. 
Likewise, the augmented ergodic states add additional complexities in terms of expanded state dimensionality $\mathcal{O}(|\mathcal{K}|^n)$. 
This poor computational scaling makes computing robust controllers prohibitively expensive. 

\subsection{Robustness via Multi-Modal Stein Variational Inference}

One way to mitigate the computational effects of computing robust ergodic controllers is to pose the ergodic control problem via exact probabilistic inference. 
Consider an optimality criteria on ergodic trajectories $\mathcal{O} : \text{Paths}(\mathcal{S}) \to \{ 0, 1 \}$. 
Then, given an ergodic metric $D(p_s, q)$, the ergodic cost likelihood function is defined by 
\begin{equation}
    p(\mathcal{O} | s) :=\exp(-\gamma D(p_s, q))
\end{equation}
where $\gamma>0$ is a hyperparameter. 
Given a prior on ergodic trajectories $p_0$ one can achieve robustness by computing a set of exact samples of the posterior distribution $p(s |\mathcal{O})$ over more general disturbance models. 

\begin{figure}
    \centering
    \includegraphics[width=\linewidth]{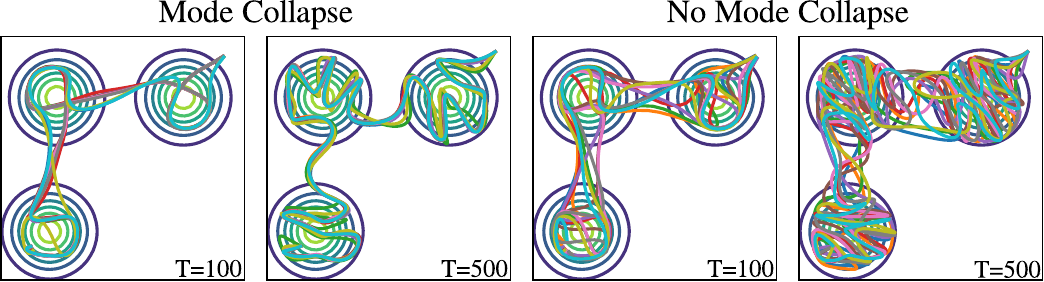}
    \caption{\textbf{SVGD Prevents Mode Collapse and Promotes Diverse Ergodic Controllers.} Here, we demonstrate mode collapse of inferring ergodic trajectories, i.e., when there is no repulsive force to diversify solutions. (Left) Ergodic trajectories randomly initialized converge to similar solutions. (Right) With the repulsive force from SVGD, the trajectory solutions leverage the multimodal nature of the ergodic metric, yielding diverse solutions, and ultimately solution robustness to uncertainty. Mode collapse can occur with increased dimensionality of the optimization variables which can be mitigated with the right choice of kernel~\citep{liu_stein_2016}. }
    \label{fig:mode_collapse}
\end{figure}

The idea it to compute $p(s | \mathcal{O})$ using Stein variational gradient descent~\eqref{eq:svgd} over a set of trajectories $\{s_i\}_{i=1}^N$. 
Since the ergodic metric encodes multiple locally optimal trajectory solutions, one can effectively achieve robustness by directly sampling the posterior distribution. 
The main result is that the Stein variational gradient prevents mode collapse\footnote{Under an appropriate norm~\citep{liu_stein_2016, lee_stein_2024}} as shown in Fig.~\ref{fig:mode_collapse}.
Through the SVGD approach, the practitioner can encode arbitrary uncertainty distributions which can expand the robustness of achieving ergodicity for an autonomous system. 

\begin{figure}
    \centering
    \includegraphics[width=\linewidth]{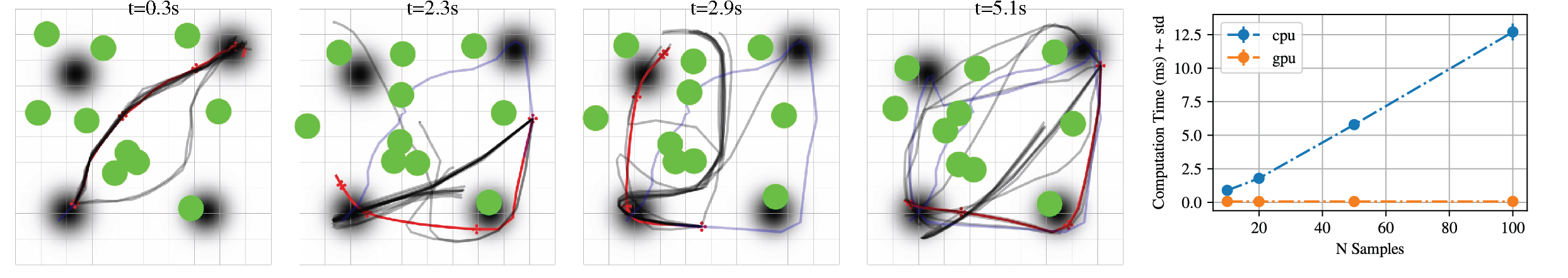}
    \caption{\textbf{GPU Acceleration of SVGD and the Ergodic metric for Robust Control.} (Left) Example SVGD for robust ergodic control in an online exploration problem (cover dark areas) with dynamics obstacles (green). A total of $12$ ergodic trajectories are calculated (shown in black). The red line indicates the best possible solution according to a heuristic (\citep{lee_stein_2024}), and blue line indicates visited areas. (Right) Computational scaling across devices. The Stein variational ergodic controller can optimize several paths that can be used as redundancies to attain robustness in uncertain and dynamic environments. }
    \label{fig:fast_svgd}
\end{figure}

Another advantage of the SVGD approach for achieving robust ergodic control is the computational speed. 
If one chooses an appropriate ergodic metric, e.g., the Fourier-based metric, calculating a set of ergodic controllers can be done quickly, leading more time to adapt to environmental changes. 
In Fig.~\ref{fig:fast_svgd}, we demonstrate the ability to achieve robustness in ergodic control. 
Note that both the ergodic metric and computing the SVGD across samples can be parallelized on a GPU leading to near constant time computational scaling (up until memory limitations). 
Subsequently, this result opens up potential future work that explores the various kinds of uncertainty that one can iterate along. 
In particular, domain uncertainty like external disturbance forces, could be considered in this formulation as a problem setting for developing robust ergodic controllers. 
Ultimately, the SVGD ergodic control approach to robustness has potential to solve several problems related to robustness and performance guarantees in ergodic control. 

\section{Time- and Energy- Optimal Ergodic Control}

In this section, we consider ergodic control under two unique settings: time and energy optimality. 
What makes these two conditions interesting is that 1) ergodicity requires infinite time to be optimal; and 2) infinite-time coverage suggests infinite energy requirements. 
As a strict mathematical optimization problem, solutions are ill-posed and non-trivial, requiring careful consideration when solving. 
In practice, ergodic coverage is typically bounded and episodic in nature: the robot eventually will return and report its findings. 
Hence, this section explores the implications of treating ergodicity as a constraint rather than a primary objective. 

\subsection{Ergodicity as a Constraint}

Here, we abandon strict minimization of the ergodic metric and instead strive to achieve \emph{sub-ergodic} control in order to compute solutions under time and energy restrictions. 
A similar approach was proposed in the previous section when dealing with robustness under disturbances. 
The intuition here is that \emph{complete} ergodicity is not realistically achievable \textemdash infinite time is required under Theorem.~\ref{thm:ergodicity}. 
However, sub-ergodic behavior, i.e., when $t\ll \infty$, is plausible to compute under the trajectory optimization formulation and has the advantage of containing many local minima, as shown in the previous section, yielding flexibility in the solution space. 
Hence, we are interesting in the setting when ergodicity is bounded from above $D(p_s, q) \le \varepsilon$, with $\varepsilon >0$ and time and energy are left as free variables.

The general formulation requires posing ergodicity as a constraint involves minimizing a limited resource $\beta$ that \emph{inversely} impacts coverage performance according to the ergodic metric.
We can write this problem as 
\begin{align} \label{eq:resource_constrained_ergodic}
    \begin{split}
    &\min_{s(\cdot) \in \mathcal{S}, u(\cdot)\in\mathcal{U}, \beta \ge 0 } \hspace{1em} \beta \\ 
    &\text{subject to} \\
    & h_1(s, u) = 0, h_2(s, u) \le 0 \\
     &D(p_s, q, \beta) \le \gamma 
     \end{split}
\end{align}
where $h_1, h_2$ are equality and inequality constraints, $s(\cdot), u(\cdot)$ may depend on the variable $\beta$, and $\gamma \in \mathbb{R}_+$ is an upper bound. 
Time and energy are both examples where $\frac{1}{f(\beta)} \propto D(p_s, q, \beta)$ over some function $f$ (e.g., polynomial). 
Thus, minimizing $D(p_s, q, \beta)$ over $\beta$ naively leads to infinite solutions. 
The most obvious example is in Theorem~\ref{thm:ergodicity} where the time-averaged statistics $p_s$ converges on the spatial statistics $q$ when $t\to \infty$, thus being a solution to minimizing an ergodic metric over time. 
By minimizing the resource (in this case time), $\lim_{t\to 0} D(p_s, q) = \infty$, which necessitates an upper bound to find a fixed point solution. 
Below we provide numerical solutions to these resource constrained problems for both time and energy considerations. 

\subsection{Time-Optimal Ergodic Control}

Letting $\beta=t_f$ in \eqref{eq:resource_constrained_ergodic} poses the time-optimal ergodic trajectory optimization problem 
\begin{align} \label{eq:time_optimal_ergodic}
    \begin{split}
    &\min_{s(\cdot) \in \mathcal{S}, u(\cdot)\in\mathcal{U}, t_f \ge 0 } \hspace{1em} t_f \\ 
    &\text{subject to} \\
    & h_1(s, u) = 0, h_2(s, u) \le 0 \\
     &D(p_s, q, t_f) \le \gamma 
     \end{split}
\end{align}
where $h_1, h_2$ encodes dynamic constraints. 
One can solve the problem time-optimal ergodic trajectory optimization problem via satisfying the Pontryagin's Maximum Principle (as done in~\citet{dong_time_2023}); however, the required time to satisfy an ergodic constraint may be quite large (as $t_f$ grows quadratically as $D(p_s, q, t_f)\to0$), leading to numerical instabilities. 
Rather, Problem~\eqref{eq:time_optimal_ergodic} is best solved via direct transcription over $N$ control knot points with discrete time-step $\Delta t = \frac{t_f}{N}$. 
The trajectory is then solved as a nonlinear program where $\mathbf{s} = [s_0, \ldots, s_{N}]$ and $\mathbf{u} = [u_0, \ldots, u_N]$.

\begin{figure}[ht!]
    \centering
    \includegraphics[width=\textwidth]{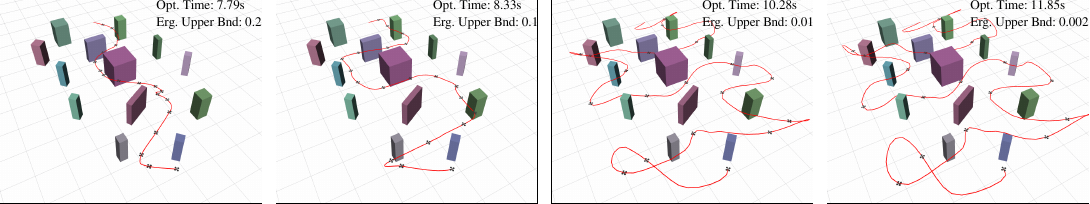}
    \caption{\textbf{Time Optimal Ergodic Trajectory Evolution~\citep{dong_time_2023} in clutter.} As one lower the upper bound $\gamma$, satisfying the ergodic metric requires the solver to search for larger $t_f$ which allows the robot trajectory to cover more of the space. Here, $\phi$ is given as a uniform distribution. }
    \label{fig:sim_cluttered_search}
\end{figure}

An example solution of solving the time-optimal ergodic control problem is presented in Fig.~\ref{fig:sim_cluttered_search}. 
The minimum optimal time value is shown to be directly impacted by the upper bound on the ergodic metric. 
Likewise, the overall trajectory length will increase, allowing for more coverage over the domain. 

\subsection{Energy-Optimal Ergodic Control}

Energy optimal considerations are less obvious than the time-optimal problem. 
This is due to the fact that energy may not directly impact the performance on ergodicity as it depends on the dynamic constraints of the robotic system.
In energy optimal ergodic control problems, the goal is to maintain persistent coverage so long as energy is minimized (or within some bound). 
The challenge is that once a robot returns to recharge, coverage is momentarily lost. 
Thus, solving the energy optimal ergodic control problem then requires the use of multiple-robots to ensure persistent coverage.

Let $b_i(t)$ be the battery state of charge for the $i^\text{th}$ robot with state $s_i(t)$. 
Furthermore, let $D(p_s, q)$ denote the joint ergodicity for the multi-agent system with $N$ agents as $s(t) = [s^{(1)}(t), \dots, s^{(N)}(t)]^\top$ subject to controls $u(t) = [u^{(1)}(t), \dots, u^{(N)}(t)]^\top$.
Then the energy optimal control problem is given as 
\begin{align}
    \begin{split}                
    &\min_{s(\cdot) \in \mathcal{S}, u(\cdot) \in \mathcal{D}} \sum_{i} \int_{0}^{t_f}(u^{(i)} (t))^\top R u^{(i)}(t) \\ 
    & \text{subject to } \\ 
    & \dot{s}^{(i)} = f^{(i)}(s^{(i)}(t), u^{(i)}(t)), s^{(i)}(0) = s^{(i)}_0 \\ 
    & b_i(t) \in (0, b_f] \forall i \in [1,\ldots, N]\\
    & h_1(s, u) = 0, h_2(s, u) \le 0 \\ 
    & D(p_s, q) \le \gamma
    \end{split}
\end{align}
where $f^{(i)}$ and $s^{(i)}(t)$ integrate the battery state $b^{(i)}(t)$ with a position-dependent recharge station, $b_f$ is a upper bound on acceptable battery levels, $h_1, h_2$ are more generic constraints, and $\gamma$ sets the coverage upper bound (see~\citet{seewald_energy-aware_2024} for more detail). 
Depending on the underlying dynamics, the battery charge will automate which agents are covering the domain and which are charging for continual operations. 

\begin{figure}[t!]
  \begin{minipage}[t!]{.5\columnwidth}
    \vspace*{-.2cm}
    \includegraphics[width=0.95\textwidth]{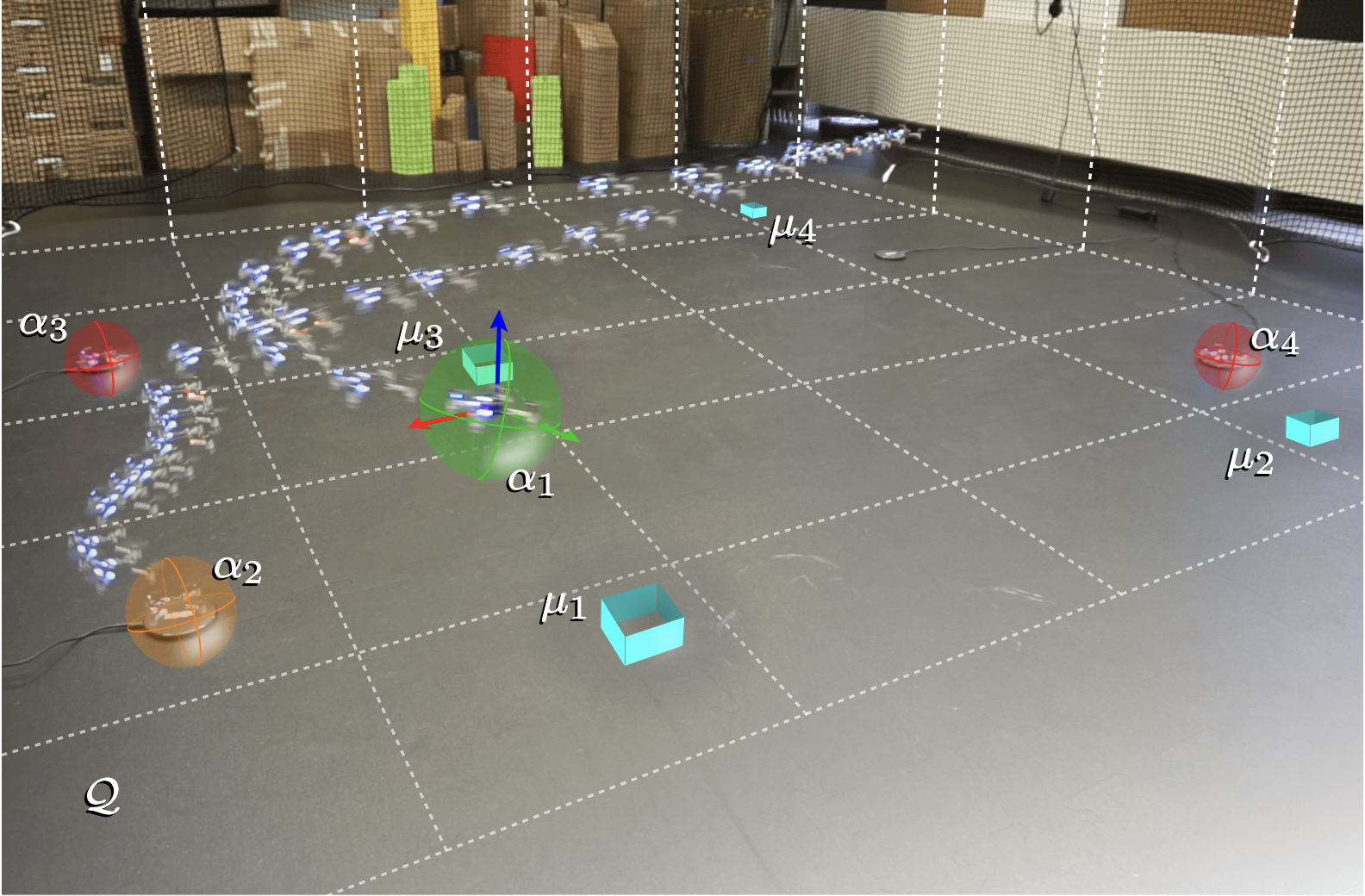}
  \end{minipage}
  \begin{minipage}[c]{.48\columnwidth}
    \vspace*{.05cm}
    \caption{\textbf{Energy-Aware Ergodic Coverage}. State of charge awareness can maintain a persistent level of ergodicity. Each agent shown trades off uniform coverage over the space and battery charge. High density areas are defined via $\mu_1, \mu_2, \mu_3, \mu_4$ whereas charging stations are denoted by $\alpha_1, \alpha_2, \alpha_3, \alpha_4$. }
    \label{fig:example_ergo}
  \end{minipage}
  \vspace*{-.4cm}
\end{figure}

\begin{figure}[t!]
  \begin{minipage}[t!]{.5\columnwidth}
    \vspace*{-.2cm}
    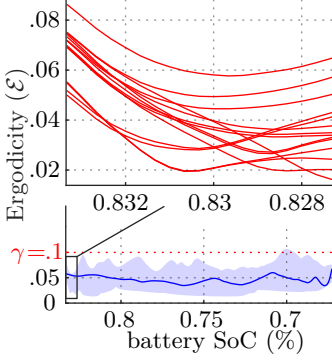
  \end{minipage}
  \begin{minipage}[c]{.48\columnwidth}
    \vspace*{.05cm}
    \caption{\textbf{Ergodicity 
    as a function of state-of-charge}. The top plot shows the evolution of the ergodicity for a set of agents. The bottom shows the average ergodicity. Initially, the agents are at their charging stations. As agents begin to lose battery charge, they they start moving to the charging stations -- ergodicity increases while other agents maintain the average ergodicity to be under the value $\gamma$.}
    \label{fig:ergo}
  \end{minipage}
  \vspace*{-.4cm}
\end{figure}

An example of persistent coverage enabled by the above optimization is presented in Fig.~\ref{fig:example_ergo}. 
Note that at least one agent will be around to continuously cover the domain while the remaining agents charge the battery to eventually trade off coverage. 
In Fig.~\ref{fig:ergo}, we can see that ergodicity is maintained on average to be within the $\gamma$ threshold as a function of battery state of charge (SoC).

\section{Ergodic Control in Curved Spaces}
Methods based on the Fourier ergodic metric are inherently limited to Euclidean domains with rectangular boundaries, as the Fourier series provides an orthonormal basis only under these geometric constraints. This severely limits their applicability to real-world robotics problems, where the state or workspace often lies in curved spaces or domains with complex internal or external boundaries. Examples include robot workspaces with obstacles, navigation tasks over uneven terrain, or interaction tasks with curved object surfaces. Beyond workspaces, many types of robotics data naturally lie on curved spaces. Direction vectors are defined on the sphere, robot joint configurations often define a torus-like manifold, and end-effector poses correspond to elements of the Lie groups. These scenarios require generalizations of the ergodic control framework that can account for curved geometries and non-Euclidean metrics. The following sections present two families of approaches for extending ergodic control to curved domains: spectral methods on low-dimensional Riemannian manifolds and alternative approaches extending to higher-dimensional Lie groups.

\subsection{Spectral Methods on Riemannian manifolds}
The Laplace–Beltrami operator is the natural extension of the Euclidean Laplacian to arbitrary Riemannian manifolds. Its eigenfunctions generalize the Fourier basis and form a geometry-aware spectral representation that can be used to define ergodic metrics consistent with the intrinsic structure of the domain. In simple geometries, such as hyperspheres, the eigenfunctions and eigenvalues of the Laplace–Beltrami operator admit closed-form solutions that reduce to spherical harmonics. \citet{jacobs_multiscale_2010} exploited this property to extend the Fourier-based ergodic metric for multiscale surveillance tasks on spheres and tori.

Most robotic applications, however, involve domains with irregular or non-uniform curvature, such as the surfaces of everyday objects or workspaces with internal and external boundaries. In these settings, no closed-form solutions exist, and the Laplace–Beltrami operator and its spectrum must be approximated numerically. A common approach is to discretize the domain and use finite element and compute a discrete Laplace-Beltrami operator as a sparse, symmetric, and positive semi-definite matrix.

\citet{xu_measure_2026} extended the Fourier ergodic metric to planar domains with complex internal and external boundaries by computing Laplace–Beltrami eigenfunctions on meshes representing the domain. Similarly, \citet{dong_ergodic_2025} used mesh-based Laplace-Beltrami eigenfunctions for ergodic trajectory optimization, though they proposed an alternative spectral weighting scheme that applied exponential decay to the square root of the eigenvalues, rather than the polynomial or exponential weighting typically used in Fourier or HEDAC formulations. 

\citet{ivic_constrained_2022} addressed similar problems by extending HEDAC from regular grids to triangular meshes using an FEM-based solver. \citet{ivic_multi-uav_2023} further scaled this approach to three-dimensional monitoring tasks with UAVs . While these methods successfully capture complex geometries, their reliance on FEM solvers and mesh preprocessing limits their applicability to offline planning scenarios.

\begin{figure}[t!]
    \centering
    \includegraphics[width=0.95\linewidth]{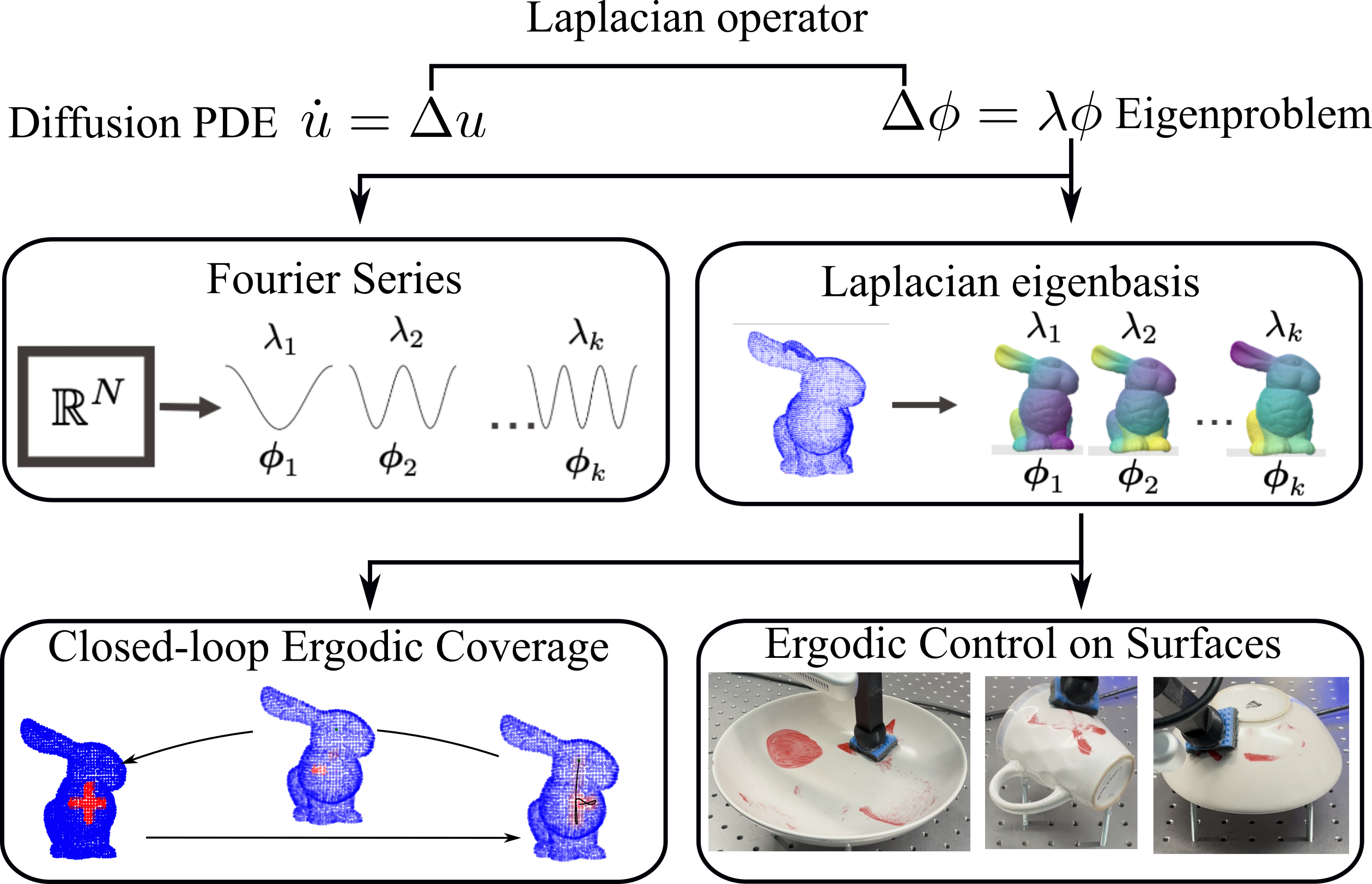}
    \caption{\textbf{The Laplacian eigenbasis extends Fourier analysis to curved manifolds, including surfaces reconstructed from point clouds. This basis enables efficient closed-loop ergodic control on such surfaces, supporting operations such as cleaning or surface inspection.
    }} 
    \label{fig:ergodic_control_on_surfaces}
  \vspace{-1em}
\end{figure}

To enable closed-loop ergodic control without requiring a prior mesh model of the object, \citet{bilaloglu_tactile_2025} extended HEDAC to operate directly on point clouds acquired online. This approach uses Laplace-Beltrami eigenfunctions of the point cloud to efficiently solve the diffusion equation in the spectral domain, allowing real-time operation. Because the point cloud can be updated online, both the object geometry and the target spatial distribution can adapt dynamically during the task, enabling ergodic coverage for interactive tasks such as cleaning as summarized in Figure \ref{fig:ergodic_control_on_surfaces}.

Overall, Laplace–Beltrami-based methods provide a principled and
geometrically consistent way to generalize Fourier-based ergodic metrics
to arbitrary curved domains. Although computational complexity scales
with the resolution of the discretization, modern numerical solvers make
real-time operation feasible for moderately complex surfaces.

\subsection{Ergodic Control Methods on Lie Groups}

Many robotic tasks require reasoning not only about positions but also about orientations and full poses. These domains are naturally represented by Lie groups, such as SO(3) for pure rotations or SE(3) for rigid-body transformations. Extending ergodic control to these spaces is challenging for two reasons. First, Lie groups are curved spaces, so the standard Fourier ergodic metric cannot be directly applied. Second, these spaces are often high-dimensional, which exacerbates the curse of dimensionality when using basis expansions.

\citet{shetty_ergodic_2022} extended the Fourier ergodic metric to the space of end-effector poses by treating the orientation component as living in the tangent space of the 3-sphere (S³) and using tensor-train decomposition to manage the computational cost of high-dimensional Fourier coefficient calculations. This approach has been demonstrated in a peg-in-hole insertion task.

Alternative approaches have sought to avoid explicit basis expansions. \citet{sun_fast_2025} proposed a kernel-based ergodic metric that replaces the Fourier basis with kernel functions, providing a computationally efficient way to approximate the ergodic metric that naturally handles curved spaces such as Lie groups, where it achieved orders-of-magnitude faster computation than spectral methods. 

\citet{hughes_ergodic_2024} introduced an ergodic metric based on the Maximum Mean Discrepancy (MMD), which operates purely on samples from the domain without requiring explicit geometric modeling or basis functions. This formulation can be applied to arbitrary domains, including surfaces or Lie groups, as long as sufficient samples are provided. However, because it does not explicitly encode the underlying geometry, it can suffer from lower computational efficiency compared to kernel-based methods.

In summary, spectral methods based on the Laplace–Beltrami operator offer a geometrically principled and multiscale framework for ergodic control on less-structured low-dimensional manifolds such as surfaces, while kernel-based and MMD-based metrics provide scalable and flexible alternatives, particularly for high-dimensional curved spaces such as SE(3). The choice among these approaches should be guided by the task requirements, balancing geometric fidelity and computational cost to achieve effective ergodic exploration and coverage in complex robotic domains.

\section{Ergodic Control for Multi-Agent Systems}

\subsection{Decentralized Ergodic Control}\label{sec:decentral}

Define a multi-agent system with $N$ agents as $s(t) = [s^{(1)}(t), \dots, s^{(N)}(t)]^\top$ over the time horizon $t\in[0, T]$, where the superscript denotes the agent index. We first extend the formulation of trajectory empirical distribution from a single-agent system toward the multi-agent system:
\begin{align}
    p_s(x) & = \frac{1}{T} \int_0^T \frac{1}{N} \sum_{i=1}^{N} \delta(x - s^{(i)}(t)) dt , \\
        & = \frac{1}{N} \sum_{i=1}^{N} \left(\frac{1}{T} \int_0^T \delta(x - s^{(i)}(t)) dt \right) = \frac{1}{N} \sum_{i=1}^{N} p_{s^{(i)}}(x), \label{eq:multi_agent_empirical}
\end{align} which is equivalent to the average of individual agents' trajectory empirical distributions. The multi-agent ergodic control problem is to generate control sequence $u(t)=[u^{(1)}(t), \dots, u^{(N)}(t)]^\top$ for the multi-agent system such that the empirical distribution (\ref{eq:multi_agent_empirical}) asymptotically converges to the target distribution $q(x)$.

Based on (\ref{eq:multi_agent_empirical}), we can extend the Fourier ergodic metric for the multi-agent system:
\begin{gather}
    D(P_s, Q) = \sum_{\mathbf{k}\in\mathcal{K}} \lambda_{\mathbf{k}} \vert c_{\mathbf{k}} - \phi_{\mathbf{k}} \vert^2, \\
    c_{\mathbf{k}} = \frac{1}{N} \sum_{i=1}^{N} \left( \frac{1}{T} \int_0^T f_{\mathbf{k}}(s^{(i)}(t)) dt \right) = \frac{1}{N} \sum_{i=1}^{N} c_{\mathbf{k}}^{(i)}, \label{eq:multi_agent_coefficients}
\end{gather} where $c_{\mathbf{k}}^{(i)}$ is the corresponding Fourier coefficient for agent $i$'s trajectory and the Fourier coefficient for the multi-agent system is the average of the Fourier coefficients of all individual agents' trajectories. For the $i$-th agent, the partial derivative of the overall ergodic metric with respect to the individual agent's trajectory is:
\begin{align}
    \frac{\partial}{\partial s^{(i)}} D(P_s, Q) = 2 \sum_{\mathbf{k}\in\mathcal{K}} \lambda_{\mathbf{k}} ( c_{\mathbf{k}} - \phi_{\mathbf{k}} ) \cdot \nabla f_{\mathbf{k}}(s^{(i)}(t)). \label{eq:decentralied_derivative}
\end{align} Based on (\ref{eq:decentralied_derivative}), given the Fourier coefficients of the overall multi-agent system $\{c_{\mathbf{k}}\}_{\mathbf{k}\in\mathcal{K}}$, each individual agent can independently optimize the overall ergodic metric without accessing to other agents' state information. Therefore, one can apply the control optimization methods for single-agent ergodic control to derive decentralized ergodic control for multi-agent systems. 

Furthermore, regarding the communication between agents, only the Fourier coefficients of each individual agent's trajectory needs to be shared with other agents. As shown in (\ref{eq:multi_agent_coefficients}), the Fourier coefficients of the multi-agent system is equivalent to the average of the Fourier coefficients of all the individual agents. Based on this intuition, \citet{abraham_decentralized_2018} shows that global optimal decentralized ergodic control can be achieved through consensus via all-to-all communication of the individual Fourier coefficients. Building on this result, \citet{taylor_safe_2024} show that global optimality for decentralized ergodic control can still be asymptotically achieved under sporadic communication, where pairs of agents exchange their individual Fourier coefficients only when they are within a limited communication range. Multi-robot ergodic control has also been extended to explicitly account for connectivity, including probabilistic connectivity and relaxed periodic connectivity constraints~\citep{liu_multi-robot_2025, liu_probabilistic_2025}.

Such decentralization paradigms also enable alternative communication approaches. \citet{taylor_stigmeric_2024} show that relying only on trajectory history for communication can enable stigmergic communication, and \citet{taylor_manufacturing_2026} apply this framework to multi-agent surface micro-patterning, providing a promising solution to the challenge of manufacturing large-scale micro-patterned surfaces.

Lastly, the empirical-distribution formulation of decentralized ergodic control provides a natural connection to mean-field methods for large-scale multi-agent systems~\citep{huang_large_2006, lasry_mean_2007}. As the number of agents increases, the finite-agent empirical distribution can be approximated by a continuous population distribution, such that each agent responds to aggregate statistics of the population rather than the states of all other agents. In decentralized ergodic control, these aggregate statistics are represented compactly through the shared Fourier coefficients, whose dimension depends on the spectral representation rather than the number of agents.

\subsection{Ergodic Control with Heterogeneous Agents}

Many applications for multi-agent systems can benefit from the use of multiple heterogeneous agents.
In particular, agents with diverse sensing and motion capabilities may allow access to more types/modes of information or regions of the search space.
However, the efficacy of a heterogeneous team is reliant on effective coordination between agents.
The unique capabilities of heterogeneous agents can be leveraged to improve team coordination, particularly towards a shared goal like search.

An approach to coordinating heterogeneous agents is to allocate agents to different parts of the search problem, based on their capabilities. 
\citet{sartoretti_spectral-based_2022} exploits the spectral nature of ergodic control to decompose the search problem and distribute agents.
Specifically, the spectral-based allocation framework introduced plans ergodic trajectories for each type of agent based on a smaller subset of the spectral coefficients associated with the information prior. 
This approach drives agents to search the domain at a spatial scale that best matches their motion and sending capabilities.
In this first work, the selection of those spectral bands was made manually, based on domain knowledge and human expertise, and shown to increase overall search performance compared to random/adversarial assignments.

\begin{figure}[h!]
    \centering
    \includegraphics[width=\linewidth]{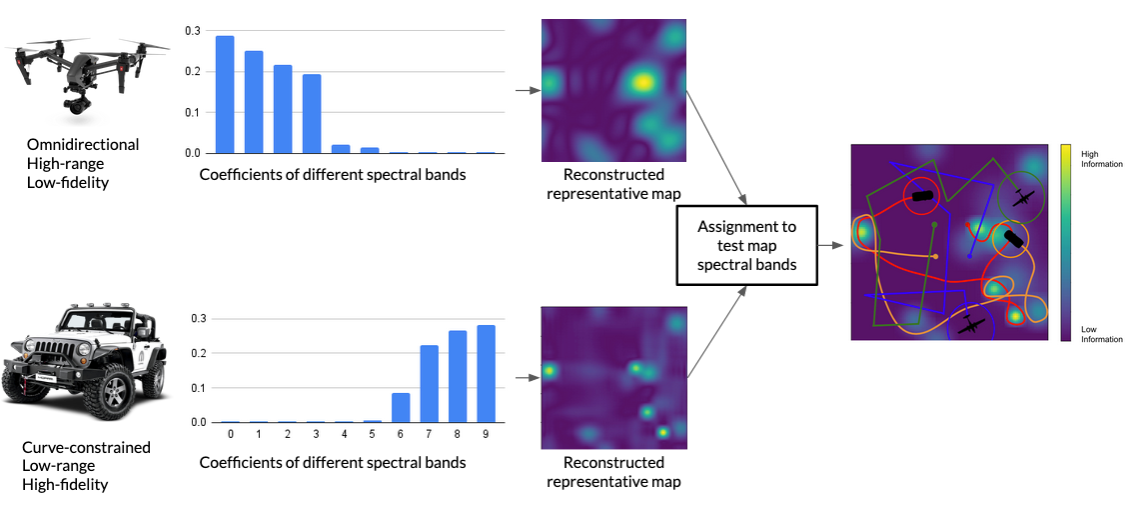}
    \caption{\textbf{Example spectral decomposition of an information prior with two heterogeneous agents:} A given information prior is decomposed directly in the spectral domain, and the spectral coefficients assigned to the different agents based on human intuition. Here, the UAV with omnidirectional motion capabilities and high-range, low-fidelity sensing (e.g., onboard camera) is assigned low-order coefficients, allowing it to quickly cover the lowest-frequency portions of the information space. Conversely, a ground vehicle with curve-constrained motion constraints, equipped with a low-range, high-fidelity sensor (e.g., LiDAR) is assigned higher-frequency coefficients, allowing it to easily chain peaks of information.}
    \label{fig:info_map_decomposition_example}
\end{figure}

The spectral decomposition of an information prior can be interpreted to correlate different frequency components to different spatial scales of information. 
As illustrated in Fig~\ref{fig:info_map_decomposition_example}, lower frequency coefficients correspond to larger-scale variations in the spatial distribution of information, while the higher frequency coefficients correspond to smaller-scale variations.

Given $M$ spectral bands (i.e., sets of frequency coefficients within particular ranges), with $M$ the number of agent types in the heterogeneous team, each band can be seen as a separate (although not completely independent) \textbf{search subtask} that can be distributed to a specific type of agent based on its motion and sensing capabilities to search at a given spatial scale.
To formally define and use these search subtasks, authors considered a modified ergodic metric that relies only on specific band(s) of coefficients:

\begin{equation}
\Phi(\gamma(t))=\sum_{k=c_1}^{c_2} \lambda_k \left| c_k (\gamma(t)) - \xi_k \right|^2,
\label{eq:spectral_band_definition}
\end{equation}

\noindent where $c_1$ and $c_2$ define the first and last coefficients of the spectral band. Note that the same result could be achieved by setting $\lambda_k = 0$ $\forall k < c_1, k > c_2$.

Once search subtasks are defined, they need to be allocated to different agents.
This assignment can be based on some intuition of the agents' abilities.
For example, agents with large, low-fidelity sensors and faster, omnidirectional motion models are better at capturing broad-stroke information and so are allocated to lower-order frequency components.
On the other hand, agents with smaller, more precise sensors and slower, curve-constrained motion models are better suited to disambiguating detailed information and so are allocated to higher-order frequencies.
This intuitive allocation improves the search and coverage performance of heterogeneous teams (see \citet{sartoretti_spectral-based_2022} for more details).

However, relying on human understanding of agent skills becomes more difficult as the agents or the search space increase in complexity.
To address this issue,~\citet{rao_learning_2024} then proposed and demonstrated that the allocation of agents to different search subtasks can also be learned from data.

\begin{figure}[ht!]
    \centering
    \includegraphics[width=\linewidth]{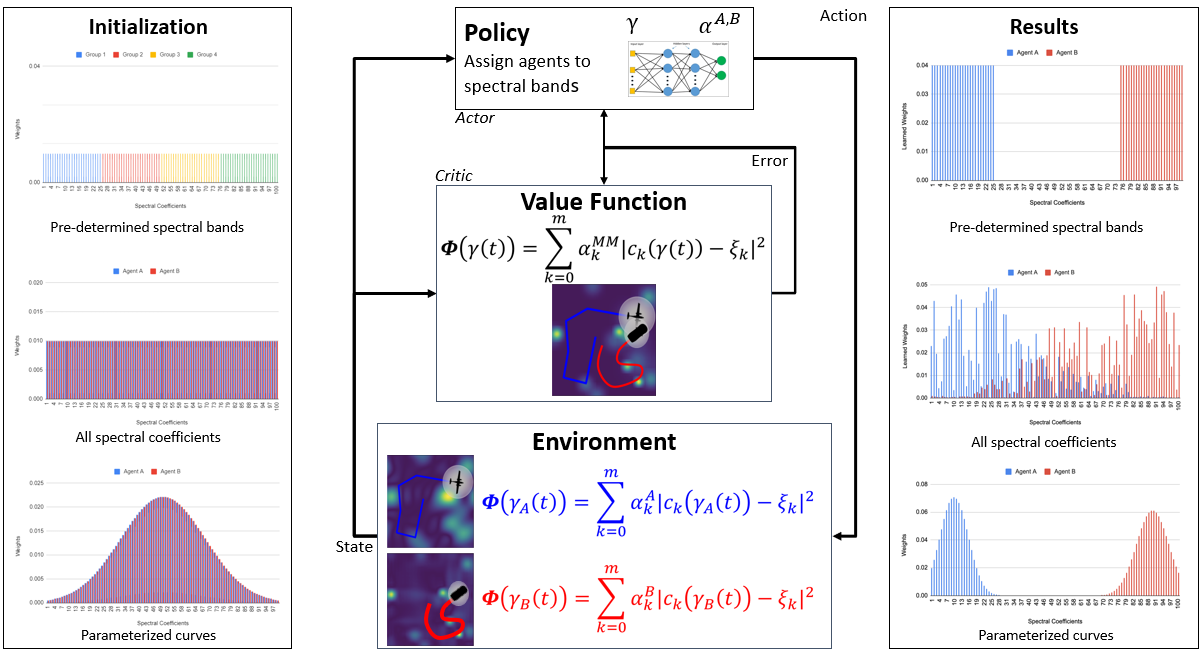}
    \caption{\textbf{Learning spectral allocations~\citep{rao_learning_2024}:} Relying on the A3C algorithm with three different distribution schemes, agents learn their allocation to search subtasks. Each distribution method has different initial conditions. In all three methods, the output is a weighted set of spectral coefficients of the information distribution being searched for each agent. The learning process seeks to minimize the joint ergodic metric of the team's trajectories.}
    \label{fig:learning_spectral_allocations}
\end{figure}

\citet{rao_learning_2024} proposed to rely on reinforcement learning to obtain data-driven mappings of agents to search subtasks (as shown in Fig~\ref{fig:learning_spectral_allocations}).
In this pipeline, each agent in a heterogeneous team is assigned a spectral scale of the search domain, after which paths are planned for all agents.
A joint reward is then calculated for all agents, based on their coverage of the search map measured by the ergodic metric.
This reward function is used to train the agents' policies, which allows them to select the band(s) they believe they are best suited for to collaboratively search the space.
In doing so, agents learn to answer the question ``which part(s) of the spectral decomposition of the underlying information distribution am I best suited to cover ergodically?''

Three specific methods are evaluated for distributing agents in a heterogeneous team to spectral scales of a search domain.
First, agents are mapped to predetermined, hand-crafted, spectral bands. 
In the second method, individual weights are learned over all spectral coefficients of the information distribution.
Finally, weights are learned for parameterized curves over the spectral coefficients of the information distribution.

In this work, the allocations learned in the spectral bands case were shown to confirm the intuition-based allocations used in~\citet{sartoretti_spectral-based_2022}.
In addition, more granular allocation methods outperform intuition-based allocation in terms of information gathered, leading to enhanced performance.

\section{When Not to Use Ergodic Control} \label{chap:when_not_to_use_ergodic_control}

Ergodic control should not be used when the temporal structure of a task is sufficiently critical, such that representing the trajectory primarily through its time-averaged spatial statistics would compromise task success. By design, the ergodic metric is largely insensitive to the order in which states are visited, thus it does not directly enforce that particular states be reached before others or within prescribed time windows. This property becomes problematic in specific tasks with strict precedence, transition, or completion requirements. Examples include an assembly operation in which one component must be inserted before another becomes accessible; a manipulation task in which a prescribed sequence of contacts is required to avoid jamming or object damage; a coordinated handoff in which two robots must reach compatible configurations at the same time; or an inspection task in which every designated feature must be observed before a fixed deadline. While certain formulations, such as time-optimality~\citep{dong_time_2023}, can address some of the issues by ensuring coverage is completed within a specific time budget, ergodic control in general does not dictate the sequence or precedence of robot actions. Under such circumstances, methods that explicitly represent temporal order, discrete transitions, and finite-horizon completion constraints are therefore more appropriate. Nevertheless, ergodic control can still remain useful locally within such tasks, where exploratory behavior is needed within a short time window while the sequential structure of the overall task is preserved by a higher-level planner. The aforementioned limitations of ergodic control are therefore not that ergodic control is inherently unsuitable for these scenarios, but that its distribution-based specification should not replace temporal structure that is essential to task success.


\chapter{Ergodic Control: Motion Synthesis Applications}

This chapter will review ergodic control for various robot motion synthesis applications across different domains. 

\section{Information Gathering}

Search and rescue (SAR) operations, inspection, and environmental monitoring all involve gathering information in dynamic and uncertain environments, making ergodic control a potential candidate for efficient motion planning. Recent experimental validations have demonstrated the effectiveness of ergodic control for autonomous UAVs conducting human search and detection in realistic wilderness environments~\citep{dumencic_experimental_2025}. In this work, HEDAC is used to direct an optimized search by generating trajectories based on the spatial probability density of finding a target. This coverage strategy is combined with YOLO, a semantic labeler acting as the detection model, and an optimal Model Predictive Control (MPC) controller to efficiently navigate and search a realistic Mediterranean karst forest environment~\citep{dumencic_experimental_2025}. By integrating these components, the UAV or multiple UAVs~\citep{lanca_probabilistic_2026} can thoroughly scan high-probability regions while maintaining sufficient coverage of the broader area, ensuring that the search is both rapid and exhaustive despite the physical constraints of the platform.

\begin{figure}[ht!]
    \centering
    \includegraphics[width=1.0\linewidth]{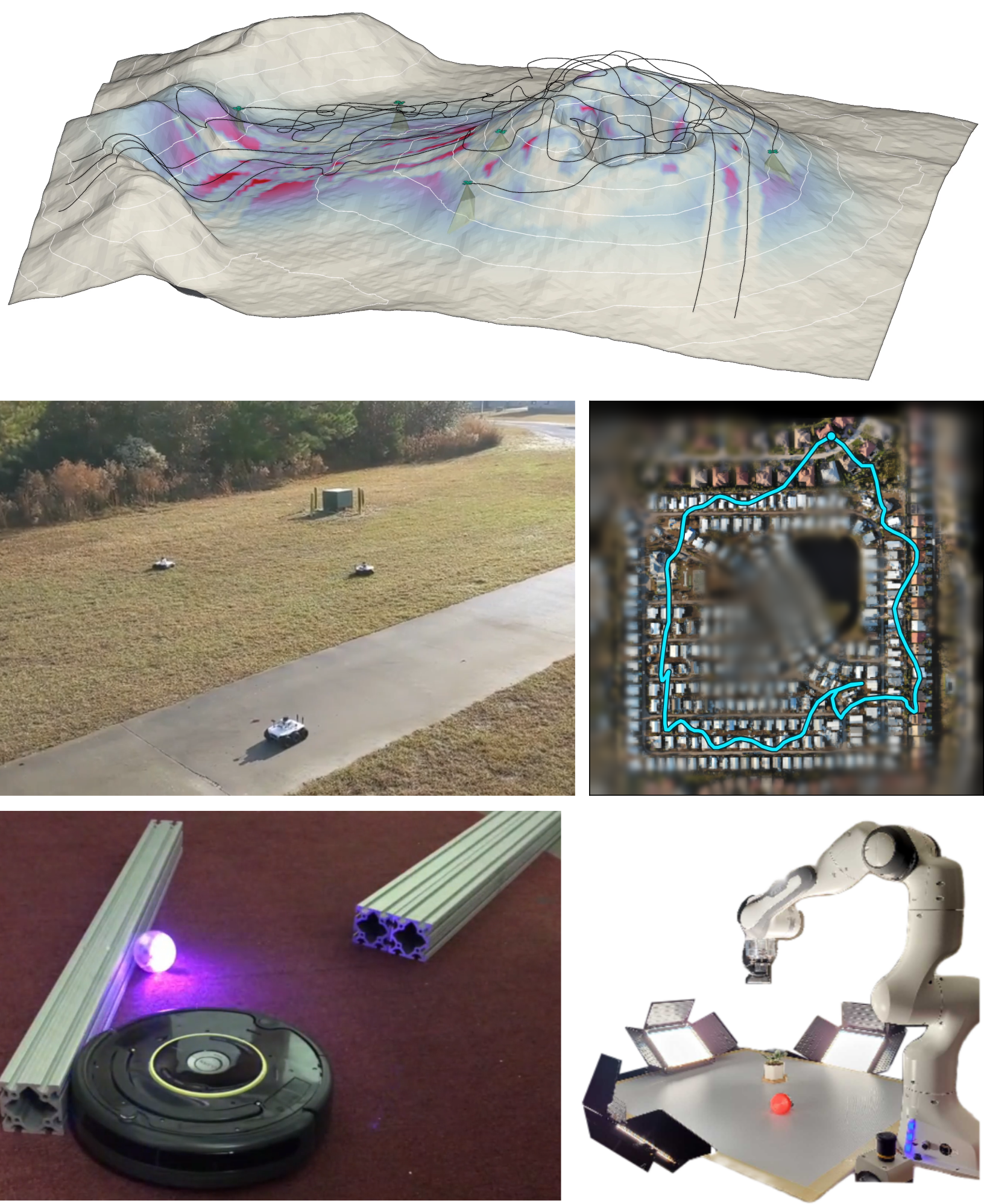}
    \vspace{-1em}
    \caption{Examples of ergodic control applications in information gathering and active perception. From top to bottom: (1) Information gathering with UAVs over uneven terrains; (2) Information gathering with human-swarm systems; (3) Information gathering over disaster sites; (4) Active perception with binary sensors; (5) Active perception with generative models.}
    \label{fig:application_info}
\end{figure}

Beyond searching according to a fixed target distribution, ergodic control can also continuously update the search trajectory as new measurements are collected. Receding-horizon ergodic exploration has been used to simultaneously provide area coverage and localize an unknown number of targets in real time, with the formulation also extending to distributed multi-robot operation~\citep{mavrommati_real-time_2018}. This ability is particularly important when robots are equipped with noisy or limited sensors. Multi-agent dynamic ergodic search has been demonstrated for locating and tracking moving targets using only low-information binary measurements, where coordinated robot motion allows information to be accumulated over time despite the limited sensing capability of each individual platform~\citep{coffin_multi-agent_2022}. Ergodic search has also been extended to multiple competing information maps by computing Pareto-optimal tradeoffs among the corresponding ergodic objectives~\citep{ren_local_2022, ren_pareto-optimal_2023}, with a multi-agent extension that jointly considers objective allocation and ergodic search~\citep{srinivasan_multi-agent_2023}.

In search and rescue, both the nature of the task and the battery life limitations of the mobile platforms make time a critical factor in efficient operations. To address this, one can formulate the ergodic metric as an inequality constraint with an upper bound such that the time of search needed to satisfy that bound can be minimized---enabling one to compute a time-optimal ergodic trajectory~\citep{dong_time-optimal_2024}. Importantly, this time-optimal formulation allows a UAV to execute complex, high-speed maneuvers while avoiding obstacles even in large, cluttered environments. Safety constraints can also be incorporated directly into ergodic trajectory optimization through control barrier functions, enabling single or multiple UAVs to perform information-directed search while maintaining collision-free trajectories in cluttered environments~\citep{lerch_safety-critical_2023}.

The probability distribution associated with a target may itself evolve during the search. This is particularly common in maritime SAR, where wind and ocean currents cause the possible target location to drift while the search platforms are being deployed. Ergodic exploration has been extended to account for such dynamic distributions by evolving the target probability field according to the environmental flow and the sensing effort already performed by the robots~\citep{lanca_ergodic_2025}. This allows multiple UAV missions to search according to the continuously changing probability of finding the target, rather than repeatedly searching according to an outdated static distribution.

The inspection of confined spaces---such as tunnel networks, ballast tanks, or industrial piping---presents a different topological challenge. \citet{wong_time-discounted_2026} introduced a time-discounted ergodicity metric on graphs in order to enable active robotic inspection in confined spaces. This formulation naturally prioritizes near-term information gathering, ensuring that the robot rapidly covers the most critical areas of the graph first, which is particularly advantageous when operating under strict time constraints or in highly restricted spaces. For building interiors, corridors, and other convoluted environments, the geometry of the free space and the locations of obstacles can instead be incorporated into the ergodic metric using Laplace--Beltrami eigenfunctions, while measure-preserving flows generate collision-free trajectories for single or multiple robots~\citep{xu_measure_2026}. These geometry-aware formulations allow ergodic search to be applied in environments where standard Fourier representations over simple rectangular domains might otherwise generate inefficient or infeasible trajectories.

Infrastructure inspection may also require the robot to gather information directly over a curved surface rather than within a planar search domain. Ergodic exploration over meshable surfaces extends the construction of the ergodic metric to arbitrary surfaces represented by triangle meshes, allowing trajectories to be generated over structures such as wind turbines and other geometrically complex objects~\citep{dong_ergodic_2025}. Together with graph-based and geometry-aware formulations, this extends ergodic inspection from open Euclidean environments to confined networks, non-convex interiors, and complex three-dimensional surfaces.

Similarly, in environmental monitoring and infrastructure inspection, sensors must often track spatiotemporal phenomena---such as gas leaks, radiation, or ocean salinity---that are not uniformly distributed. For underwater monitoring, ergodic exploration has been used for adaptive sampling of scalar fields using a gliding robotic fish~\citep{ennasr_ergodic_2018}. Gaussian process regression is used to estimate the field and update an expected information density, while ergodic control combines two-dimensional trajectory optimization with spiral motion to sample the full water column. More generally, ergodic control allows a mobile sensor network to persistently monitor these environments, leveraging both prior data collected from on-board sensors and feedback from human operators (e.g., \citet{meyer_scale-invariant_2023,schlafly_collaborative_2024}), naturally balancing the exploitation of known hotspots with the exploration of unmapped regions. In wildfire monitoring, for example, smoke not only evolves over time but also changes the visibility of the on-board sensors. By repeatedly updating the expected information distribution according to smoke diffusion and the corresponding sensor visibility, a team of UAVs can prioritize regions where informative measurements can currently be obtained while continuing to monitor an evolving wildfire environment~\citep{wittemyer_multi-agent_2025}.

Persistent monitoring further requires the robots to account for their available energy and the need to periodically recharge. Adaptive ergodic search has been combined with energy-aware scheduling to coordinate a team of rechargeable aerial robots and a shared mobile charging station, allowing information gathering to continue as the underlying spatiotemporal process evolves~\citep{naveed_adaptive_2025}. In this setting, the desired search distribution can be constructed from the uncertainty accumulated since a region was last observed, while the scheduling component determines when individual robots should temporarily leave the monitoring task to recharge.

For large-scale environmental monitoring, the spatial process of interest may be both unknown and time-varying, requiring the robots to estimate the information distribution while simultaneously exploring it. A distributed HEDAC-based framework has been developed for multi-robot estimation and monitoring of unknown, time-varying spatial processes in complex non-convex environments, with real-world UAV experiments demonstrating fully on-board planning and coordination using limited communication~\citep{mantovani_distributed_2026}. 

Finally, many environmental phenomena evolve according to flow fields that affect both the information distribution and the motion of the sensing platforms. This is especially relevant for oceanographic monitoring, where currents deform the search domain and constrain the motion of underactuated autonomous systems. Ergodic coverage over generalized motion fields explicitly incorporates this flow-induced evolution into the coverage objective, enabling information gathering over dynamic domains while preserving long-term coverage guarantees~\citep{hughes_asymptotically_2026}. Such formulations have been demonstrated for ocean exploration and the tracking of moving populations, as well as through physical experiments with aerial and legged robotic platforms, further extending ergodic information gathering to dynamic, flow-dominated field environments.

\section{Active Perception}

Most of the applications of ergodic control in active perception follows the standard procedure of specifying the ergodic control method as shown in the previous chapters, with the focusing of developing an effective information utility formulations that are spatially-dependent as the target distribution. 

In~\citet{miller_ergodic_2016}, ergodic control is applied to an underwater robot platform with an electrolocation sensory module for a target localization task. Inspired by weakly electric fish such as the black ghost knifefish, the electrolocation sensory module is an active sensory module, similar to LiDARs, that actively generates a weak electric field and measures disturbances within. Such sensory module design is particular suitable for low-velocity robots operating in environments where conventional vision-based sensory modules are ineffective due to factors such as darkness or clutters (e.g., muddy waters). On the other hand, electrolocation sensing is a near-field sensing mechanism with inherent ambiguity in the sensor measurements. Compared to conventional far-field sensors such as cameras, near-field sensors particularly motivates the need of active perception, as the near-field sensory leads to sparse distribution of information, making it necessary to explicitly consider the movements of the sensor as part of the perception process. To tackle this active perception task with electrolocation sensor, the target distribution of ergodic control in~\citet{miller_ergodic_2016} is formulated as the Fisher information-based expected information density (EID), a widely used information utility metric in optimal experimental design. By modeling the sensory module as a probabilistic distribution of the sensor measurement, conditioned on the sensor state and parameterized by the target location, Fisher information evaluates the amount of information provided by a potential measurement at a specific sensor state regarding the target location. Intuitively, Fisher information quantifies how ``quickly'' the likelihood of the parameter of interest---in this case, the target location---changes given a measurement. As a result, EID-based metric encourages the sensor to visit the nearby areas of the estimated target location, instead of simply visiting areas with high likelihood of the target location. Compared to three other baselines that greedily maximize the information utility and one baseline based on random walk, ergodic control consistently outperforms other methods in estimation accuracy, especially when in the scenarios with multiple targets, and is the one method that achieves 100\% success rate across tasks, thanks to the asymptotic coverage guarantee of ergodic control. 

The application of ergodic control for target localization has also been extended in later works, with focuses on real-time receding-horizon control optimization~\citep{mavrommati_real-time_2018} and decentralized control optimization~\citep{abraham_decentralized_2018}. Other than Fisher information, another commonly used information utility for ergodic control is entropy, or related, mutual information (expected entropy reduction). Entropy-based utility formulation is particular suitable for perception models based on Gaussian processes (GPs)~\citep{rasmussen_gaussian_2006}, since the entropy of the GP posteriors can be computed in closed-form. In~\citet{coffin_multi-agent_2022}, the mutual information formulation for ergodic control is further extended for low-information, binary sensors. Beyond the information utility, the sensing footprint itself can also be incorporated into the ergodic formulation, including dynamic footprints whose shape and resolution vary with the robot state~\citep{kwon_volumetric_2025, zheng_ergodic_2026}.

Another near-field sensory modality that plays a critical role in robotics is tactile sensing. The generation of tactile sensory measurements is intrinsically linked to the motion of tactile receptors, and it has long been established that human hand movements are both “purposeful” and “exploratory,” serving a central function in tactile perception~\citep{lederman_hand_1987}. Moreover, the significance of active movement for tactile perception extends beyond humans and has been widely documented in other species. In particular, for whisker-based tactile sensing in rats, both body movements and vibrissal motion have been demonstrated to be essential for perception~\citep{kleinfeld_active_2006}.

In~\citet{abraham_ergodic_2017}, ergodic control is used for shape estimation with a binary sensor for low-resolution tactile perception. The target distribution of ergodic control is formulated as the Fisher information-based expected information density, with a Gaussian process (GP) being the non-parametric model of the shapes of the environment. In the follow-up work~\citep{abraham_data-driven_2018}, the purpose of active perception is extended to estimating the binary sensor measurement model itself, and ergodic control is applied for active localization using the estimated measurement model in sparse environments. \citet{ketchum_active_2024} expands on this idea further by using a conditional variational antoencoder (CVAE) to estimate a richer set of tactile features, such as textures and object geometries, and use the spatial distribution of the predictive entropy as the target distribution for ergodic control. 

While active perception methods are mainly developed as control strategies for robotic tasks, a different perceptive on the use of active perception methods is to study and provide mechanistic models for biological behaviors. It has been a long interest to study the search-related movements of animals through the perspective of active perception, with examples include infotaxis~\citep{vergassola_infotaxis_2007}, gain adaptation, spectral whitening, and high-pass filtering. However, these methods do not incorporate the energetic cost of movements, thus often fail to produce realistic movements and perform poorly in naturalistic conditions. On the other hand, ergodic control as an active perception strategy intrinsically incorporates the control sequences and the nonlinear dynamics as the energetic cost. In~\citet{chen_tuning_2020}, ergodic control is implemented as a energy-constrained proportional betting strategy for study the sensing movements across different species, and is shown to generate more realistic movements compared to standard approaches. 

In \citet{dressel_optimality_2018}, it is pointed out that one specification of an information acquisition task, in which ergodic control is the optimal strategy, is submodularity regarding repeated measurements of the same region. In other words, the accumulated information density of a region decreases at a fixed rate with each sensory measurement taken from the region. Repeated measurements play an important role in practical robot applications, such as the loop closure in SLAM problems. While existing submodular optimization methods are often developed for discrete space, ergodic control has the advantage of optimizing in the continuous time and space domain subject to the nonlinear dynamics of the sensor. 

Lastly, active perception is also closely related to belief-space planning~\citep{kaelbling_planning_1998, platt_belief_2010}, optimal experimental design~\citep{pukelsheim_optimal_1993}, and Bayesian optimization~\citep{jones_efficient_1998}, which provide general frameworks for quantifying the value of potential sensory measurements under uncertainty. These frameworks are not necessarily alternatives to ergodic control; rather, they can be used to construct its target distribution. For example, a belief maintained by a POMDP or belief-space planning method can be mapped to a spatial distribution of expected information~\citep{kaelbling_planning_1998, platt_belief_2010}, while criteria from optimal experimental design---such as A-, D-, and E-optimality---can be used to evaluate the information provided by measurements at different sensor states~\citep{pukelsheim_optimal_1993, miller_ergodic_2016, abraham_ergodic_2021}. Similarly, acquisition functions from Bayesian optimization can define spatial utilities based on quantities such as uncertainty, expected improvement, or expected reduction in posterior uncertainty~\citep{jones_efficient_1998, frazier_knowledge-gradient_2008, hennig_entropy_2012}. Ergodic control then addresses the complementary problem of converting such an information utility into a dynamically feasible sensing trajectory. Unlike methods that greedily select the next most informative measurement, ergodic control optimizes the long-term visitation statistics of the trajectory, enabling non-myopic reasoning over multiple measurements. Its proportional coverage behavior encourages repeated sampling of highly informative regions while preserving exploration of other regions, allowing the sensor to escape local optima in non-convex information landscapes.

\section{Manipulation}

Everyday tasks such as opening a locked door by inserting a key into the keyhole may seem very easy for many but are hard to achieve with standard robot controllers. The main reason is that such a manipulation task cannot be solved by simply tracking a trajectory reference. If we analyze the way we achieve this task more carefully, we can observe that while we will likely not reach the keyhole in one straight shot. We are still capable of doing the task by skillfully moving around the expected location of the keyhole, which can take more or less time depending on the size of the region we need to search in. Indeed, if we open a door that we do not know well, if it is in the dark, or if the door handle occludes the location of the keyhole, will have impact on the uncertainty of our initial guess. We are often very smart at handling this variation by searching for the keyhole with a motion adapted to the situation. Solving this underlying coverage problem is typically done unconsciously. 

\begin{figure}[t!]
    \centering
    \includegraphics[width=1.0\linewidth]{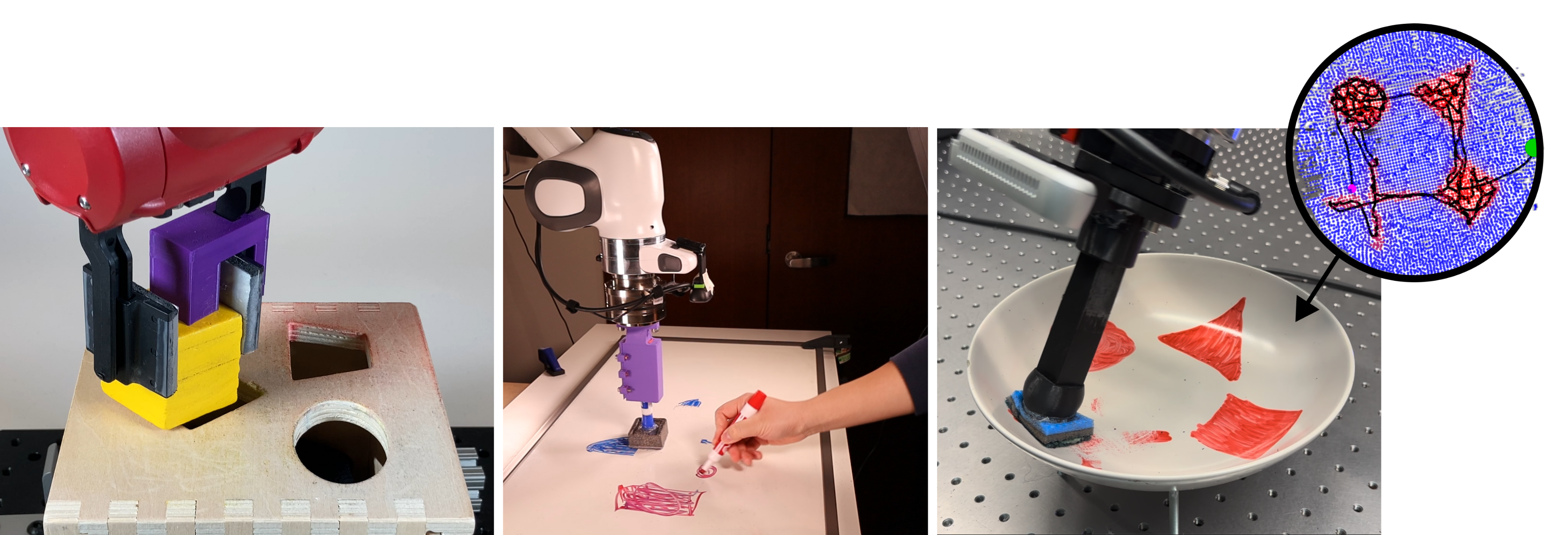}
    \vspace{-1em}
    \caption{Examples of ergodic control applications in manipulation. From left to right: (1) Learning insertion policies from human demonstrations; (2) Reactive erasing in dynamic environments; (3) Erasing over curved surfaces with tactile feedback.}
    \label{fig:application_manipulation}
\end{figure}

This simple example already shows that for such tasks, \emph{tracking a distribution} is more appropriate than tracking a reference point. If we examine this task even further, we can also realize that searching for the keyhole requires an exploration behavior combining both position and orientation information, which can potentially be coupled. Indeed, searching for the keyhole while maintaining a wrong orientation of the key during the search would not be successful. Similarly to the position of the key, the range of orientation will be influenced by several sources of uncertainty, including the orientation of the key held in our hand and the orientation of the door. Vision here is not the only sensing information that we exploit here: we instead combine adaptively multiple sources including tactile sensing. Moreover, the task does not stop at kinematic level. Indeed, the dissection of this example task can also be extended to dynamics, as we also require a skillful usage of compliance to smoothly achieve the task, which can also be extrapolated to an exploration/coverage problem in the space of possible impedance parameters that can be used by a robot to insert the key with varying degree of stiffness.

The above example represents a recurring challenge of robot manipulation, that can be more generally related to the achievement of skillful tasks despite poor sensing and localization, poor estimate of the tools or objects held i the hand/gripper of the robot, poor modeling of the gripper contact with the object that can easily slide/shift, being hard to be estimated by vision due to the occluding hand/fingertips. For many robotics platforms, these challenges are exacerbated by also including poor actuation capability in the list. The rapid spread of humanoid platforms currently being developed will be particularly faced with this challenge, as this legged design choice for mobile manipulation will typically involve lightweight hands with flexible links, limited actuation capability and/or sensorless fingers.

When doing a robot manipulation experiment, if a robot fails at inserting a key into a keyhole, the naive deduction is that we need more expensive cameras to increase the precision of a reference point and/or that we need more expensive actuators to precisely track a reference trajectory. Our claim is that, rather than improving the hardware, there is already a long way to go to improve the control strategies that we are familiar with to go beyond the widespread---but naive---concept of tracking a reference. We believe that ergodic control as a coverage problem formulation is a key component to move research forward. In other words, we should praise the poor sensing and actuation capability of the current hardware to help us shed new lights on the control strategies required to robustly solve manipulation tasks with robots. 

In addition to the hardware/sensing aspects, manipulation inherently requires contact with the environment, which gives a non-negligible advantage to the use of torque-controlled robots (when possible), or robots that can be controlled to provide some form of compliance (active or passive compliance, admittance control, etc.). In such a case, the concept of tracking a reference trajectory can be even worse if naive controllers are employed, as the compliance will produce imprecise reproduction of trajectories with a delay and smoothness that can have direct impact on task success. The increase of the tracking gains is a quick fix that hinders the more essential problem of defining what the robot has to do at first place. Here again, it can be beneficial to specify the task that the robot has to do in the form of a coverage problem rather than a specific reference to reach. 

In \citet{shetty_ergodic_2022}, the Siemens Gears benchmark was used to assess the capability of ergodic control, formulated as a spectral multiscale coverage (SMC) problem, to be used for the gear shaft insertion part, which is representative of a generic peg-in-hole problem in robotics. In this work, the distribution was provided from human demonstration, by observing a user solve the insertion task through kinesthetic teaching. The co-variations observed for the position and orientation of the robot gripper holding the gear shaft were used as inputs to SMC for the reproduction of the task. This skill transfer strategy provides a very intuitive interface to program robots by demonstrating, within a few seconds of continuous recording, the expected range of co-variations that the robot needs to search for the entry point when reproducing the task. Note here that the expected range is not necessarily unimodal (e.g. Gaussian distribution). Indeed, depending on the shapes or symmetries of the pegs or shafts, insertion tasks often offer different options for insertion, forming a multimodal distribution with separated options (e.g., a square peg can be inserted in 4 different configurations). Even in the case of a unimodal distribution, the correlations between the variables often matter, so that the search is position and orientation is coordinated instead of exploring the different variables separately. A unimodal distribution in this case could for example be encoded as a Gaussian with full covariance matrix (instead of an isotropic Gaussian with diagonal covariance, corresponding to uncorrelated variables). 

Standard SMC cannot be used in practice when handling a 6D exploration problem. This is because the Fourier decomposition that needs to be considered along all dimensions does not scale well to problems with more than 3 dimensions. In \citet{shetty_ergodic_2022}, it was proposed to handle this issue through a tensor factorization approach, which allows 6D exploration problems to be solved by exploiting low-rank decomposition techniques that allows the 6D problem to be computed fast while keeping the most essential coordination patterns between the variables.

The proposed SMC approach using tensor factorization was compared to two common baselines: a spiral search and a point sampling strategy. The first baseline consists of pre-defining a geometric pattern followed by the robot to cover a desired area. For rectangular distributions in 2D, the pattern can take the form of a regular back and forth sweep (a.k.a.\ lawnmower strategy). For circular or elliptical distributions in 2D, the pattern can take the form of an ellipse. While easy and convenient to specify in the 2D case, such patterns can hardly be extended to more dimensions, such as the 6D problem treated in the peg-in-hole task that requires a joint distribution in both position and orientation space. The second baseline consists of sampling points in the distribution and connecting the points (a.k.a.\ traveling salesman problem). Compared to these two widely used baselines, ergodic control could solve the peg-in-hole problem faster. This work also showed the importance of using a controller instead of a planner, which can both rely on the same SMC cost. Indeed, the robot in the experiment was controlled with an impedance controller, in order to allow the stiff shaft to be moved in contact with the stiff gear components. This compliant control strategy introduces tracking errors due to unmodeled friction between the components, whose effect was largely toned down by using a myopic ergodic control strategy that was re-evaluating at each time step which next control commands to select to improve the coverage. The approach presented in \citet{shetty_ergodic_2022} was compared to a planning problem using the same cost function, which would pre-compute the ergodic control path, further tracked by the torque controller. The results indicated that an ergodic control strategy with a recomputation strategy (myopic or with a time window as in  model predictive control) was beneficial in manipulation and insertion problems.

In \citet{bilaloglu_whole-body_2023}, the ergodic control problem for manipulation was extended from targeting a specific point on the kinematic chain, for example the 6D gripper pose, to a coverage problem that exploits the entire surface of the robot body. This extension enables the use of additional sensing modalities such as tactile skin distributed across the arm, thereby greatly increasing the effective sensor footprint in tactile exploration tasks. The work demonstrated that formulating ergodic control through diffusion (HEDAC) provides a more effective framework for handling complex sensor footprints and for flexibly balancing local and global exploration. In this setting, the target distribution corresponds to an implicit representation of a surface with an unknown shape or pose in the three-dimensional workspace of the robot. An alternative scenario is to use an explicit representation of the surface, for example a mesh or a point cloud, and to perform exploration constrained to the known surface.

In many manipulation tasks, accurately modeling the physical interaction between a robot and an object is prohibitively complex. For example, in cleaning, achieving sufficient coverage depends on factors such as material properties and the type of debris, which are often unknown or difficult to quantify. As a result, motion planning under such uncertainty is prone to failure. To address this, \citet{bilaloglu_tactile_2025} proposed reframing the problem as a closed-loop ergodic control task that mirrors how humans rely on continuous visual and tactile feedback rather than detailed preplanning. The method captures the object’s surface geometry and the evolving dirt distribution as a point cloud using an RGB-D camera and then guides the robot through ergodic control, ensuring that dirtier regions are revisited more frequently. This closed-loop tactile ergodic control strategy enables robots to adapt online, overcome modeling limitations, and achieve effective coverage even on complex curved surfaces.

In other scenarios, directly measuring coverage with a camera is challenging, as in surface inspection, sanding, or mechanical palpation. In such cases, cleaning can serve as a proxy task in which a human expert marks regions of interest with an easy-to-remove marker, allowing the robot's progress to be monitored visually. This strategy can enable closed-loop ergodic control to facilitate intuitive human–robot interaction, using expert annotations and markings to guide tactile robotic tasks. 

\section{Human-Robot Interaction}
Rapid development of exoskeletons, quadrupeds, and humanoids have put robots physically closer to humans than ever before, generating both higher risk and opportunities for shared human-robot activities. Although healthy human participants can be modeled as energy-minimizing or minimum jerk under tightly controlled experimental conditions. We generally cannot expect the even the same individual to follow the same trajectory from one iteration of a task to the next. The variance in trajectories will generally increase with more diversity in the pool of users, uncertainty in the environment, and when the users are non-experts. Probabilistic representations of human motions enable one to capture this natural variation such that ergodic metrics can be used to analyze the quality of human motion, provide robot support as needed, and even control multiple robot agents. Using ergodic metrics relaxes the requirements on the human users in providing task specifications or demonstrations when the robot is receiving information from the human. When the robot acts in a supportive role to the human---the robot communicates to the human via physical interactions---ergodic metrics enable the autonomy to correctly identify what support is needed. Ergodic shared control can then support a range of normative movements without overrestricting the joint human-robot trajectory.

\begin{figure}[t!]
    \centering
    \includegraphics[width=1.0\linewidth]{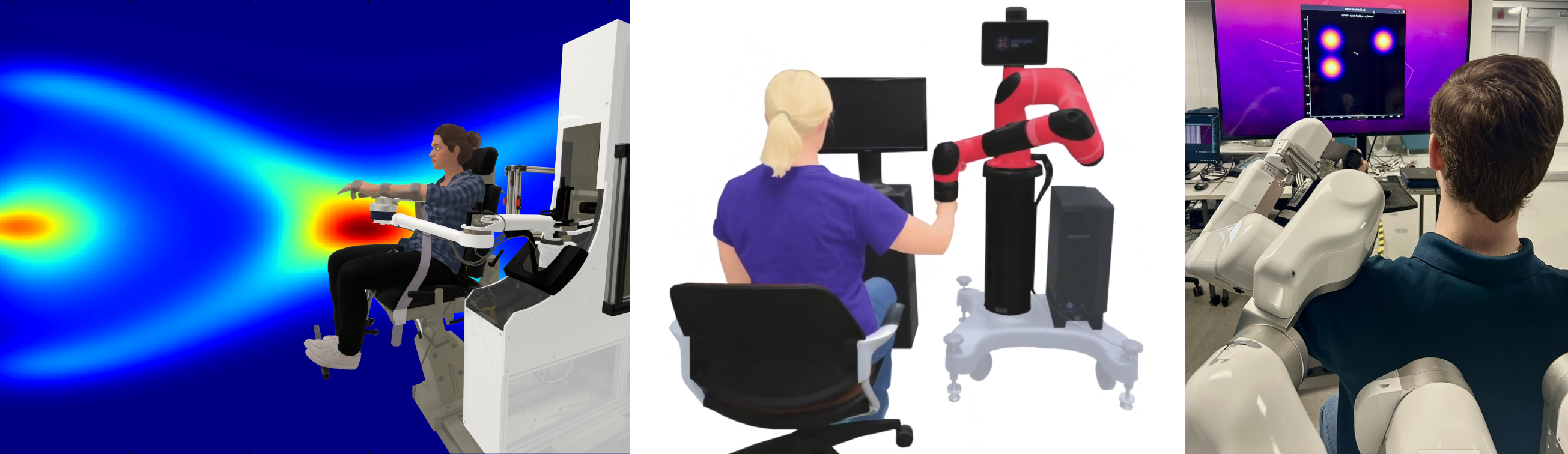}
    \vspace{-1em}
    \caption{Examples of ergodic control applications in human-robot interaction. From left to right: (1) Ergodicity as a measure of task-relevant information in patients' motion during rehabilitation; (2) Ergodic control ensures safety while maintaining adaptivity in physical therapy; (3) Ergodic control for learning from physical demonstration.}
    \label{fig:application_hri}
\end{figure}

For simple tasks, users can provide high level control objectives to robot agents by specifying  a complete trajectory via kinesthetic teaching or providing waypoints on a map. However, as with direct control, providing waypoints becomes slower and more cumbersome for the human user as the complexity of the task and the environment increases. Waypoints or trajectories provided also do not fully represent the possible set of good solutions to the task. If the user is providing information about locations in the environment where data is to be collected by the robot agent, a trajectory or set of waypoints, does not capture the high-level task objective and lacks information about areas where the user knows there will not be useful data. 
\citet{meyer_scale-invariant_2023} have developed a method for capturing user specifications using a touchscreen interface in which inputs are translated into swarm control via spectral decomposition. The resulting distributions can use information about areas the user is interested in and areas to avoid. Leveraging the decentralized formulation of ergodic control developed in~\citep{abraham_decentralized_2018} (discussed in~\ref{sec:decentral}) enables the user to control the swarm with real-time updates regardless of the number of agents. In simulations and field tests in the DARPA OFFSET field setting, this human-swarm interaction approach demonstrated that performance degraded gracefully as agents were lost. Furthermore, \citet{popovic_measuring_2023} showed that this approach reduced the burden on users as it required fewer instructions compared to waypoint methods. Waypoint commands grow proportional to the swarm size while ergodic specification is scale invariant such that users can shift their focus from operating the swarm to other tasks in the environment---enhancing the performance of the human-swarm team overall. Specifically, when swarms provide task-relevant information about the environment, shifting the level of autonomy from waypoint to coverage-based controllers improves human decision-making~\citep{schlafly_collaborative_2024}. Although cognitive availability is highest when the swarm operates fully autonomously based on environmental information only, sharing control (using environmental information and user specification to specify coverage goals) was the only paradigm that actually improved the performance of the human-swarm team in the given task. Augmenting environmental information with user specifications provides a balanced approach to reduce cognitive load while leveraging human knowledge of high level goals and priorities.

We can also use ergodic control to leverage human teachers in the context of imitation learning. Many learning from demonstrations paradigm assume that the teacher provides `expert' or `near-optimal' demonstrations of a task. Yet assistive robots are being developed to be deployed on factory floors, healthcare facilities, and in people's homes, where they would be most beneficial if end users are able to customize their behavior for specific tasks. Providing high quality demonstrations may be challenging, so ergodic imitation learning aims use imperfect demonstrations to learn a reference distribution rather than a reference trajectory~\citep{kalinowska_ergodic_2021}. This enables each demonstration to add information such that the learner can actually outperform an individual demonstration. Demonstrations are added using a weighted average of the Fourier coefficients computed as part of the spectral method for measuring ergodicity. This unique learning strategy also enables the use of demonstrations of `what not to do' or negative demonstrations. These demonstrations can highlight regions where a collision is likely or capture preferences, and in some cases may be easier for the user to provide. In the case of obstacle avoidance, it might take several positive demonstrations to learn the area that the robot should avoid. Using a single demonstration trajectory that spends time at the obstacle, the reference distribution can be updated by applying a negative weight to the obstacle-seeking demonstration in the weighted average of distributions. Leveraging this unique capability to add both negative and positive demonstrations, ergodic imitation learned was used by \citet{pang_ergodic_2025} to learn from a human's physical corrections (positive demonstrations) as well as the behavior that was being corrected (negative demonstrations). Incorporating both the corrected trajectories and originally planned trajectories improved learner performance, since both explicit and implicit information from the correction was used to update the reference distribution.

Ergodic metrics can also play a significant role in physical human-robot interactions, since they can enable the robot partner to identify when it should intervene. Especially in applications like assistive robotics and rehabilitation robotics, the correct thing for the robot to do to support the human is nothing. For therapeutic tasks, the patient must have a high level of engagement and actively participate in executing a task for the benefits to translate to skill acquisition and retention outside the clinical setting. This is challenging for typical measures such as error from a normative trajectory, since two equally succesful task executions may have significant differences between them at each time point due to time misalignment, multiple good task solutions, and the natural variation that we expect in human motion. \citet{fitzsimons_ergodicity_2019} demonstrated that ergodic metrics could capture differences in skill level and performance differences due to learning. While task-specific metrics of success could differentiate users of different skill levels and trajectory error showed significant but subtle changes due to learning, an ergodic metric  was able to capture both phenomena statistically. This introduced the idea that ergodic measures reflect the information encoded in motion about the task, such that higher skill users have more information about the task and that assistance adds information. Given this, ergodic metrics are potentially useful for closing the loop on pHRI, especially when the goal is motor learning. Ergodic metrics enable the controller to allow variability during practice repetitions as in \citet{fitzsimons_ergodic_2022}, where ergodic shared control improved training compared to an error-based shared control framework.

\begin{figure}[t!]
    \centering
    \includegraphics[width=1.0\linewidth]{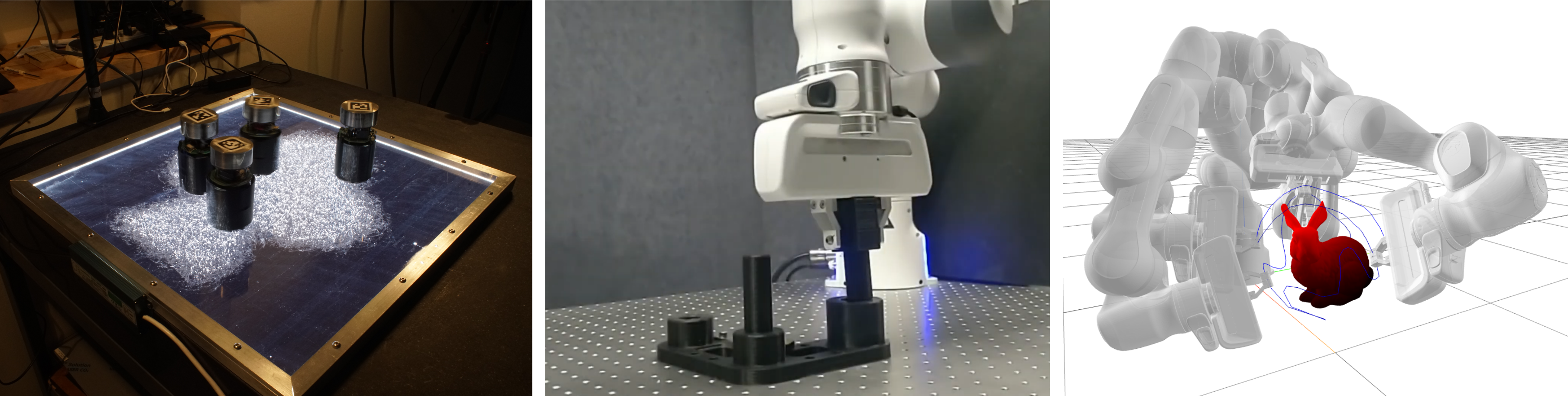}
    \vspace{-1em}
    \caption{Examples of ergodic control applications in manufacturing. From left to right: (1) Micro-patterned surface manufacturing with swarm robots; (2) Learning assembly from demonstrations; (3) Autonomous inspection of manufactured components.}
    \label{fig:application_manufacturing}
\end{figure}

\section{Other Applications} 

Many processes in manufacturing can be framed as spatial coverage problems including milling or surface finishing (e.g. sanding, polishing, and painting) where a surface must be covered according to a specific density or material deposition rate. Ergodic control provides a framework to generate continuous, dynamically feasible toolpaths that deposit material or apply finishing tools proportionally to a target spatial distribution.  \citet{bilaloglu_tactile_2025} demonstrated that by applying ergodic covereage directly to point clouds, tactile coverage akin to sanding or polishing could be achieved on curved surfaces. The robot was able to maintain contact with the surface and exert the desired force profile while dynamically adjusting its coverage based on real-time visual feedback. Such a system can even be enhanced with learned preferences from an expert human~\citep{schneyer_ergodic_2026}. Another manufacturing-related application of ergodic control is surface micro-patterning, the process of incorporating arrays of structural features (e.g., dimples and grooves) to enhance functionalities beyond the material's bulk properties, including friction control, drag reduction, and anti-fouling properties. Mobile, multi-robot systems have great potential in manufacturing large-scale micro-patterned surfaces, as mobile systems are not constrained by the scale of the manufacturing equipment. \citet{taylor_image_2024} designed a prototype mobile robot system integrated with ergodic control for density-specified patterning of micro-structured surfaces, and this system was later extended to multi-robot systems and was shown to enable effective friction reduction~\citep{landis_large-scale_2025, taylor_manufacturing_2026}.

A related tactile coverage problem arises in minimally invasive surgery, where limited access restricts direct palpation of subsurface abnormalities. \citet{beber_autonomous_2026} combined online force-based estimation of tissue elasticity with Gaussian Process Regression and a HEDAC controller driven by an expected information density incorporating model uncertainty, tissue stiffness, and spatial stiffness gradients. Experiments on a silicone tissue phantom showed that the resulting adaptive palpation trajectories could reconstruct the elasticity map and detect stiff inclusions, demonstrating the potential of ergodic exploration for autonomous tissue characterization in robot-assisted minimally invasive surgery.

\begin{figure}[ht!]
    \centering
    \includegraphics[width=1.0\linewidth]{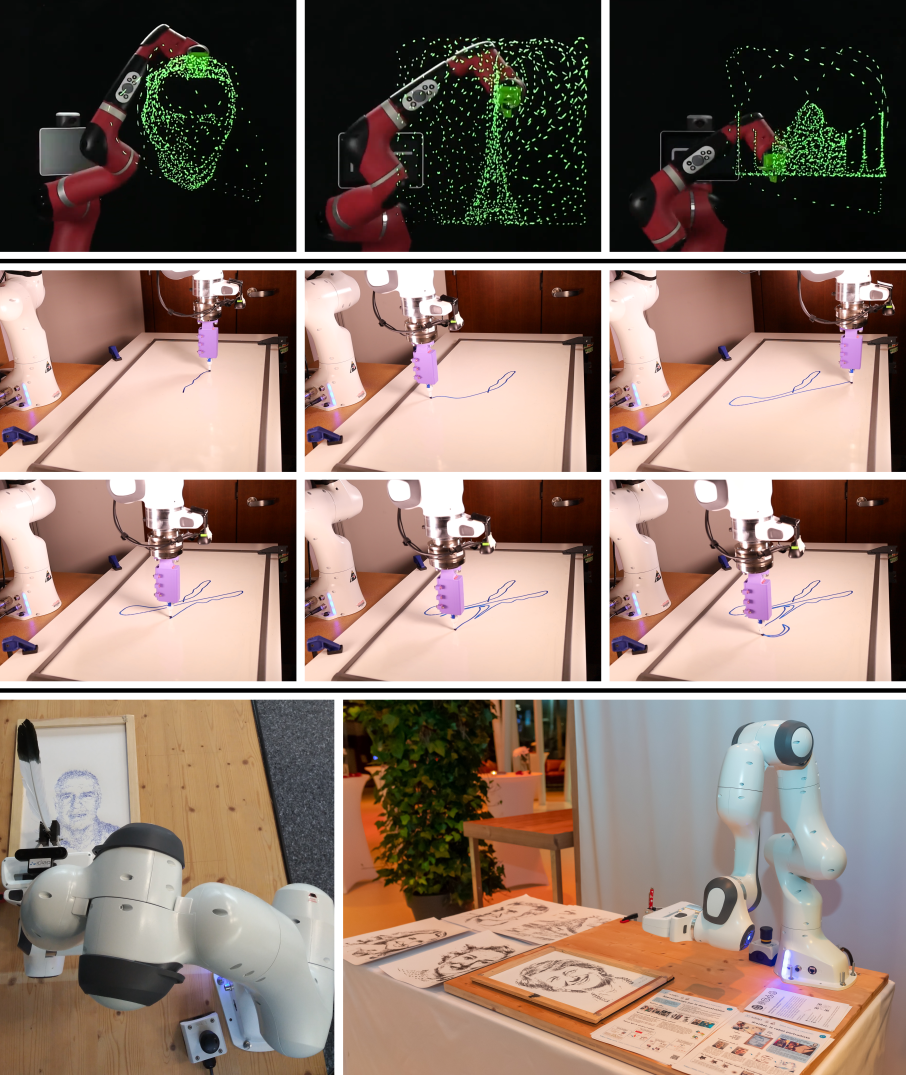}
    \vspace{-1em}
    \caption{Examples of ergodic control applications in robot art.}
    \label{fig:application_art}
\end{figure}

While drawing and erasing are common benchmark tasks in the ergodic control literature~\citep{prabhakar_autonomous_2020, fitzsimons_ergodic_2022, sun_flow_2025, kwon_volumetric_2025, pang_ergodic_2025}, the framework is less frequently applied to generate portraits for purely aesthetic purposes. In the realm of robot art, \citet{low_drozbot_2022} introduced the ``drozBot'' system, which treats an image's pixel intensity as a spatial probability distribution. By applying ergodic control, the system synthesizes a continuous trajectory for a robotic marker. The resulting variations in line density recreate the original image, demonstrating how mathematical coverage formulations can yield compelling and aesthetically pleasing artistic renderings.

\chapter{Controlled Diffusion for Perception Learning}

\section{Overview}

In the era of generative models, the capability of capturing complicated data distributions and generating new samples from the learned distribution has advanced significantly. For robotics, this offers exciting opportunities for modeling sensors, dynamics, and environments with a level of fidelity that is not achievable by conventional modeling approaches.

However, there is a key difference between the training of generative models for standard tasks such as language and image generation versus the training of generative models for robots. For standard tasks, the focus is on designing the model architecture such that the model can accurately extract the underlying distribution of existing large-scale datasets. On the other hand, robot learning problems do not have the luxury of large-scale datasets readily available. Furthermore, given the multifaceted nature of robotics tasks, it is also unclear whether it is practically possible to create large-scale datasets for all different kinds of robotics tasks. For example, while text and image generation are challenging tasks, the tasks themselves are well-structured, and the dataset format is unified, which plays a central role in the creation of existing large-scale datasets. But for robotic tasks, different tasks involve different robot dynamics, sensors, different kinds of environmental interactions, and different types of environments themselves. Therefore, to leverage the power of generative models for robot learning, we not only need to address the model capability aspect of the problem, as in text and image generation, but also, equally importantly, address the problem of data collection.

While the above analysis questions the feasibility of creating unified large-scale datasets for robot learning, robots, as embodied agents with the capability of actively interacting with the environment, offer another solution to the data collection challenge. Instead of focusing on creating unified, large-scale datasets before training the model, robots offer a platform to incrementally create specialized datasets concurrently with model training, and provide the embodied agency to close the feedback loop of the process, such that the robot can actively interact with the environment to generate the most informative data to accelerate the training process. We name this process closed-loop data collection.

Closed-loop data collection is a decision-making process. Similar to conventional decision-making problems for robots, it involves choosing the optimal control sequences under the robot’s dynamic constraints with respect to a task-relevant reward function. Unlike these problems, for which the reward function is a state-based metric related to task performance (e.g., tracking error), in closed-loop data collection, the reward function evaluates the quality of the generated data as the outcome of the robot’s actions with respect to the model at the current training stage. While there exists a wide range of metrics for evaluating dataset quality, such as the i.i.d. property and mutual information, what is common among them is that the metrics focus on the distribution that the generated data represents. This shift in task evaluation between conventional robot motion planning and closed-loop data collection makes the broad family of distribution-based motion synthesis approaches— in particular, ergodic control—a natural solution to the decision-making problem that is closed-loop generation.

The value of ergodic control in robot learning is twofold:
(1) It enables non-myopic search behavior over information distributions;
(2) It generates trajectories that converge to a stationary distribution, decorrelating states and producing i.i.d. data for machine learning.
Our core argument for using ergodic control in robot learning is that non-myopic behavior alone—often achieved heuristically by other methods—is not sufficient. We must also systematically synthesize the statistical properties of the data. In other words, distribution-driven coverage and decorrelation inherently lead to non-myopic search behavior, but the reverse is not necessarily true.

In this chapter, we will discuss the use of ergodic control for closed-loop data collection in in robot perception learning. In the next chapter, we will discuss its use in embodied reinforcement learning.

\section{What Is Robot Perception Learning?}

Robot perception learning refers to the task of learning the sensor-objective relationship, with the sensor subject to the robot’s movements. Since ImageNet~\citep{deng_imagenet_2009}, learning-based pipelines have become the dominant approaches for reasoning over perception inputs, especially images, to extract a rich set of information from the sensor inputs, such as classification labels, semantic segmentation, and potential action plans.

However, applying conventional learning pipelines developed for image data—that is, methods that rely on training with large-scale datasets prior to the task—to robot perception learning faces two main challenges. First, compared to conventional image-based tasks, robot tasks more often encounter scenarios that are drastically different from the training data. This is particularly true because sensory inputs in robot perception depend on the robot's motion, geometry, and dynamics. These factors introduce additional sources of variation and further widen the distribution gap between training and testing, calling for a \emph{robot-centric} rather than human-centric design principle for perception models. Second, robotic tasks involve perception channels beyond vision, for which large-scale training data may be unavailable and realistic simulation tools may not exist to generate synthetic data. One possible solution is to leverage vision to emulate other sensing modalities, such as vision-based tactile perception~\citep{yuan_gelsight_2017, lambeta_digitizing_2024, goncalves_punyo-1_2022}. However, these approaches not only limit the fundamental capabilities of the sensing modality itself~\citep{kim_heterogeneous_2020, pattabiraman_eflesh_2025, he_deep_2026}, but also fail to generalize to many perception channels that are valuable for robotics. Olfactory perception is a notable example, as neither vision-based methods nor realistic simulation tools are currently available~\citep{zhang_advanced_2026, liu_scensory_2026, bassil_scalable_2026}. Lastly, the novelty of the sensory input can come not only from the inherent design of the sensor, but also from the damage to the sensor or changes in sensor or environmental properties (e.g., tactile sensing)~\citep{avtges_damage_2026}.

While we have seen significant progress in robot perception learning to address the above challenges, such as new model architectures that can generalize across more diverse training data and low-cost platforms that collect training data more quickly~\citep{chi_universal_2024, thakkar_clamp_2025, zorin_ruka_2025}, a more principled approach to address the fundamental needs of robot perception learning is through online learning with closed-loop data collection. This principle also aligns with how animals interact with and adapt to their environments, where learning is continual and kinetic energy is allocated not only to carry out actions for survival tasks but also to purposefully interact with the environment in order to generate more data~\citep{vergassola_infotaxis_2007, stamper_active_2012, yang_active_2016, goldshtein_acoustic_2024}.

Consider an arbitrary robot, with knowledge of its own kinematic model, equipped with an unknown sensor and a learning architecture that is able to extract information from past sensor observations. How should the robot act to build a perception model between the sensor and the environment from scratch, without external intervention? To answer this question, there are two steps. The first step is to identify what architecture the learning model should use and what information it should extract from past observations. The second step is to synthesize the motion of the robot in response to the information extracted from the current data by the model in order to generate more sensory inputs to further stimulate and refine the model.

Since there is no external supervision, the learning needs to be self-supervised. The sensory input is subject to the robot’s motion, thus the learning of the perception model needs to be conditioned on the robot’s state, and potentially velocity and acceleration. Combining these two requirements, we consider a dataset $\mathcal{D}=\{ d_i \}_{i=1,\dots,N}$ where each data point $d_i=(x_i, y_i)$ consists of a sensor measurement $y_i$ and the corresponding robot state $x_i$ where the measurement is generated. The perception model is formulated as a generative model that predicts the anticipated sensor measurements given a robot state, based on the dataset containing previous observations. We denote such a model as $p_{\theta}(y|x)$---interpreted as a probability distribution of sensor measurements conditioned on the robot state---and formulate the training process through the maximum likelihood estimation (MLE) formula:
\begin{align}
    \theta^* & = \argmax_{\theta} \frac{1}{N} \sum_{i=1}^{N} \log p_{\theta} (y_i | x_i) .
\end{align} Once learned, the conditional generative model can support several downstream perception and decision-making tasks. The model can be used directly to predict sensor measurements at unobserved robot states~\citep{finn_deep_2017}, reconstruct incomplete observations~\citep{sohn_learning_2015}, detect novel objects or anomalous sensor responses~\citep{zollicoffer_novelty_2025, jiang_anomalies-by-synthesis_2025}, and identify changes in the sensor or environment through deviations between predicted and measured observations~\citep{chen_learning---drive_2024, lupu_magicvfm-meta-learning_2025}. Because the model captures how sensory inputs vary with the robot state, it can also serve as a measurement model for state estimation, belief-space planning, active perception, and control. Furthermore, the autonomously collected dataset and the representations learned by the generative model can be used to adapt existing perception architectures to the specific robot, sensor, and environment. In particular, they can provide robot-specific data for fine-tuning large pretrained perception, multimodal, or vision-language-action models~\citep{kim_openvla_2025}, allowing the capabilities of large-scale offline training to be combined with online adaptation to previously unseen sensing conditions.

\section{A Case for Conditional Variational Autoencoders (CVAEs)}

While the the general formulation of the robot perception learning is compatible with arbitrary choice of generative models, a particularly common choice for the generative perception model in practice is the conditional variational autoencoder (CVAE)~\citep{sohn_learning_2015}---an extension of the widely used variational autoencoder (VAE)~\citep{kingma_auto-encoding_2013} architecture. 

\subsection{Overview of CVAEs}

A CVAE introduces a latent variable $z$ to represent variations in the sensor measurements that are not directly specified by the conditional robot state $x$. The generative model is formulated as
\begin{align}
    p_{\theta}(y|x) = \int p_{\theta}(y|z,x)p(z)\,dz,
\end{align} where $p(z)$ is a prior distribution over the latent variable, commonly chosen to be a standard Gaussian distribution, and $p_{\theta}(y|z,x)$ is a conditional decoder that predicts the sensor measurement distribution given the robot state and latent variable. Under this formulation, the generative process first samples a latent variable $z\sim p(z)$ and then generates a sensor measurement $y\sim p_{\theta}(y|z,x)$ conditioned on the robot state.

Similar to the general generative perception model introduced in the previous section, the CVAE could in principle be trained by maximizing the conditional data likelihood:
\begin{align}
    \theta^* & = \argmax_{\theta} \frac{1}{N} \sum_{i=1}^{N} \log p_{\theta} (y_i|x_i) \\
    & = \argmax_{\theta} \frac{1}{N} \sum_{i=1}^{N} \log \int p_{\theta}(y_i|z,x_i)p(z) dz.
\end{align} However, directly solving this maximum likelihood estimation problem requires marginalizing the decoder over all possible latent variables. For nonlinear decoder architectures such as neural networks, this integral is generally intractable. Furthermore, the true posterior distribution
\begin{align}
    p_{\theta}(z|x,y) = \frac{p_{\theta}(y|z,x)p(z)} {p_{\theta}(y|x)}
\end{align} is also intractable because its normalization requires evaluating the same marginal likelihood.

To address this challenge, \citet{kingma_auto-encoding_2013} introduces a variational encoder $q_{\phi}(z|x,y)$ to approximate the intractable posterior $p_{\theta}(z|x,y)$. The encoder takes both the robot state and the observed sensor measurement as inputs and outputs a distribution over the corresponding latent representation. In a common formulation, this distribution is modeled as a diagonal Gaussian:
\begin{align}
    q_{\phi}(z|x,y) = \mathcal{N} \left( \mu_{\phi}(x,y), \operatorname{diag} \left( \sigma_{\phi}^{2}(x,y) \right) \right).
\end{align} Instead of directly maximizing the generally intractable conditional likelihood, the CVAE jointly optimizes the encoder and decoder by maximizing a tractable lower bound on the conditional log-likelihood, known as the evidence lower bound (ELBO). Applying Jensen's inequality gives
\begin{align}
    \log p_{\theta}(y|x) & = \log \int q_{\phi}(z|x,y) \frac{ p_{\theta} (y|z,x)p(z) }{ q_{\phi}(z|x,y) } dz \\
    & \geq \underbrace{\mathbb{E}_{z\sim q_{\phi}(z|x,y)} \left[ \log p_{\theta}(y|z,x) \right] - D_{\mathrm{KL}} \left( q_{\phi}(z|x,y) \| p(z) \right)}_{\mathcal{L}_{\mathrm{ELBO}} (\theta,\phi;x,y)}.
\end{align}
The gap between the conditional log-likelihood and the ELBO is
\begin{align}
    \log p_{\theta}(y|x) - \mathcal{L}_{\mathrm{ELBO}}
    (\theta,\phi;x,y) = D_{\mathrm{KL}} \left( q_{\phi}(z|x,y) \|  p_{\theta}(z|x,y) \right),
\end{align} which is nonnegative. Therefore, maximizing the ELBO simultaneously improves the conditional data likelihood and encourages the variational encoder to approximate the true latent posterior. More specifically, the first term of the ELBO,
\begin{align} 
    \mathbb{E}_{z\sim q_{\phi}(z|x,y)} \left[ \log p_{\theta}(y|z,x) \right],
\end{align} evaluates the reconstruction error, which encourages the decoder to accurately reproduce the observed sensor measurement from the robot state and its latent representation. The second term,
\begin{align}
    D_{\mathrm{KL}} \left( q_{\phi}(z|x,y)  \| p(z) \right),
\end{align} regularizes the latent representation by encouraging the encoded distributions to remain close to the latent prior. The resulting learning objective over the collected dataset is
\begin{align}
    \theta^*,\phi^* = \argmax_{\theta,\phi} \frac{1}{N} \sum_{i=1}^{N} \mathcal{L}_{\mathrm{ELBO}} (\theta,\phi;x_i,y_i).
\end{align} In practice, latent samples can be differentiated through using the reparameterization
\begin{align}
    z = \mu_{\phi}(x,y) + \sigma_{\phi}(x,y) \cdot \epsilon, \qquad  \epsilon \sim \mathcal{N}(0,I),
\end{align} enabling the encoder and decoder to be trained jointly through standard gradient-based backpropagation.

\subsection{Benefits of CVAEs}

The latent-variable formulation of CVAEs provides several advantages over other generative model architectures, such as the diffusion models and flow-based models, that are particularly useful for closed-loop data collection and robot perception learning.

First, the latent variable $z$ enables the model to separate properties of the observations of the environment from variations caused by the robot state. Given a measurement $y$ collected at state $x$, the encoder extracts a latent representation $z\sim q_{\phi}(\cdot|x,y)$ of information that is not fully explained by the conditional robot state itself. The same latent representation $z$ can then be passed to the decoder at a different candidate state $\hat{x}$ through $y \sim p_{\theta}(\cdot|z,\hat{x})$. Conceptually, this allows the robot to ask what the same object or environmental feature would look like from another sensor configuration. For example, consider a robot equipped with a camera observing a rubber duck from a particular viewpoint. The robot state $x$ specifies the camera pose, while the latent variable $z$ can encode properties of the observed rubber duck, such as its color and geometry. After observing the rubber duck from one camera pose, the robot can combine the inferred latent representation with a different candidate pose $\hat{x}$ to predict how the same object would appear from that new viewpoint. In contrast, a model formulated only as $p_{\theta}(y|x)$ must represent all sensor measurements that could occur at a given state without conditioning on the particular object or feature currently being observed. The latent variable therefore provides a mechanism for maintaining perceptual context while reasoning across different robot states.

Second, CVAEs are computationally well suited for closed-loop data collection because both model training and uncertainty evaluation can be performed on the same time scale as motion synthesis. The CVAE supports incremental online training through stochastic gradient updates of the ELBO as new sensor measurements are collected. At inference time, latent estimation and sensor prediction require only a single forward pass through the encoder and decoder, respectively. When the decoder output is modeled using a tractable distribution, such as a Gaussian distribution with learned mean and variance, the predictive entropy
\begin{align}
    h(x) = \mathbb{H}_{y}[p_{\theta}(y|z,x)], \qquad z\sim p(z)
\end{align}
can also be evaluated analytically. Therefore, data collection, model training, uncertainty evaluation, and motion synthesis can occur concurrently within the same closed-loop process. In contrast, generative models such as diffusion models and flow-based models commonly generate samples through iterative denoising processes or numerical integration. Estimating predictive uncertainty may require generating multiple such samples at every candidate robot state, creating a significant computational bottleneck when the information distribution must be repeatedly evaluated across the robot's reachable state space. CVAEs avoid this iterative sampling process, making them particularly suitable for online closed-loop decision-making in which the evolving perception model must inform the robot's motion in real time.

Third, the latent space learned by a CVAE provides a compact perception representation that can be reused for downstream tasks. While the decoder represents how sensor measurements vary with robot state, the latent variable captures recurring structure across the collected observations. Measurements generated by the same object or environmental feature can therefore be compared and grouped in the latent space even when their raw sensor values differ due to changes in viewpoint or robot configuration. This enables the learned model to support tasks such as feature clustering, object fingerprinting, classification, and object identification without requiring a separate supervised representation-learning pipeline. In this sense, the CVAE serves not only as a predictive sensor model for guiding data collection, but also as a self-supervised feature-learning architecture whose internal representation remains useful after the data collection process is complete. This capability is not inherently available in generative architectures that do not learn an explicit encoder into a structured latent space, for which a separate representation-learning model may be required to support the same downstream perception tasks.

Lastly, the aforementioned advantages do not imply that the CVAE must serve as the final perception model after the data collection process is complete. From the perspective of reconstruction fidelity and predictive accuracy, CVAEs may be limited compared to more expressive generative architectures, such as diffusion models and flow-based models, which have been introduced in part to better represent complicated and high-dimensional data distributions. Instead, the particular value of the CVAE in closed-loop data collection is that it serves as an efficient intermediate model that can be continually updated and queried while the robot is operating. Its evolving predictive uncertainty guides the motion synthesis process toward informative regions of the state space, while its latent representation provides self-supervised features that can be used to organize, cluster, or annotate the collected data. The primary output of this process is therefore not necessarily the CVAE itself, but a high-quality dataset whose distribution has been actively shaped by the robot's interaction with the environment. Once data collection is complete, more expressive or task-specific models can be trained on this dataset for downstream tasks that require higher reconstruction fidelity, predictive accuracy, or specialized perception capabilities.


\section{I.I.D. Property, Ergodicity, and PAC-Bayes Bound}

We have now established why the CVAE is a suitable intermediate perception model for guiding closed-loop data collection. The remaining question is how the robot should use the information provided by this model to make data-collection decisions. Unlike conventional control problems, in which the objective may be defined directly through quantities such as trajectory-tracking error, the objective of data collection is less immediate: the robot must choose actions according to the statistical quality and information content of the dataset those actions will generate. In this section, we argue that ergodic control provides a suitable template for formulating this decision-making problem for two main reasons. First, from the perspective of probably approximately correct (PAC) learning, we show that ergodic control enables the robot to generate trajectories with stable long-term visitation statistics, which are important for relating a finite, trajectory-generated dataset to an underlying data distribution. Second, it allows the desired allocation of sensing effort to be specified directly as a target distribution over the robot state space, enabling the robot’s behavior to adapt to the evolving information provided by the perception model.

\subsection{PAC-Bayes Generalization Bounds}

A model is learned using a finite dataset, but its objective is to perform well on future data, assumed to be generated by the same underlying process. Statistical learning theory formalizes this distinction through the concept of \emph{empirical risk}---model performance on the observed training data---and the \emph{population risk}---the expected performance of the model on the underlying data distribution. Tools from statistical learning theory, such as the PAC-Bayes bound, have been increasingly applied in robot learning, both as certification tools and as optimization objectives, to learn policies with provable generalization guarantees to unseen environments~\citep{majumdar_pac-bayes_2021, veer_probably_2021, ren_generalization_2021, agarwal_stronger_2022, sharma_pac-bayes_2023}.

Let $\mathcal{H}$ denote a hypothesis space, where each hypothesis $w\in\mathcal{H}$ represents a possible model (e.g., a set of neural network parameters). Let $\rho$ denote the data-generating distribution and let $\ell(w;d)$ denote the loss of hypothesis $w$ on a data sample $d$. Given a finite dataset $\mathcal{D}=\{d_i\}_{i=1}^{N}$, the population and empirical risks of a hypothesis $w$ are defined as:
\begin{align}
    R_{\rho}(w) &= \mathbb{E}_{d\sim\rho} \left[ \ell(w;d) \right],\\
    \widehat{R}_{\mathcal{D}}(w) &= \frac{1}{N} \sum_{i=1}^{N} \ell(w;d_i).
\end{align} The central question is whether a hypothesis with low empirical risk also has low population risk.

The PAC-Bayes framework addresses this question by considering probability distributions over the hypothesis space. Let $P(w)$ denote a prior distribution over hypotheses, chosen independently of the observed dataset, and let $Q_{\mathcal{D}}(w)$ denote a posterior distribution selected after observing the dataset $\mathcal{D}$.\footnote{The term ``posterior'' here does not necessarily refer to an exact Bayesian posterior; it may be any data-dependent distribution over possible hypotheses.} PAC-Bayes evaluates the expected performance of a randomized predictor obtained by sampling $w\sim Q_{\mathcal{D}}$. Its population and empirical risks are defined as:
\begin{align}
    R_{\rho}(Q_{\mathcal{D}}) &= \mathbb{E}_{w\sim Q_{\mathcal{D}}} \mathbb{E}_{d\sim\rho} \left[ \ell(w;d) \right],\\
    \widehat{R}_{\mathcal{D}}(Q_{\mathcal{D}}) &= \mathbb{E}_{w\sim  Q_{\mathcal{D}}} \left[ \frac{1}{N} \sum_{i=1}^{N} \ell(w;d_i) \right].
\end{align} Assuming the loss $l(w;d)$ is a bounded function, a typical PAC-Bayes bound has the schematic form
\begin{align}
    R_{\rho}(Q_{\mathcal{D}}) \lesssim \widehat{R}_{\mathcal{D}}(Q_{\mathcal{D}}) + \sqrt{ \frac{ D_{\mathrm{KL}} \left( Q_{\mathcal{D}}\|P \right) + \log(1/\delta) }{N} },
\end{align} which holds with probability at least $1-\delta$ over the sampled training dataset. The precise form of the bound depends on the chosen PAC-Bayes theorem, but the underlying interpretation is consistent.

The bound balances three quantities. The empirical-risk term favors hypotheses that explain the observed data well. The KL-divergence term $D_{\mathrm{KL}} ( Q_{\mathcal{D}}\|P )$ measures how much the learned posterior departs from the prior and therefore quantifies the amount of model specialization introduced by the training data. Finally, the generalization penalty term:
\begin{align}
    \sqrt{ \frac{ D_{\mathrm{KL}} \left( Q_{\mathcal{D}}\|P \right) + \log(1/\delta) }{N} }
\end{align} decreases as the number of samples increases. PAC-Bayes therefore formalizes the principle that a model is more likely to generalize when it achieves low training loss without becoming excessively specialized to a finite dataset.

The bound does not prescribe a particular data-generating distribution $\rho$. However, the empirical and population risks must refer to the same underlying distribution. Crucially, standard PAC-Bayes results establish this connection by assuming the dataset contains i.i.d. samples from the data generation distribution:
\begin{align}
    d_1,\ldots,d_N \overset{\mathrm{i.i.d.}} \sim \rho.
\end{align} The i.i.d. condition ensures that every training sample represents the same population, while independence enables the empirical risk to concentrate around the corresponding population risk. Therefore, the statistical properties of the data-generation process are central to establishing a meaningful generalization guarantee.

\subsubsection{PAC-Bayes Bounds for CVAEs}

CVAEs provide a particularly concrete instantiation of the PAC-Bayes framework because the latent-variable formulation naturally provides a prior distribution, a data-dependent posterior distribution, and a KL-divergence regularization term. Unlike the generic formulation above, in which hypotheses may represent complete sets of model parameters, the hypothesis space in the CVAE-specific construction is the latent space $\mathcal{Z}$. The latent prior $p(z)$ serves as the PAC-Bayes prior, while the encoder distribution $q_{\phi}(z|x,y)$ serves as a data-dependent posterior for each observed state-measurement pair.

Given a latent hypothesis $z\sim q_{\phi}(\cdot|x,y)$, the decoder generates a prediction conditioned on the robot state. Let $g_{\theta}(z,x)$ denote the predicted sensor measurement, and define the reconstruction loss as
\begin{align}
    \ell_{\theta}(z,x,y) = \left\| y-g_{\theta}(z,x) \right\|.
\end{align}
The corresponding population reconstruction risk is
\begin{align}
    R_{\rho}(\theta,\phi) = \mathbb{E}_{(x,y)\sim\rho} \mathbb{E}_{z\sim q_{\phi}(\cdot|x,y)} \left[ \ell_{\theta} (z,x,y) \right],
\end{align}
while the empirical risk over the collected dataset
$\mathcal{D}=\{(x_i,y_i)\}_{i=1}^{N}$ is
\begin{align}
    \widehat{R}_{\mathcal{D}}(\theta,\phi) = \frac{1}{N} \sum_{i=1}^{N} \mathbb{E}_{z\sim q_{\phi}(z|x_i,y_i)} \left[ \ell_{\theta}(z,x_i,y_i) \right].
\end{align}

Under suitable regularity assumptions, including Lipschitz continuity of the encoder and decoder, a PAC-Bayes bound can be established for the CVAE reconstruction risk. Schematically, the result takes the form
\begin{align}
    R_{\rho}(\theta,\phi) \lesssim \widehat{R}_{\mathcal{D}}(\theta,\phi) + \frac{1}{\lambda} \sum_{i=1}^{N} D_{\mathrm{KL}} \left( q_{\phi}(z|x_i,y_i) \| p(z) \right) + \mathcal{C},
\end{align}
where $\lambda>0$ is a bound parameter and $\mathcal{C}$ collects the remaining finite-sample, confidence, and continuity-dependent terms. Because the decoder is conditioned on the robot state $x$, the CVAE bound also contains additional terms that account for how the prediction and reconstruction loss vary across conditional states.

This result directly connects the CVAE training objective to population-level generalization. The empirical reconstruction risk corresponds to the reconstruction term of the ELBO, while $D_{\mathrm{KL}} ( q_{\phi}(z|x,y) \| p(z) )$  serves both as latent regularization and as a complexity term in the PAC-Bayes bound. Thus, maximizing the ELBO improves empirical reconstruction while controlling quantities that bound expected reconstruction performance. This gives CVAEs an additional advantage for closed-loop data collection: their probabilistic structure naturally provides the prior, posterior, and tractable KL-divergence required for a concrete generalization bound.

\subsection{Ergodicity in Sequential Data Collection}

In the generic PAC-Bayes formulation, the result relies on assumptions about the statistical properties of the training dataset. In particular, the collected state-measurement pairs must represent i.i.d. samples from a fixed underlying data distribution. However, in closed-loop robot data collection, measurements are generated sequentially along a dynamically constrained trajectory, which compromises the i.i.d. assumption crucial for the generalization of the learned perception model.

Denote the control policy that governs robot data collection as $\pi(u|x)$ and the control induced state visitation distribution as $p_{\pi}(x)$. The data generation distribution of the robot data collection process can be written as:
\begin{align}
    \rho_{\pi}(x, y) = p(y | x) p_{\pi}(x), 
\end{align} where $p(y|x)$ is the measurement model of the sensor. More importantly, consecutive samples from the distribution $\rho_{\pi}(x, y)$ are not independent, but instead correlated by the physical dynamics of the robot. As a result, for a finite dataset $\mathcal{D}_{N}=\{(x_i,y_i)\}_{i=1}^{N}$, the joint distribution of the collected samples generally does not factor as
\begin{align}
    p(\mathcal{D}_{N}) = \prod_{i=1}^{N}\rho_{\pi}(x_i,y_i),
\end{align} and the PAC-Bayes bound cannot be applied directly. The compromise of the independence property is fundamentally tied to the embodied nature of the robot. Since the robot must move continuously under its physical dynamics, it cannot generally draw arbitrary independent samples from the state space. 

On the other hand, the control policy determines the long-term state-visitation statistics of the robot. Under a fixed policy $\pi$, the robot state evolves as a controlled Markov process. If this process is ergodic, it admits a unique invariant distribution $p_{\pi}(x)$. An invariant distribution, also called a stationary distribution, is a distribution that remains unchanged under the state-transition dynamics. Specifically, if $q_{\pi}(x' \mid x)$ denotes the transition kernel induced by the control policy, then the state-visitation distribution $p_{\pi}$ is invariant if
\begin{align}
    p_{\pi}(x') = \int q_{\pi}(x' | x) p_{\pi}(x) dx.
\end{align} Thus, if the current state is distributed according to $p_{\pi}$, the state remains distributed according to $p_{\pi}$ after the next transition. Although ergodicity cannot eliminate the temporal dependence between consecutive
samples, it can induce stable long-term sampling behavior so that the collected dataset represents a well-defined underlying distribution. The significance of ergodicity can be formalized by Birkhoff's ergodic theorem. Suppose that the state-measurement process generated under $\pi$ is ergodic with invariant distribution $\rho_{\pi}(x,y)$. Then, for any integrable function $f(x,y)$, we have
\begin{align}
    \lim_{N\rightarrow\infty} \frac{1}{N} \sum_{i=1}^{N} f(x_i,y_i) = \mathbb{E}_{(x,y)\sim\rho_{\pi}}  [ f(x,y) ].
\end{align} Thus, even though the samples remain temporally correlated, empirical quantities computed from a single robot trajectory converge to their population counterparts under a stationary data-generating distribution. In particular, by taking $f(x,y)$ to be the reconstruction loss of the perception model, Birkhoff's ergodic theorem establishes that the empirical reconstruction risk appearing in the PAC-Bayes analysis converges to the corresponding population risk under $\rho_{\pi}$. Ergodicity therefore provides the statistical connection between the loss evaluated on the sequentially collected dataset and the expected loss that the PAC-Bayes bound seeks to control.

The first significant advantage of ergodic control for closed-loop perception learning is therefore that it regulates the long-term visitation statistics of the robot. Rather than eliminating the unavoidable temporal correlations introduced by the robot's embodied and dynamically constrained motion, ergodic control ensures that the empirical state-visitation distribution converges to a well-defined stationary distribution. This makes population-level learning performance meaningful and provides a foundation for extending the PAC-Bayes analysis to sequentially collected data.

\section{Ergodic Control for Perception Learning}

While ergodicity establishes an asymptotic connection between the empirical risk computed from sequentially collected data and the population risk appearing in the PAC-Bayes bound, in practice, we must also specify the spatial distribution toward which the robot motion should converge. This requires the control policy not only to exhibit ergodic stability, but also to synthesize control with respect to arbitrary target distributions. This is the second advantage of ergodic control for closed-loop robot data collection: it not only induces ergodicity, but also allows the distribution toward which the robot motion converges to adapt as the perception model evolves during data collection and model regression.

Concurrently with the regression of the generative model over the current dataset, the predictive uncertainty of the model indicates varying levels of confidence across the state space. Closed-loop data collection therefore requires the robot to synthesize motion that provides both sufficient variety in the collected data and high information value. Since the perception model is conditioned on the robot state, a natural choice for the target information distribution is the predictive uncertainty, such as entropy, conditioned on the robot state. Ergodic control can then synthesize non-myopic coverage with respect to this distribution. Compared to a random walk, ergodic control improves data-collection efficiency by biasing coverage toward regions with higher information density. Compared to greedy information maximization, it preserves coverage of the broader state space and enables the robot to escape regions with locally optimal information density.

\citet{prabhakar_mechanical_2022} studies this non-myopic search capability of ergodic control for perception learning. Compared to a greedy information-maximization strategy, ergodic control generates a more comprehensive dataset across the search space. This capability is particularly important when the environment contains multiple regions with distinctive features. In such cases, information maximization can repeatedly sample a locally informative region and fail to investigate other regions of interest, whereas ergodic control guarantees the discovery of features across the space given sufficient time. Compared to a random sampling-based strategy that uniformly covers the space, ergodic control produces better predictive accuracy with lower energy consumption. The improvement in predictive accuracy arises because information is unevenly distributed across the space; uniform coverage can therefore generate a dataset whose sampling distribution is disproportionate to the information distribution of the environment. The reduction in energy consumption arises from the tendency of ergodic control to bias the robot's motion toward regions with high information density. As a result, the robot spends less time exploring regions with lower information value while still guaranteeing asymptotic coverage of the space.

\citet{pinosky_embodied_2024} experimentally demonstrates the effectiveness of ergodic control for closed-loop perception learning through hardware experiments. Across all active exploration trials, the learned CVAEs maintained information-rich latent spaces, and the collected data rapidly converged toward the conditional entropy distributions of the models. In contrast, random-walk exploration led to latent-space collapse in two of the five trials and produced data that did not match the target information distribution. The learned models successfully identified all objects in their training environments and more than \(75\%\) of the objects in previously unseen test environments. The complete learning process was performed online on hardware within approximately 15 minutes and without any pre-existing dataset.


\chapter{Controlled Diffusion for Reinforcement Learning}

\section{Overview}
Embodied reinforcement learning is a process where an embodied agent learns an action policy from experience. The action policy is often modeled as a probability distribution of actions conditioned on the current state of the robot. The quality of an action policy is evaluated based on a reward function defined by the task. The evolution of the agent state is governed by a state transition model, which captures the dynamics of the agent and the interaction between the agent and the environment. The overall goal of reinforcement learning is to learn an optimal action policy under the reward function from the past experience of interacting with the environment.

Over the past decade, reinforcement learning has seen significant progress in robotics and has become one of the most used approaches for developing control and motion planning strategies. It has proven especially effective for robots in complex, dynamic, and unstructured environments, and has seen success across a wide range of tasks, such as locomotion and manipulation. On the other hand, reinforcement learning algorithms are still limited by several fundamental issues, some of which are inherently related to the embodied nature of robotic systems. One of the most significant bottlenecks for reinforcement learning on embodied agents is related to sampling in the physical space. RL algorithms require the agent to gather experience as training data through interaction with the environment, and these algorithms are built on the assumption that such interactions lead to i.i.d. data. However, while the i.i.d. assumption often holds—at least to a certain extent—for conventional non-embodied agents such as in text and image generation, this assumption is fundamentally against the nature of embodied agents, leading to a significant performance gap in RL on embodied agents.

Embodied agents, due to their physical nature, move continuously across space and time, inherently introducing correlations between their interactions with the environment. While numerous techniques exist to address this issue for embodied RL, these techniques are only technical tricks that address the problem at a surface level. To address the issue fundamentally, we need to re-think the data collection process for embodied RL as a motion synthesis problem with the explicit goal of decorrelating states.

\section{Preliminary}

\subsection{Problem Formulation}

Denote a Markov decision process (MDP) as $\mathcal{M} = (\S, \A, p, r, \gamma, \rho)$, where $\S$ denotes the state space, $\A$ denotes the action space, $p:\S\times\A\mapsto\Delta(\S)$ denotes the probabilistic state transition model ($\Delta(\S)$ denotes the space of all probability distributions over $\A$), $r:\S\times\A\mapsto\Real{}$ denotes the reward function, $\gamma\in[0, 1)$ is the discount factor, and $\rho$ is the prior distribution of the initial state of the system. 

We denote a trajectory as $\tau = (s_0, a_0, s_1, a_1, \dots)$, which is a sequence of state-action pairs. Given a policy $\pi$, the policy-induced trajectory distribution, denoted as $p^{\pi}(\tau)$, is defined as:
\begin{align}
    p^{\pi}(\tau) = \rho(s_0) \prod_{t=0}^{T} \pi(a_t | s_t) p(s_{t+1} | s_t, a_t). \label{eq:traj_distr}
\end{align}

The problem of reinforcement learning (RL) is to find a policy $\pi:\S\times\A\mapsto\Delta(\A)$ that optimizes the objective function $J(\pi)$ that is the expected discounted reward:
\begin{align}
    J(\pi) = \E_{\tau\sim p^{\pi}(\cdot)} \left[ \sum_{t=0}^{T} \gamma^{t} r(s_t, a_t) \right]. 
\end{align}

For a given $\pi$, the evaluation of the policy's future expected discount reward is often through the Q function $Q^{\pi}(s, a)$---based on the current state-action pair---and the value function $V^{\pi}(s)$---based on the current state:
\begin{gather}
    Q^{\pi}(s, a) = \E_{\tau\sim p^{\pi}(\cdot)} \left[ \sum_{t=0}^{T} \gamma^{t} r(s_t, a_t) \Big| s_0=s, a_0=a \right], \\
    V^{\pi}(s) = \E_{a\sim\pi(\cdot|s)} \left[ Q^{\pi}(s, a) \right].
\end{gather} The Q-function and value function provide alternative representations of the decision-making process, from which an optimal policy can be derived. 

\subsection{Overview of RL Methods}

We provide an overview of commonly used RL methods by categorizing them according to which components of the decision-making process are learned from data, specifically those represented by neural networks and trained through regression.

\noindent\textbf{Learning the policy. } NNs can be used to directly represent a parameterized policy $\pi_{\theta}(a|s)$. To optimize and learn the policy in this case, derivative of the policy with respect to the expected discounted reward, often approximated from samples using stochastic policy rollouts, is needed. Representative methods in this category are the policy gradient methods, such as REINFORCE~\citep{williams_simple_1992} and trust region policy optimization (TRPO)~\citep{schulman_trust_2015}. While these methods benefit from the simplicity of the formula and the resulting effectiveness during deployment, learning such policies might be data-intensive. 

\noindent\textbf{Learning the value function. } NNs are also used to represent the value function (or Q function), in which case policy is formulated as the optimization of the Q- or value-function. Representative methods from this category are the Q-learning methods, such as Deep Q-Networks (DQN)~\citep{mnih_human-level_2015}. The most common way to learn a value function is through temporal-difference (TD) learning~\citep{sutton_learning_1988}. In TD learning, the learned function is trained against a bootsrapped target formed by combining the immediate reward with a estimate of future value from the learned function itself. Value-based representation improves the data-efficiency of the reinforcement learning, but reliable learning of the value function remains an open challenge. 

\noindent\textbf{Learning the dynamics. } As another alternative, NNs can be used to represent the dynamics of the system or environment. In this case, the policy is not represented directly; instead, it is implicitly obtained by solving an online model-based control or planning problem using the learned representations. Methods for policy optimization include model-based control---such as model predictive path integral (MPPI) control~\citep{williams_information_2017, chua_deep_2018}---and planning methods---such as Monte Carlo tree search~\citep{schrittwieser_mastering_2020}. These methods offer advantages in data efficiency, flexibility, and adaptation. However, policy performance depends heavily on model accuracy, and the computation required for runtime optimization can be prohibitively expensive for real-time applications. Both challenges become more severe in long-horizon decision-making problems.

More importantly, more recent and currently widely used RL methods often combine these techniques by using neural networks (NNs) to represent multiple components of the decision-making process. We discuss two major families of such methods.

\noindent\textbf{Actor-critic methods.} A major family of such methods is actor-critic methods. The core idea is to represent both the policy and a value function, or (Q)-function, with NNs and train them iteratively so that each facilitates the learning of the other: the learned value function provides a more informative learning signal for improving the policy, while the learned policy determines the data distribution used to learn the value function. Although actor-critic methods share this general framework, they differ in how the policy is updated. One common approach uses the learned value function to estimate advantages for policy-gradient updates, with representative methods including proximal policy optimization (PPO)~\citep{schulman_proximal_2017}, advantage actor-critic (A2C), and asynchronous advantage actor-critic (A3C)~\citep{mnih_asynchronous_2016}. These methods require data generated by the current policy and are therefore referred to as \emph{on-policy methods}. Another common approach optimizes the policy using the learned (Q)-function, with representative methods including soft actor-critic (SAC)~\citep{haarnoja_soft_2018}, deep deterministic policy gradient (DDPG)~\citep{lillicrap_continuous_2019}, and Twin Delayed DDPG (TD3)~\citep{fujimoto_addressing_2018}. Unlike on-policy methods, these methods can learn from data collected using previous or different policies and are therefore referred to as \emph{off-policy methods}. The actor-critic methods discussed above learn the policy without explicitly modeling the system dynamics, and are therefore referred to as \emph{model-free methods}. On the other hand, some more recent methods, such as Dreamer~\citep{hafner_dream_2019}, also learn a dynamics model of the environment---often referred to as a \emph{world model}---to bootstrap the actor-critic learning process by generating rollouts from the learned model.

\noindent\textbf{Planning-based methods.} Another major family of methods combines multiple learned components by formulating the policy implicitly as an online planning or control problem, while learning auxiliary components to improve the efficiency of that process. Representative examples include AlphaGo~\citep{silver_mastering_2016}, AlphaZero~\citep{silver_general_2018}, and MuZero~\citep{schrittwieser_mastering_2020}. These methods learn both a policy and a value function---and, in the case of MuZero, a latent dynamics model---but instead of directly executing the learned policy at runtime, they use it as an action prior to guide and accelerate Monte Carlo tree search (MCTS). A similar idea has also been extended to continuous-control problems in robotics. For example, temporal-difference model predictive control (TD-MPC)~\citep{hansen_temporal_2022, hansen_td-mpc2_2024} learns both a value function and a latent dynamics model to improve the efficiency and long-horizon performance of online model predictive control.

\section{Data Generation in Reinforcement Learning}

Among all the aforementioned methods, one thing is common: reinforcement learning methods require generating data by interacting with the environment through policy rollouts, using either the current policy or previously learned policies. The collected trajectories are then used to learn the components of the decision-making process, such as the policy, value function, or dynamics model. Given unlimited interaction with the environment and unlimited computational resources, repeatedly rolling out the policy would eventually generate sufficient data for learning. In practice, however, interactions with the environment are often expensive, making the quality and efficiency of the generated data critical to successful learning.

This challenge is commonly referred to as the exploration problem in reinforcement learning. The objective of exploration is to generate informative trajectories that improve learning with as little interaction as possible. A common strategy is to introduce stochasticity into the policy rollout, either by directly perturbing the actions or policy, or by incorporating stochastic behavior into the learning objective itself. Among these approaches, maximum entropy reinforcement learning (MaxEnt RL)~\citep{ziebart_maximum_2008, haarnoja_soft_2018} plays a central role by explicitly encouraging stochastic policies through the optimization objective, leading to more diverse data generation while simultaneously maximizing the expected return.

Note that MaxEnt RL is not a particular algorithm but a general problem formulation and framework associated with a family of different algorithm specifications. Instead of the standard RL objective, MaxEnt RL simultaneously maximizes the expected reward and the action entropy of the policy:
\begin{align}
    \pi^* = \argmax_{\pi} \E_{\tau\sim p^{\pi}(\cdot)} \left[ \sum_{t=0}^{T} \gamma^t \left( r(s_t, a_t) + \alpha \cdot \H\Big(\pi(\cdot|s_t)\Big) \right) \right], 
\end{align} which can be re-written as:
\begin{align}
    \pi^* = \argmax_{\pi} \E_{\tau\sim p^{\pi}(\cdot)} \left[ \sum_{t=0}^{T} \gamma^t \Big( r(s_t, a_t) - \alpha \cdot \log \pi(a_t | s_t) \Big) \right]. \label{eq:maxent_obj}
\end{align}

By explicitly incorporating the entropy term in the objective, MaxEnt RL maintains the stochasticity of the optimal policy, which encourages exploration and in turn produces more diverse data for learning. In particular, the optimal policy under MaxEnt RL can be written in the Boltzmann formula as a distribution:
\begin{align}
    \pi^*(a|s) \propto \exp\left( \frac{1}{\alpha} \cdot Q^*(s, a) \right). \label{eq:maxent_boltsmann}
\end{align} $Q^*(s, a)$ is the optimal Q function and it can be written in the form of Bellman backup as:
\begin{align}
    Q^*(s, a) = r(s,a) + \gamma \cdot E_{s^\prime\sim p(\cdot|x,a)} \left[ \alpha \cdot \log \left( \int \exp\left( \frac{1}{\alpha} \cdot Q^*(a^\prime|s^\prime) \right) da^\prime \right) \right].
\end{align} Therefore, MaxEnt RL provides a direct connection between policy sampling and exploration through the soft optimality distribution induced by the Q-function. In standard RL, policy improvement is based on hard maximization: the policy is encouraged to select the action with the largest Q-value. In MaxEnt RL, this hard maximization is replaced by a soft distribution over actions, where actions with larger Q-values receive higher probability, but actions with comparable Q-values are still sampled with non-negligible probability. Practical algorithms such as SAC implement this idea by jointly learning a soft Q-function and a stochastic policy that approximates the soft optimality distribution induced by the learned critic.

At a higher level, MaxEnt RL can be understood as a strategy for improving data generation by maintaining stochasticity during policy rollouts. Rather than injecting arbitrary action noise, the agent samples actions according to their expected usefulness under the current learning objective, preventing the policy from prematurely collapsing to a narrow behavior based on incomplete or inaccurate value estimates. This action-level view of stochastic data generation is not limited to model-free actor-critic methods such as SAC. Sampling-based model predictive control methods, such as NN-MPPI, follow a related principle at the level of control sequences: they sample stochastic action sequences, evaluate them through model rollouts, and bias future control toward higher-performing samples. Although these methods differ substantially in implementation, they all rely on stochasticity in the action or control generation process to improve exploration and data acquisition. 

From the perspective of data collection, however, RL data consist not only of sampled actions but of state transitions produced by the interaction between the policy and the environment dynamics. Consequently, diverse action or control samples do not necessarily imply diverse or decorrelated state trajectories. In particular, in embodied, constrained, or poorly controllable systems, the dynamics can transform diverse action samples into highly correlated state trajectories. Thus, while MaxEnt RL encourages stochastic action or control sampling, it does not directly guarantee diverse or decorrelated experience in embodied settings.

\section{Maximum-Diffusion Reinforcement Learning}

Instead of maximizing the action entropy of the policy for exploration, for embodied agents, a more suitable objective for physical exploration is the entropy of the state trajectory distribution induced by a policy (\ref{eq:traj_distr}). We denote a state trajectory as $\tau_s = (s_0, s_1, \dots, s_T)$ and a general distribution over state trajectories as $p(\tau_s)$. However, unlike the action sequences, state trajectories are subject to the constraints of the physical dynamics of the embodied agents. To characterize the dynamics constraints of the agent without involving specific control policies, we first define the local velocity fluctuations of the agent with respect a general trajectory distribution $p(\tau_s)$ as follow:
\begin{align}
    \langle \Delta s_t, \Delta s_t^\top \rangle_{s^\prime}  = \E_{\tau_s\sim p(\cdot)} \left[ \sum_{t=0}^{T} \langle \Delta s_t, \Delta s_t^\top \rangle \cdot \delta(s_t - s_t^\prime) \right], \label{eq:vel_fluc}
\end{align} which measures the variety of the state velocity at a particular state $s^\prime$, with the velocity denoted as $\Delta s_t = s_{t+1} - s_t$. To represent the physical dynamics as a constraint, the local velocity fluctuations (\ref{eq:vel_fluc}) is constrained to be equivalent to the autocovariance of the trajectory at any particular state $s^\prime$, denoted as $C[s^\prime]$, given the same state trajectory distribution $p(\tau_s)$:
\begin{align}
    C[s^\prime] = \sum_{k=1}^{K} \mathbb{E}_{\tau_s\sim p(\cdot)} \left[ \left(s_{t+k}-\mu_{k|s^\prime}\right) \left(s_{t+k}-\mu_{k|s^\prime}\right)^\top \,\middle|\, s_t=s^\prime \right], \label{eq:local_autocov}
\end{align}
where
\begin{align}
    \mu_{k|s^\prime} = \mathbb{E}_{\tau_s\sim p(\cdot)}
    \left[ s_{t+k} \mid s_t=s^\prime \right].
\end{align} Therefore, we can formulate the following constrained optimization problem---a maximum calibre problem---for maximizing the state trajectory entropy:
\begin{align}
    p_{max}(\tau_s) & = \argmax_{p(\tau_s)} -\int p(\tau_s) \log p(\tau_s) d\tau_s, \\
    \text{s.t., } & \int p(\tau_s) d\tau_s =1 \\
    & C[s^\prime] = \langle \Delta s_t, \Delta s_t^\top \rangle_{s^\prime}, \forall s^\prime \in \S. 
\end{align} The above problem has an analytical solution to the global optimum~\citep{berrueta_maximum_2024}:
\begin{align}
    p_{max}(\tau_s) & \propto \exp\left( -\frac{1}{2} \sum_{t=0}^{T} \Delta s_t^\top C^{-1}[s_t] \Delta s_t \right) \\
    & = \rho(s_0) \prod_{t=0}^{T} \underbrace{\exp\left( -\frac{1}{2} \Delta s_t^\top C^{-1}[s_t] \Delta s_t \right)}_{\propto p_{max}(s_{t+1} | s_t)} \\
    & = \rho(s_0) \prod_{t=0}^{T} p_{max}(s_{t+1} | s_t). 
\end{align} Note that, $p_{max}(\tau_s)$ does not involve any specific policy, task and reward, it solely describes the state trajectory distribution with maximum entropy under physical dynamics constraints of embodied agents. Furthermore, $p_{max}(s_{t+1} | s_t)$ describes the state transition under the stochastic process under $p_{max}(\tau_s)$:
\begin{align}
    p_{max}(s_{t+1} | s_t) & \propto \exp\left( -\frac{1}{2} \Delta s_t^\top C^{-1}[s_t] \Delta s_t \right) \\
    & = \exp\left( -\frac{1}{2} (s_{t+1}-s_t)^\top C^{-1}[s_t] (s_{t+1}-s_t) \right),
\end{align} which indicates that the evolution of states under $p_{max}(\tau_s)$ is governed by a Gaussian distribution with covariance being the autocovariance of the system trajectory:
\begin{align}
    s_{t+1} = s_t + w_t, \quad w_t \sim \mathcal{N}(w ; 0, C[s_t]). 
\end{align} Therefore, a stochastic process that maximizes the state trajectory entropy is equivalent to a diffusion process.

To incorporate a reward function $r(s, a)$, we can formulate a Boltzmann distribution, similar to MaxEnt RL (\ref{eq:maxent_boltsmann}), based on $p_{max}(\tau_s)$:
\begin{align}
    p_{max}^r(\tau) \propto p_{max}(\tau_s) \cdot \exp\left( \frac{1}{\alpha} \sum_{t=0}^{T} \gamma^t r(s_t, a_t) \right). \label{eq:diffusive_distr}
\end{align} To incorporate a particular policy $\pi$, we first define the state trajectory distribution induced by a policy $\pi$ as:
\begin{align}
    p^{\pi}(\tau_s) = \rho(s_0) \prod_{t=0}^{T} \int p(s_{t+1} | s_t, a_t) \pi(a_t | s_t) da_t.
\end{align} The objective for maximum-diffusion reinforcement learning (MaxDiff RL) is formulated to minimize the KL-divergence between the state-action trajectory distribution (\ref{eq:traj_distr}) $p^{\pi}(\tau)$ and the reward-biased diffusive trajectory distribution (\ref{eq:diffusive_distr}) $p_{max}^r(\tau)$:
\begin{align}
    \pi^* = \argmin_{\pi} D_{KL}\left( p^{\pi}(\tau) \| p_{max}^r(\tau) \right). 
\end{align} Importantly, this objective can be re-written in parallel to the standard maximum-entropy reinforcement learning objective (\ref{eq:maxent_obj}) as follow~\citep{berrueta_maximum_2024}:
\begin{align}
    \pi^* = \argmax_{\pi} \E_{\tau\sim p^{\pi}(\cdot)} \left[ \sum_{t=0}^{T} r(s_t, a_t) + \frac{\alpha}{2} \log\det C_{\pi}[s_t] \right],
\end{align} where $C_{\pi}[s_t]$ is the local velocity fluctuations given the policy-induced state trajectory distribution:
\begin{align}
    C_{\pi}[s^\prime] = \sum_{k=1}^{K} \mathbb{E}_{\tau_s\sim p_{\pi}(\cdot)} \left[ \left(s_{t+k}-\mu_{k|s^\prime}\right) \left(s_{t+k}-\mu_{k|s^\prime}\right)^\top \,\middle|\, s_t=s^\prime \right]. \label{eq:local_autocov_2}
\end{align} Compared to the objective of maximum-entropy reinforcement learning objective (\ref{eq:maxent_obj}), which simultaneously optimizes the reward and the action entropy of the policy, the maximum-diffusion reinforcement learning objective simultaneously optimizes the reward and the local entropy of the policy-induced state trajectory distribution. The term $\log\det C_{\pi}[s_t]$ measures the volume of the local trajectory covariance ellipsoid, thus maximizing the term encourages the agent to spread its state trajectories across many dynamically feasible directions, rather than merely randomizing its actions. As a result, MaxDiff RL encourages the policy to maximize expected reward while encouraging the policy-induced state trajectory distribution to approximate the maximally diffusive trajectory distribution.

\section{Diffusion Process, Ergodicity and I.I.D. Data}

\subsection{What is I.I.D. Data?}

Aside from empirically maximizing the state trajectory entropy and inducing diffusive state evolution, the MaxDiff RL framework has deeper formal implications on controlling the properties of data produced by embodied agents, in particular the independent and identically distributed (i.i.d.) property. 

The i.i.d. property is one of the most ubiquitous assumptions underlying machine learning methods, including reinforcement learning methods. A dataset is i.i.d. when each datum is sampled independently from the other samples and all samples are drawn from the same underlying distribution, which also implies a stationary data distribution. Given a set of i.i.d. data $\mathcal{D}=\{z_i\}_{i=1}^N$ sampled from $p_{\mathcal{D}}(z)$, the probability distribution of the batch factorizes into the product of the probabilities of individual samples:
\begin{align}
p_{\mathcal{D}}(z_1, \dots, z_N) = \prod_{i=1}^{N} p_{\mathcal{D}}(z_i).
\end{align} The practical significance of the i.i.d. assumption can be demonstrated through stochastic gradient descent for batched backpropagation, which is essential in most model regression frameworks. In machine learning, the optimization problem can generally be written as minimizing a loss function $\mathcal{L}(\theta)$ over parameters $\theta$, where the loss is defined as an expectation over the data distribution:
\begin{gather}
\theta^* = \argmin_{\theta} \mathcal{L}(\theta), \
\mathcal{L}(\theta) = \E_{z\sim p_{\mathcal{D}}}[l(\theta; z)].
\end{gather} In batched stochastic gradient descent, the gradient of the loss function $\nabla_{\theta} \mathcal{L}(\theta)$ is approximated using a mini-batch $\{z^\prime_j\}_{i=1}^{B} \subset \mathcal{D}$:
\begin{align}
\nabla{\theta} \mathcal{L}(\theta) \approx \frac{1}{B} \sum_{i=1}^{B} \nabla_{\theta} l(\theta; z^\prime_i), \ z^\prime_i \overset{\text{i.i.d.}}{\sim} \mathcal{D}.
\end{align} When the data is produced i.i.d., the variance of the gradient estimation reduces proportionally to the batch size $B$:
\begin{gather}
    \text{Var}\left[ \frac{1}{B} \sum_{i=1}^{B} g_i \right] = \frac{1}{B} \text{Var}[g_i], \ 
    g_i = \nabla_{\theta} l(\theta; z^\prime_i).
\end{gather} This is based on the fact that i.i.d. data has no correlation between samples:
\begin{align}
    \text{Cov}(z^\prime_i, z^\prime_j) = \text{Cov}(g_i, g_j) = 0, \ \forall i\neq j. 
\end{align} On the other hand, for data that is not produced i.i.d., the variance of the gradient estimation becomes:
\begin{align}
    \text{Var}\left[ \frac{1}{B} \sum_{i=1}^{B} g_i \right] = \frac{1}{B} \text{Var}[g_i] + \frac{2}{B^2} \sum_{i<j} \text{Cov}(g_i, g_j), \label{eq:batch_grad_cov}
\end{align} where the non-zero covariance between the samples compromises the gradient estimation, and in turn, affects the model regression and data efficiency. 

\subsection{The Challenge of Producing I.I.D. Embodied Data}

In reinforcement learning, data are produced sequentially through the interaction between the agent and the environment, forming transition tuples $z_t=(s_t, a_t, r_t, s_{t+1})$. Batched stochastic gradient descent is then used to regress different learned components in different RL methods. For example, to learn a parameterized state transition model $p_{\theta}(s_{t+1}|s_t, a_t)$, the interaction data are used to minimize the following loss:
\begin{align}
    \theta^* = \argmin_{\theta} \E_{(s,a,s^\prime)\sim\mathcal{D}} \left[ \log p_{\theta}(s^\prime | s, a) \right].
\end{align} For learning a parametrized Q-function $Q_{\theta}(s,a)$ under a policy $\pi$, the transition data are used to minimize a temporal-difference (TD) regression loss:
\begin{align}
    \theta^* = \argmin_{\theta}
    \E_{(s,a,r,s^\prime)\sim\mathcal{D}} \left[ \left( Q_{\theta}(s,a) - \left( r + \gamma \E_{a^\prime\sim\pi(\cdot|s^\prime)} \left[ Q_{\bar{\theta}}(s^\prime,a^\prime) \right] \right) \right)^2 \right],
\end{align} where $\bar{\theta}$ is lowly-updated copy of the Q function parameter that serves as the target for TD learning. Given the learned Q-function, a parameterized policy $\pi_{\theta}(a|s)$ can then be updated by maximizing the expected Q-value under the current policy:
\begin{align}
    \theta^* = \argmax_{\theta} \mathbb{E}_{s\sim\mathcal{D},\,a\sim\pi_{\theta}(\cdot|s)} \left[ Q(s,a) \right].
\end{align} 

However, unlike a supervised learning dataset whose samples can be treated as draws from a fixed and stationary data distribution, reinforcement learning data are generated sequentially by policy rollouts. Therefore, the transition tuples $z_t=(s_t,a_t,r_t,s_{t+1})$ are not necessarily drawn independently from a stationary distribution, as consecutive samples are coupled through the state transition distribution $p(s_{t+1} | s_t, a_t)$. The potential violation of the i.i.d. assumption is especially critical for embodied agents, whose states evolve continuously through physical dynamics. Such agents cannot arbitrarily jump between independent states; their experiences must unfold through temporally correlated trajectories. As a result, the data generated by embodied reinforcement learning agents can violate the i.i.d. property at the level of data generation.

Although methods such as \emph{replay buffer} mitigate this issue by storing previously collected transitions $\mathcal{B}=\{z_t\}_{t=0}^{T}$ and sampling mini-batches $\{z_j^\prime\}_{j=1}^{B}\sim\mathrm{Unif}(\mathcal{B})$ during optimization, this only decorrelates data after collection; fundamentally addressing the i.i.d. violation requires revisiting how data are generated by embodied agents, which is simultaneously a control problem and an exploration problem.

\subsection{Diffusion Processes and Ergodicity}

As derived above, the maximum-caliber solution induces the following state transition distribution:
\begin{align}
    p_{\max}(s_{t+1}|s_t) \propto \exp\left( -\frac{1}{2} (s_{t+1}- s_t)^\top C^{-1}[s_t](s_{t+1}-s_t) \right),
\end{align}
or equivalently,
\begin{align}
    s_{t+1} = s_t + w_t, \qquad w_t \sim \mathcal{N}(0, C[s_t]).
\end{align} One of the key properties of this diffusion process is \emph{ergodicity}. The transition kernel $p_{\max}(x_{t+1}\mid x_t)$ is Markovian, and under assumptions that the state space $\S$ is compact and connected, and that the diffusion tensor is full-rank, invertible, Lipschitz, and bounded, this transition kernel has nonzero probability of moving between any two discretized states. Therefore, all states communicate. Moreover, because the Gaussian transition kernel also assigns nonzero probability to remaining at the same state, $p_{\max}(x^*\mid x^*)>0$, the induced Markov chain is aperiodic. Hence, the diffusion process that maximizes the state trajectory entropy is an ergodic Markov chain.

The significance of inducing ergodicity in embodied agents through control policies is that it eliminates the temporal correlation from the physical dynamics over time, such that the trajectory of the embodied agent, as a batch, asymptotically becomes statistically equivalent to a set of i.i.d. samples. 

First, ergodicity implies the existence of a stationary (also called invariant) distribution. Let $\mu_{\max}(s)$ denote this invariant distribution of the diffusion process, satisfying:
\begin{align}
    \mu_{\max}(s^\prime) = \int_{\S} p_{\max}(s^\prime|s)\mu_{\max}(s)\,ds.
\end{align} Since the diffusion process induced by $p_{\max}(s_{t+1}|s_t)$ is ergodic, for any integrable function $f:\S\mapsto\mathbb{R}$, the time average along a single trajectory converges to the ensemble average under the invariant distribution:
\begin{align}
    \frac{1}{T}\sum_{t=0}^{T} f(s_t) \rightarrow \int_{\S} f(s)\mu_{\max}(s)\,ds, \qquad T\rightarrow\infty.
\end{align}
Equivalently, the long-time statistics of a single trajectory become representative of the stationary state distribution.

From the perspective of reinforcement learning, ergodicity provides a formal mechanism for mitigating the i.i.d. data challenge. The issue with embodied data is not merely that samples are generated sequentially, but that nearby transition tuples can be strongly correlated due to the continuity of the physical dynamics. These correlations enter directly into batched regression through covariance terms in the gradient estimate shown in (\ref{eq:batch_grad_cov}). On the other hand, for an ergodic diffusion process with sufficient mixing, the dependence between temporally separated samples decays:
\begin{align}
    \mathrm{Cov}(g_t,g_{t+k}) \rightarrow 0,
    \qquad k\rightarrow\infty.
\end{align}
Thus, the generated data become increasingly representative of the invariant distribution while the effect of temporal correlation is reduced over longer horizons.

Therefore, MaxDiff RL connects exploration, control, and data generation through ergodicity. Rather than treating decorrelation as a post-processing step after data collection, as in replay buffers, maximum diffusion shapes the state trajectory distribution during interaction with the environment. In this view, the i.i.d. data challenge for embodied reinforcement learning is mitigated by synthesizing physically feasible trajectories whose long-time statistics approximate a stationary state distribution and whose temporal correlations decay through diffusive state-space exploration.

\subsection{Connections between MaxDiff RL and Ergodic Control}

MaxDiff RL and ergodic control methods provide similar functionality for control synthesis in the sense that, both provide the tools to specify the behavioral of embodied agents not through temporal representations, but time-averaged spatial representations. However, one of the key differences is that, ergodic control methods explicitly specify the stationary spatial distribution that the trajectory will converge to, while MaxDiff RL guarantees the existence such as a stationary distribution, while implicitly specifying the agent behavior through the reward function and the temperature term. 

Another connection between ergodic control and MaxDiff RL can be made through the perspective of kernel ergodic control~\citep{sun_fast_2025} as discussed in Section \ref{sec:kernel_ergodic_control}. Both frameworks formulate the task as an explicit combination of task-specific reward maximization and state entropy maximization. In kernel ergodic control, state entropy maximization is achieved through the double time integral over a repulsive kernel function. The kernel function measures correlation between states from different time steps globally across the trajectory. On the other hand, the $\log\det C_{\pi}[s^\prime]$ term in the MaxDiff RL objective measures the volume of local trajectory covariance ellipsoid, and can be considered as a log-determinant functional applied to a local covariance kernel, which specifically takes into account the dynamics properties of the system. In particular, for linear time-varying systems, this local trajectory covariance is equivalent to the controllability Gramian~\citep{berrueta_maximum_2024}, meaning that MaxDiff RL explicitly rewards exploration along directions that are locally reachable under the system dynamics.


\chapter{Grand Challenges in Ergodic Control}

\section{Where Does The Target Distribution Come From?}

Let’s revisit the recipes of ergodic control. While there are numerous combinations of search space specification, target distribution representation, ergodic metric formula, and control optimization method, one crucial question remains unanswered: what should the target distribution be? Indeed, while specific implementations matter in practice, at its core, ergodic control is about specifying the temporal actions of the robot based on desired spatial statistics. Yet, in most of the literature on ergodic control, the process of specifying these spatial statistics has been largely overlooked. Therefore, one of the grand challenges in ergodic control is to develop a more systematic view of the role of the target distribution. In particular, we consider this challenge from the following three perspectives.

\textbf{Robo-centric uncertainty prediction.} Closed-loop data collection for robot learning requires models that not only produce predictions but also indicate where and how data should be collected. Prediction uncertainty provides a natural way to characterize what data is needed, and to what extent, to refine the model. While modern generative models are inherently probabilistic and implicitly capture uncertainty, closed-loop data collection requires explicit, quantifiable metrics such as entropy or Fisher information that can be evaluated over the robot’s state or action space and used to define a target distribution. Designing models with this property is therefore often essential. For example, the CVAE-based perception model in~\citet{prabhakar_mechanical_2022,pinosky_embodied_2024} conditions anticipated sensor measurements on the robot’s state and supports analytical entropy evaluation. Systematic design of such robo-centric generative models, potentially extending to advanced architectures such as diffusion models~\citep{ho_denoising_2020} and flow-based models~\citep{lipman_flow_2022}, is an important direction for future work.

\textbf{Optimal data distributions for learning.} Ergodic control offers two formal properties that are highly relevant to the performance of a learned model: the asymptotic coverage guarantee and the i.i.d. property. These properties make ergodic control a potential solution for certifiable robot learning: the coverage guarantee ensures thoroughness of the collected data, while the i.i.d. property ensures that the data is properly decorrelated despite the inherent temporal correlation of robot motions. However, these guarantees only establish that the robot samples a specified target distribution well; they do not establish that the target distribution itself is optimal for learning. For example, in~\citet{pinosky_embodied_2024}, the i.i.d. property bridges the gap between robot learning and PAC-learning theory. Yet PAC-learning theory only guarantees generalization error given a fixed but arbitrary data distribution, requiring only i.i.d. data collection from the chosen distribution. As a result, the choice of distribution in ergodic control remains largely heuristic. A key challenge is therefore to derive target distributions with formal optimal properties for robot learning, potentially drawing on tools from optimal experimental design. In particular, \citep{golovin_adaptive_2011} uses submodular optimization to derive provably near-optimal data distributions for learning; such approaches could be extended to design optimal target distributions for ergodic control and move toward certifiable learning.

\textbf{Meta-level reasoning of target distributions.} Conventional ergodic control is a single-stage decision process: given a target distribution, the robot generates actions to produce a trajectory whose spatial distribution matches that target. However, as motion planning problems grow in complexity, a single fixed target distribution can become insufficient for reasoning across multiple spatial and temporal scales. Frameworks such as task and motion planning (TAMP)~\citep{garrett_integrated_2021} address a similar challenge through hierarchical decision-making that integrates discrete reasoning, discrete-continuous motion specification, and continuous planning. A corresponding challenge for ergodic control is to extend this hierarchy to the target distribution itself, shifting from selecting a single fixed target to a process that iteratively generates and updates distributions at the task level. Recent advances in large models, such as vision-language models (VLMs) and vision-language-action models (VLAs), offer reasoning capabilities well suited for this role. For example, vision-language models have been used to generate spatial uncertainty distributions in response to natural language queries, guiding lower-level exploration for embodied question answering~\citep{das_embodied_2018,ren_explore_2024,ginting_enter_2025}. Combining such reasoning with ergodic control could pair its optimal exploration guarantees with the generative and reasoning capabilities of large models.

\section{Ergodic Control for Large-Scale Problems}

Ergodic coverage problems can span arbitrary domains of variable sizes. 
For example, searching a multi-story building requires reason about the global domain size while locally exploring the hallways and corridors that are significantly smaller than the size of the building~\citep{cao_tare_2021, cao_representation_2023}. 
Likewise, ergodic search can be most useful at planetary scales where a satellite can orbit a planet searching for landing sites, key geological and geomagnetic features, and even signs of life. 
However, the numerical conditioning of these problems\textemdash large-scale domains with small-scale sensing\textemdash can make optimization for ergodic control challenging. 

Solvers used in planning planetary trajectories of satellites and rockets~\citep{ortolano_autonomous_2021}, as well as hierarchical approaches~\citep{wittemyer_bi-level_2023} are promising solutions to the ergodic control problem at large scales. 
These solvers take advantage of governing physical laws that exist at larger-scales to mitigate any numerical challenges while being amenable to local Newtonian physics.
A hybridized approach that integrates the larger-scale planners to coordinate overall coverage goals, with local ergodic control to refine coverage is a promising approach. 
More recently, attempts at formulating an ergodic metric that captures the multiscale nature of large-scale coverage problems like~\eqref{eq:fourier_metric}, with the small-scale resolution of relevant features shows promise as a future research direction~\citep{lahrach_search_2025}. Likewise, a promising direction is exploiting the structure of the large-scale formulation to capture low-dimensional manifolds where information flow is integrated into the ergodic metric to achieve asymptotic coverage~\citep{hughes_asymptotically_2026}.

Additional scalability issues arise when extending exploration beyond the physical workspace to high-dimensional domains, such as configuration spaces or controller parameter manifolds (e.g., stiffness and damping matrices on the $S_{++}^d$ manifold). While low-rank approximations~\citep{shetty_ergodic_2022} and kernel-based formulations~\citep{sun_fast_2025} have proven effective for higher-dimensional ergodic control, achieving efficient coverage in these spaces remains notoriously difficult without exploiting the underlying problem symmetries. Geometric priors that respect topology and symmetry are well-established in the machine learning literature~\citep{bronstein_geometric_2017}. Similar insights can be directly extended to ergodic control to enable scalable exploration on lower-dimensional manifolds. For instance, while a complex system—such as a humanoid or a multi-fingered hand—may possess over 20 degrees of freedom (DoFs), its coordinated movements typically lie on much lower-dimensional manifolds. To exploit this property, recent work \citep{low_cooperative_2026} employed similarity transformations of geometric primitives to represent lower-dimensional cooperative task spaces. Lie algebra of these transformations can be used for ergodic control providing a dimensionality reduction from more than 20 dimensions to fewer than 7. A similar structural efficiency applies to object interaction. Relying on a single fixed Cartesian frame (whether world- or body-fixed) is often inefficient for exploring curved geometries. Instead, adopting cylindrical or spherical coordinate systems exploits natural symmetries \citep{ti_geometric_2023}, while object-centric local frames can align the search space with task-relevant directions even if the objects lack global symmetry~\citep{bilaloglu_object-centric_2026}. By capturing these intrinsic symmetries and geometric invariants, the search space can be effectively reduced from more than 6 dimension to fewer than 3 and can be used for decomposing controller parameters to shape-invariant, decoupled translational and rotational components.

\section{Lifelong Learning and Long-horizon Problems}
Lifelong learning is essential for robots operating in previously unseen settings to start operation using prior knowledge, explore to acquire new information, and continuously adapt over their operational lifespan. Ergodic control offers a different perspective for this paradigm, as it provides a principled way to balance and schedule exploration and exploitation according to the statistics of the target spatial distribution. However, deploying ergodic control in dynamic, lifelong settings necessitates establishing asymptotic coverage guarantees under evolving target distributions, as well as long-horizon scalability.

Ergodic control can encode a continuous spectrum of exploration and exploitation behaviors by changing the statistics of the target distribution (i.e. distribution tracking). At one extreme, one can set the target distribution for the ergodic controller to a Dirac delta distribution and obtain a simple reaching controller (pure exploitation). At the other extreme, a uniform target leads to a coverage controller (pure exploration). In realistic applications, the true underlying objective is rarely static or known with absolute certainty. The target distribution inherently lies between these two extremes: it cannot be a Dirac delta due to persistent sensorimotor uncertainity and unmodeled dynamics; nor should it be uniform, as prior information bias where the robot should explore. Furthermore, as a robot interacts with its environment over an extended operational lifespan, newly visited states and incoming observations must continuously refine this target distribution. Previous work has explored nonparametric updates \citep{abraham_ergodic_2017, ivic_search_2020} and parametric updates of the target distribution \citep{miller_ergodic_2016, mavrommati_real-time_2018}. Nevertheless, establishing systematic target updates for lifelong learning within the ergodic control framework is challenging. Additionally, it remains an open question whether the stationarity of the underlying distribution is sufficient for the asymptotic coverage guarantees of ergodic controllers when the target distribution provided to the controller evolves over time.

An inherent challenge of lifelong learning is scaling to long-horizon settings and what to keep in memory and what to forget~\citep{billard_roadmap_2025}. Fourier ergodic metric and field-based approaches like HEDAC circumvent this by recursively accumulating trajectory statistics into a fixed-dimensional state vector with a constant update cost per time step. Conversely, kernel-based methods struggle in long-horizon scaling because explicit history of visited state points leads to quadratic scaling bottlenecks. Recent work by~\citet{hughes_infinite-horizon_2026} directly addresses this by deriving an extended error state that recursively accumulates past visitation history, decoupling past coverage from future trajectory optimization. Similarly kernel approaches can be made more computationally efficient over long horizons by leveraging Random Fourier Features (RFF) or explicit basis function approximations to project the kernel into a low-dimensional feature space, restoring constant recursive accumulation. However, domain sampling schemes and spectral approximations still suffer from the curse of dimensionality, quickly becoming intractable as the state-space dimension grows. Consequently, designing ergodic control formulations that simultaneously scale to high spatial dimensions and infinite time horizons remains an open challenge.

\section{Ergodic Control Beyond Exploration and Coverage}
Existing methods in ergodic control predominantly focus on tasks that can be framed as exploration or spatial coverage and time-averaged trajectory statistics alone fall short for the vast majority of robotic tasks. Nevertheless, a majority of tasks might benefit from ergodic control principles to achieve robustness against sensorimotor uncertainty and distribution shifts. Crucially, it remains an open research question how far ergodic control principles can be extended beyond coverage tasks, and how they can be effectively interfaced with contemporary frameworks for multi-task and long-horizon autonomy.

As discussed in detail in Chapter~\ref{chap:when_not_to_use_ergodic_control}, ergodic control is, by design, insensitive to the temporal ordering of a trajectory. Still, a promising middle ground is to introduce \emph{soft temporal guidance}, enabling ergodic controllers to realize a continuous spectrum between strict trajectory tracking and pure distribution coverage rather than requiring discrete mode switches. For example, while ergodic control has proven effective for surface cleaning tasks, pre-operation grasping of the cleaning tool and post-operation placement typically demand separate reaching controllers governed by explicit state-machine switching. An appealing alternative is to retain a unified ergodic formulation across all phases, dynamically modulating the statistics of the target distribution and embedding temporal priors through the initialization of the trajectory optimizer. A key insight is that the non-convexity of ergodic trajectory optimization naturally yields multiple locally optimal trajectories under different initial conditions \citep{miller_trajectory_2013}. This property has previously been exploited to generate diverse exploratory trajectories via Stein variational ergodic search \citep{lee_stein_2024}. Extending this concept, preliminary work on adaptive ergodic imitation \citep{xu_ergodic_2026} combines pre-trained imitation learning policies with ergodic control to enable exploratory online adaptation to environmental shifts. Soft temporal sequencing is injected through the initialization of ergodic trajectory optimization exploiting the non-convexity. Pretrained imitation learning policy act as an observation-conditioned target generator and this target is modulated to an adaptive target distribution for the ergodic controller using anisotropic diffusion. Consequently, this pipeline aims to bridge imitation with exploratory adaptation to extend ergodic imitation \citep{kalinowska_ergodic_2021, shetty_ergodic_2022} toward more general tasks with potentially sequential components. A promising alternative for adaptive ergodic imitation might be to use Stein variational formulations~\citep{lee_stein_2024,sirigiri_diversifying_2025} since they can leverage the learned score function from the diffusion models directly.

While bridging modern imitation learning with ergodic control offers a compelling pathway several key challenges remain. In particular, it remains an open question how to systematically balance sequential aspects against trajectory statistics, maintain computational scalability in high-dimensional, long-horizon settings, and preserve theoretical asymptotic coverage guarantees under non-stationary target distributions.




\backmatter  

\printbibliography

\end{document}